# Text Classification of the Precursory Accelerating Seismicity Corpus: Inference on some Theoretical Trends in Earthquake Predictability Research from 1988 to 2018

A. Mignan*

*Version date: 24 August 2018*

\* Swiss Federal Institute of Technology Zurich, ETHZ, NO H66, Sonneggstrasse 5, CH-8092 Zurich

*Contact address:* arnaud.mignan@sed.ethz.ch

*Abstract:* Text analytics based on supervised machine learning classifiers has shown great promise in a multitude of domains, but has yet to be applied to Seismology. We test various standard models (Naïve Bayes, *k*-Nearest Neighbors, Support Vector Machines, and Random Forests) on a seismological corpus of a hundred articles related to the topic of precursory accelerating seismicity, spanning from 1988 to 2010. This corpus was labelled in a previous study [Mignan, *Tectonophyscs*, 2011] with the precursor whether explained by critical processes (i.e., cascade triggering) or by other processes (such as signature of main fault loading). We investigate rather the classification process can be automated to help analyze larger corpora in order to better understand trends in earthquake predictability research. We find that the Naïve Bayes model performs best, in agreement with the machine learning literature for the case of small datasets, with cross-validation accuracies of 86% for binary classification. For a refined multiclass classification ('non-critical process' < 'agnostic' < 'critical process assumed' < 'critical process demonstrated'), we obtain up to 78% accuracy. Prediction on a dozen of articles published since 2011 shows however a weak generalization with a F1-score of 60%, only slightly better than a random classifier, which can be explained by a change of authorship and use of different terminologies. Yet, the model shows F1-scores greater than 80% for the two multiclass extremes ('non-critical process' versus 'critical process demonstrated') while it falls to random classifier results (around 25%) for papers labelled 'agnostic' or 'critical process assumed'. Those results are encouraging in view of the small size of the corpus and of the high degree of abstraction of the labelling, and demonstrate the potential of supervised learning to reveal textual patterns. Domain knowledge engineering remains essential but can be made transparent by an investigation of Naïve Bayes keyword posterior probabilities.

## 1. Introduction

Text analytics (or text mining) uses tools from information retrieval, natural language processing (NLP), and machine learning, to autonomously extract information from corpora (Sebastiani, 2002; Aggarwal, 2018). Prominent in Web data classification (Sebastiani, 2002; Tsytsarau and Palpanas, 2012; Kharde and Sonawane, 2016), text analytics is also commonly used in science, such as social sciences (Grimmer and Stewart, 2013) or biomedical sciences (Ng and Wong, 1999). The present article is the first to investigate the potential of text analytics in Seismology.

Analysis of corpora from the seismological literature remain rare, and are so far based on domain knowledge with no use of text analytics (Mignan, 2011; 2014). These early investigations, with manual extraction of information and categorization (so-called knowledge engineering), remain somewhat non-transparent and/or difficult to scale to larger corpora. Those types of meta-analyses however provide useful conclusions regarding the scientific process, including status-quo biases, paradigm shifts, and other patterns (Kuhn, 1970), which can then be used as input to improve the scientific debate. For instance, Mignan (2011) used the precursory accelerating seismicity corpus (86 articles) to identify a scientific cycle composed of three phases: early works, criticality paradigm, and divergent directions (Fig. 1). This meta-analysis also showed the implications of different theoretical trends on the stated predictability of large earthquakes and that this scientific process was dynamic, with current research being at a "crisis" stage (see Hough (2010) for a novelized review of earthquake prediction research cycles). Despite the present "earthquake prediction winter", a better understanding of the history of earthquake predictability remains crucial for the next research phase.

Here we will reinvestigate the precursory accelerating seismicity corpus already labelled by Mignan (2011) and we will extend it by including the articles published since then, yielding a corpus of 101 articles for the past 30-year period (1988-2018). We will test various machine learning algorithms for supervised learning (Naïve Bayes, *k*-Nearest Neighbors, Support Vector Machines, and Random Forests) and apply them to the text domain (Sebastiani, 2002; Aggarwal, 2018). We will discuss the specificities of text classification (sparse high-dimensional non-negative data) and of our corpus (small and slightly unbalanced dataset) to be considered for both feature and classifier selection. Our aim is threefold: (1) to illustrate, for the first time, text classification procedures in Seismology, (2) to improve the meta-analysis initiated by Mignan (2011), by making information extraction (i.e., theoretical trends) more transparent and scalable with autonomous learning, and (3) to engage the Seismological community in debating scientific issues using the concept of metascience (Pearce and Rantala, 1983).

## 2. Methods & Data

*2.1. Corpus Definition & Text Analytics Prerequisites*

Our corpus is composed of $n_{all}$ = 101 articles that explore the phenomenon of accelerating seismicity observed prior to large earthquakes. The training set is composed of $n_{train}$ = 86 articles published between 1988 and 2010, and which have been labelled by Mignan (2011). The test set is composed of $n_{test}$ = 15 articles, corresponding to the articles published since 2011. Following the same approach as in Mignan (2011), works not considered here are the ones where precursory accelerating seismicity is not the main topic, the term is used to mean short-term foreshocks (events occurring in the near-field during the nucleation phase; see review by Mignan, 2014), or where tectonic earthquakes are not the target (e.g., volcanic precursors, laboratory acoustic emissions, *etc.*).

The corpus is treated as a bag-of-words, from which we define an $n \times d$ document-term matrix (DTM), which is a sparse high-dimensional non-negative matrix $D = \{\bar{X}_1, ..., \bar{X}_n\}$ with $\bar{X}_i$ a $d$-dimensional vector, following Aggarwal (2018)'s notation. Each possible word from the corpus lexicon corresponds to one dimension, or to one feature. We extract text from our corpus following two different approaches, one considering the full articles (from PDF files) and another the meta-data only, consisting of the title, publication year, author list, abstract and keywords (which can be extracted from each PDF, or from website queries). While the first approach yields a larger set of tokens (or total number of words), the second approach gives a subset of the full texts that condenses the authors' vision while avoiding journal paywalls. Using the meta-data instead of the PDF provides a natural approach to feature reduction and can take advantage of web scraping tools to automatize the corpus creation (Glez-Peña et al., 2013).

We use the unigram model, which is equivalent to the bag-of-words, and yields a lexicon of $d$ = 21,017 terms (or dimensions) for the full texts and of $d$ = 1,891 terms for the meta-data, with only terms composed of at least 3 characters considered in the full text corpus. The tokens are subsequently cleaned by removing stop-words (common English words, such as articles, pronouns, *etc.*) and punctuation, by converting upper cases into lower cases, and by consolidating related words which have the same root into a single term (i.e. process of stemming, e.g., 'accel' = {'accelerating', 'accelerated', 'acceleration'}). All of those preliminary text mining steps are done with the *quanteda* R package (Benoit, 2018).

We test as feature-value weighting scheme both the term frequency and the term frequency-inverse document frequency (tf-idf) model, in which the inverse document frequency $\log(n/n_i)$ is multiplied with the term frequency, with $n_i$ the number of documents in which the $i$th term occurs (Salton and McGill, 1983). Normalization by document length is only done when it increases cross-validation accuracy.

*2.2. Class definition and labelling*

We first define the following binary class: 1 = 'critical process assumed or demonstrated' versus 0 = 'else'. We then refine the labelling with an ordered multi-class categorization of the theoretical trends with: 0 (= 'non-criticality assumed or demonstrated') < 1 (= 'theoretically agnostic') < 2 (= 'criticality assumed') < 3 (= 'criticality demonstrated').

Let us now explain the rationale behind categorizing the precursory seismicity literature around the concept of Criticality (Sornette, 2000). Criticality is a generic term encompassing Self-Organized Criticality (SOC; Bak and Tang, 1989) and the Critical Point theory (Sammis and Sornette, 2002). It is related to Complexity theory, which can be summed up as dynamic processes controlled by bottom-up triggering leading to emergent phenomena at the system level (i.e., holistic system). In contrast to criticality would be non-criticality, where processes may be static and are controlled by top-down triggering (often referred to as loading, e.g., from a main fault). In that case, the system can be reductionist. The debate of criticality versus non-criticality is important since the two diametrically opposed views offer different conclusions as to the precursory seismicity phenomenology and to the predictability of earthquakes (Mignan, 2011; 2014). It is also a fundamental question, as most earthquake theories, if not all, can be related to one or the other view.

On top of the obvious reason for using the precursory accelerating seismicity corpus, which is the availability of an already defined and labelled dataset (Mignan, 2011), another one is the relatively simple debate illustrated by this precursory pattern, which is defined by a power-law time-to-failure equation (Bufe and Varnes, 1993). Power-laws are often described as the signature of a critical process (Bak and Tang, 1989) although they can also be explained by geometric top-down processes (King, 1983; Mignan et al., 2007; Mignan, 2012). Then, whether the process is truly critical or non-critical has tremendous implications for earthquake predictability. If the system is controlled by SOC, then earthquakes would be unpredictable (Geller, 1997) and accelerating seismicity patterns would only represent random fluctuations of the natural clustering behavior of seismicity (Hardebeck et al., 2008), which is commonly described by the Epidemic-Type Aftershock Sequence (ETAS) model (Ogata, 1988; Seif et al., 2017 and references therein). If the system is a variant of SOC however, the system-level event (i.e., mainshock) could show some degree of predictability with accelerating seismicity representing some cascading triggering towards the critical point (Sammis and Sornette, 2002). In a non-critical model, accelerating seismicity would be the signature of loading on the fault that will host the mainshock (Mignan, 2012). While the critical point and top-down loading expect some degree of predictability, the accelerating seismicity phenomenology differs. Also, in criticality, patterns are expected to emerge on average by timeseries stacking while in non-criticality, patterns may change from timeseries to timeseries with stacking having for effect to smooth out the precursory pattern. Theoretical choices have therefore a clear impact on the outcome, as proven also for the prognostic value of short-term foreshocks (Mignan, 2014).

Figure 1 shows the number of articles per class as a function of publication year. The Kuhnian cycle already described by Mignan (2011) is visible and composed of three successive phases: early works, criticality paradigm, and divergent directions. For the training set, the binary labelling follows the one of Mignan (2011) with 1 ('critical process assumed or demonstrated') = {'x', '+'} and 0 ('else') = {'-', ' '} (see their Table 1). Mignan (2011) labelled as 'unclear' ('~') 6 articles. Here we decided whether they correspond to 0 or 1 (see the supplementary material for the full list of labels). The multiclass labelling also tries to follow Mignan (2011) with 0 ('non-criticality assumed or demonstrated') = '-', 2 ('criticality assumed') = 'x' and 3 ('criticality demonstrated') = '+'. For their empty class ' ', most were related to our class 1 ('theoretically agnostic') but the articles where non-criticality was assumed were moved to our class 0. As previously mentioned, the knowledge engineering approach that led to this labelling was not explicit in Mignan (2011). It is also not simple to describe, as illustrated in Table 1 for the test document set. Obviously, the meta-analysis would become more transparent and objective if one could autonomously classify the articles per theoretical trend. More generally, this would have important implications if the process could be scaled to larger corpora, for example to the complete literature on earthquake predictability research.

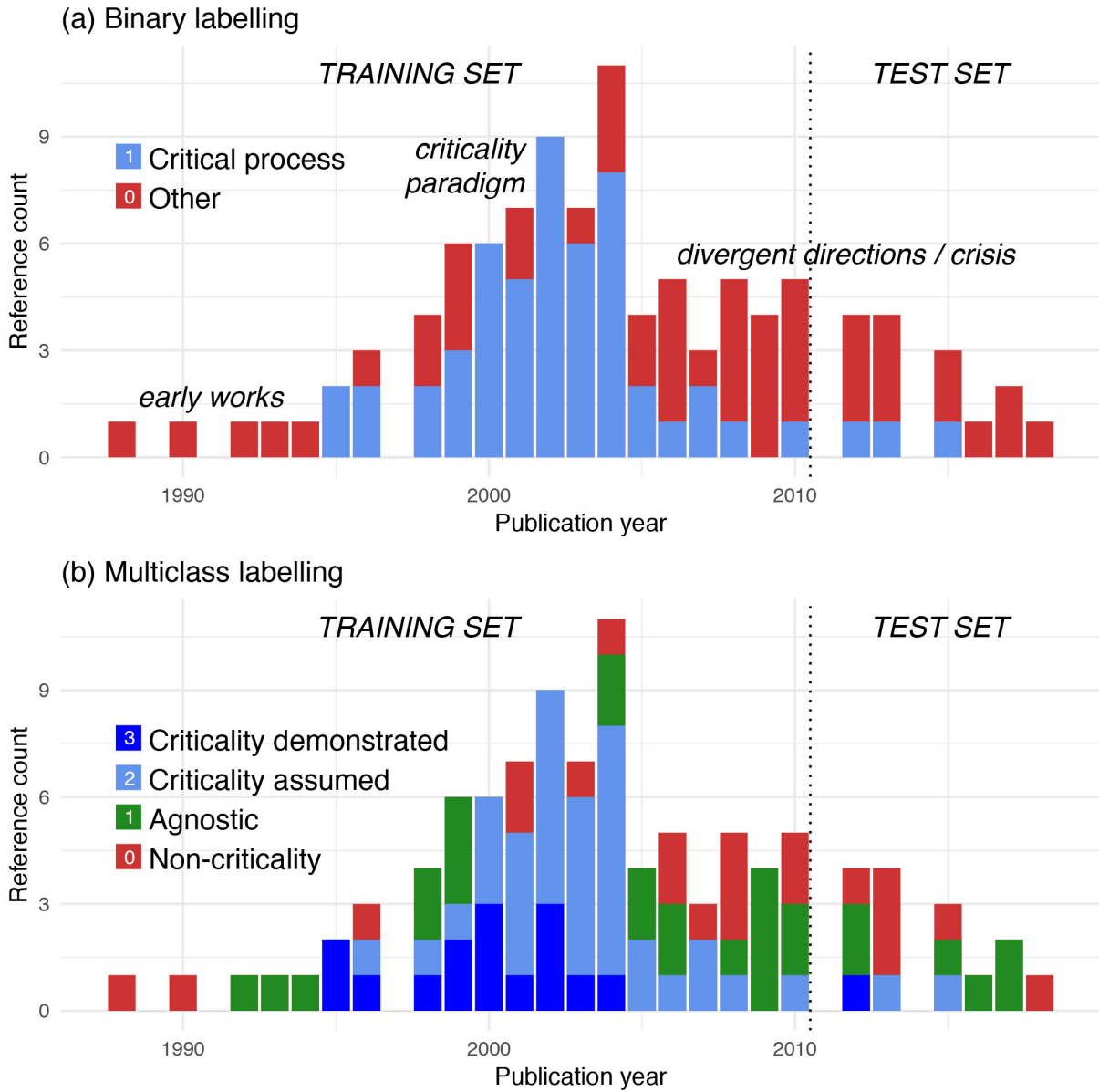

**Fig. 1.** The two categorizations defined to represent the precursory accelerating seismicity 1988-2018 corpus: (a) Binary classes 'critical process' or 'else', based on the results of the meta-analysis by Mignan (2011), updated for the period 2011-2018; (b) Multiple ordered classes 'non-criticality demonstrated or assumed' < 'agnostic' < 'criticality assumed' < 'criticality demonstrated' - Labels were determined using domain knowledge (see e.g., Table 1).

### 2.3. Supervised Learning for Text Classification

The DTMs defined in section 2.1 are used as input to test different machine learning classifiers. We consider some of the most common methods: Naïve Bayes (Domingos and Pazzani, 1997), *k*-Nearest Neighbors (Cover and Hart, 1967), Support Vector Machines (Cortes and Vapnik, 1995), and Random Forests (Breiman, 2001). The *k*-Nearest Neighbor, Support Vector Machines, and Random Forest are sketched in Figure 2. Deep learning approaches such as artificial neural networks (Rumelhart et al., 1986) are not considered since our corpus is too small, nor are boosting techniques (Freund and Schapire, 1999) used to improve any of the above classifiers. We also do not experiment with the ensembling of different classifiers (Rokach, 2010). Hyperparameter tuning based on cross-validation results will be described in section 3.1.

The Naïve Bayes classifier is based on Bayesian probability theory with the term 'naïve' referring to the assumption that all features are independent. Although this is rarely true, naïve Bayes models perform relatively well in many problems (Domingos and Pazzani, 1997). Bayes' theorem defines the posterior probability as

$$P(c|\bar{X}) = \frac{P(c)P(\bar{X}|c)}{P(\bar{X})} \qquad (1)$$

with $P(\bar{X}|c)$ the likelihood, $P(c)$ the prior probability and $P(\bar{X})$ the marginal likelihood, with $\bar{X}$ the feature vector of the document and *c* the class. For *d* features, we get

$$P(c|\bar{X} = \{x_1, \ldots, x_i, \ldots, x_d\}) \propto P(c) \prod_{i=1}^{d} P(i|c)^{x_i} \qquad (2)$$

which is the multinomial Naïve Bayes classifier of a given document, with $x_i$ the weighted frequency of term *i*. Note that the marginal is constant and thus only acts as scaling factor. We test two different priors *P(c)*, the uniform distribution and the document frequency of *c*. Note that the corpus is unbalanced in opposite directions for the training set (58% in class 1) and test set (20% in class 1) for the binary case. We use the Naïve Bayes classifier *textmodel_nb* of the *quanteda* R package (Benoit, 2018) with Laplace smoothing of 1 to avoid null probabilities.

The *k*-Nearest Neighbors (*k*NN) classifier (Fig. 2a) is a simple non-parametric method where documents are classified to the majority class in the cluster composed of the *k* closest instances in the feature space (Cover and Hart, 1967). We use Euclidean distance combined to a cosine (similarity) kernel function to reduce weights on orthogonal documents (Hechenbichler and Schliep, 2004). The value of the parameter *k*, which thus specifies the number of neighbor observations that contribute to the predictions, is optimized via cross-validation. We did not test other Minkowski distances (i.e., other than Euclidean). We use the *k*NN classifier of the *KernelKnn* R package (Mouselimis, 2018).

A Support Vector Machine (SVM) divides the *d*-dimensional space into partitions of similar classes, by searching for the Maximum Margin Hyperplane (MMH) that creates the greatest separation between two classes. The so-called support vectors are the points from each class which are the closest to the MMH (Fig. 2b). When the data are non-linear, kernels are used to map the data into a linear problem in a higher dimensional space. We here test the following kernels: linear, polynomial and radial basic function networks (Joachims, 1998). Regularization is used to avoid overfitting, allowing instances to fall off the MMH but subject to the 'slack penalty' $C = 1/\lambda$ (here = 1), with $\lambda$ the regularization parameter. We refer the reader, who wants to learn more about the mathematics of SVMs, to Cortes and Vapnik (1995); Bennett and Campbell (2000); Steinwart and Christmann (2008). We use the *ksvm* classifier of the *kernlab* R package that uses Euclidean distance (Karatzoglou et al., 2004).

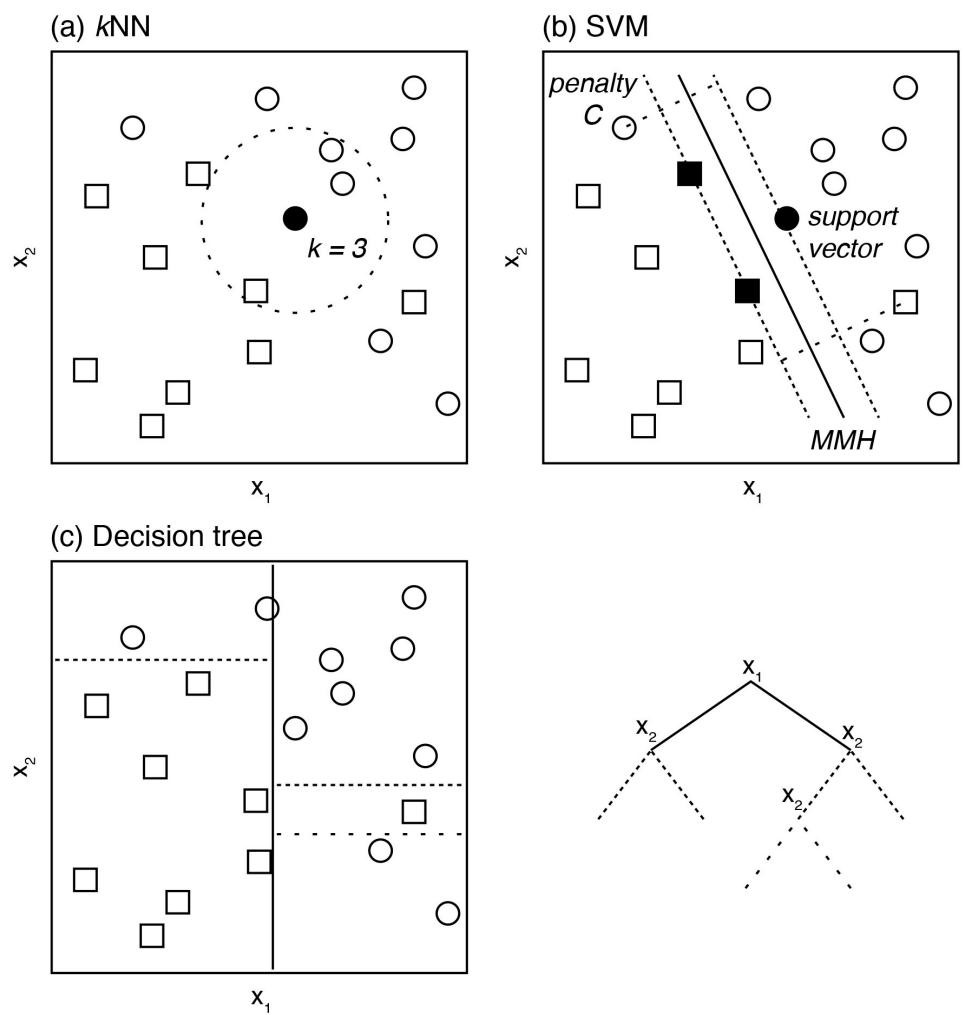

**Fig. 2.** Sketches illustrating three classic machine learning classifiers: (a) *k*-Nearest Neighbor *k*NN; (b) Support Vector Machine SVM; and (c) Decision tree (one instance of a Random Forest). The squares and circles represent two different classes characterized by the two features $x_1$ and $x_2$.

A decision tree (Fig. 2c) makes a hierarchical partitioning of the data space, in which the partitioning is done recursively, top-down, using split conditions on the features (Breiman et al., 1984). For text, the split condition corresponds to constraints on the frequency of one term (Aggarwal, 2018). At each step, two new branches are created, corresponding to the presence or absence of a term. The final nodes (leaf nodes) are labelled, meaning that a test document will be checked for the presence or not of each term of the tree, following the matching branches down to the label given by the leaf node. To avoid overfitting, pruning is generally done (by holding out part of the training data). Another option, used here, is the Random Forest classifier, which bootstraps decision trees grown to full height (Breiman, 2001). The main parameters are the number of features $n_{try}$ randomly selected at each node and the number of trees $m_{tree}$ from which an average is made (i.e., bagging approach). We use the *randomForest* R package (Liaw and Wiener, 2018) in which the split condition is based on a variant of the Gini index (Breiman et al., 1984).

**3. Results**

*3.1. Cross-Validation & Model Selection (Binary Class only)*

We select the feature weights and machine learning hyperparameters based on a leave-one-out (i.e., *n*-fold) cross-validation, which consists in training on all the data except for one document on which the prediction is made (Kohavi, 1995). The classifier is then evaluated by averaging the *n* results. This is computationally non-intensive since we have only $n_{train} = 86$ training documents. Results are shown in Table 2 in terms of accuracy, F1-score, precision and recall. The accuracy is defined as the fraction of test instances in which the predicted class value matches the label. The F1-score is the harmonic mean of precision and recall. Precision is defined as the proportion of positive instances (i.e., true positives + false positives) that are truly positive and recall is defined as the number of true positives over the total number of positives (i.e., true positives + false negatives).

A random classifier would yield a 50% accuracy, but since our binary classes are 1 ('critical process assumed or demonstrated') and 0 ('else'), we must benchmark our machine learning classifiers against the obvious search of the keyword 'critic*'. This is a simple rule-based classifier that predicts 1 if the term 'critic*' is present and 0 if absent from the document. As we notice from Table 2, the accuracy of a classifier should be higher than 80% for the meta-data (composed of the list of authors, title, abstract, and keywords) and 63% for full texts (F1-score > 75% and 82%, respectively). The difference in accuracy of this baseline classifier is due to the fact that studies assuming or demonstrating criticality are more likely to use that term in the title and/or abstract than other studies, while any study may refer to criticality within the main text, for example when discussing the background literature.

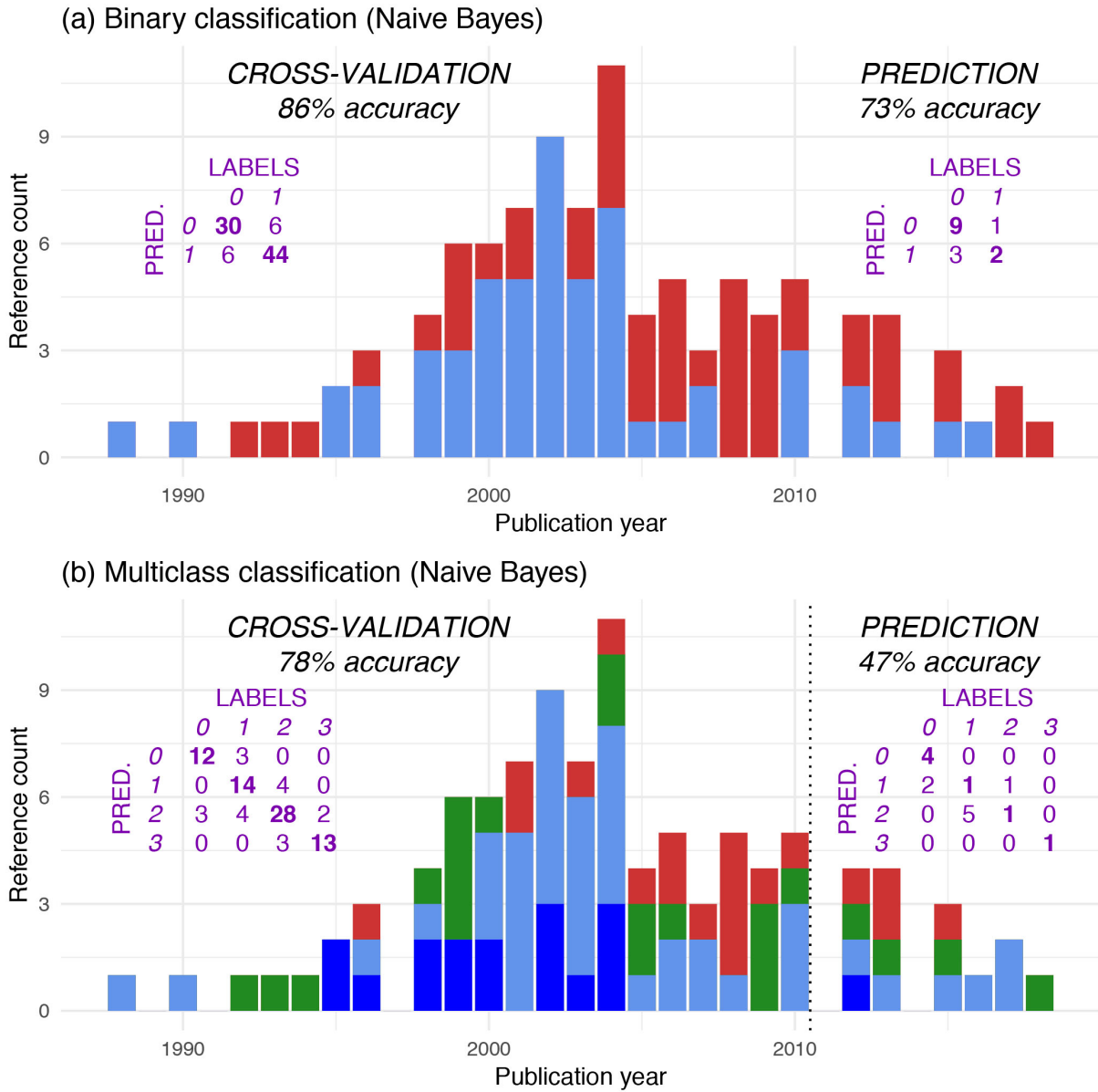

**Fig. 3.** Naïve Bayes results (including confusion matrices) for leave-one-out cross-validation on training document set (1988-2010) and for test document set prediction (2011-2018) - both cases for full texts: (a) Binary class; (b) Multiclass. See Figure 1 for comparison with the original labels and for the class color scheme.

The best results of the five machine learning classifiers, as listed in Table 2, were obtained as follows: for all cases, better results were obtained for full texts if the number of features was reduced by removing all terms with a frequency of occurrence lower than 5. We also found that using the term frequency gave similar or better results than the tf-idf model (Forman, 2008), which suggests that some frequent words that tf-idf possibly penalizes too strongly are important for classification. Further document length normalization only improved results for *k*NN and was automatically done by *textmodel_nb* for Naïve Bayes. Specific hyperparameter tuning was as follows: we used a document frequency prior for Naïve Bayes, $k = 3$ for *k*NN, a linear kernel for the SVM, and $n_{tree} = 500$ and $n_{try}$ the square root of the total number of features for the Random Forest. All classifiers perform better than the keyword search for full texts with an accuracy equal to or greater than 78% (F1-score $\geq 81\%$). For metadata however, only the Naïve Bayes classifier performs better than a simple keyword search with an accuracy of 84% and F1-score of 86%. Those scores are reasonable compared to, for instance, sentiment classification on the Web (Tsytsarau and Palpanas, 2012; Kharde and Sonawane, 2016). Overall, the Naïve Bayes classifier performs better than all other classifiers, although the differences range from 10% to as low as 1%. Note that using different strategies (e.g., different hyperparameter tuning, use of boosting) may also yield better accuracies. Results of Table 2 are however consistent with the bias-variance tradeoff that states that models with a high bias (such as Naïve Bayes) perform better on rather small training sets (Ng and Jordan, 2001). For the rest of this work, we will only consider the multinomial Naïve Bayes model with document frequency prior. This model has also the advantage to be computationally cheap and transparent.

*3.2. Interpreting Naïve Bayes Results*

Let us now investigate the Naïve Bayes results for the binary class in more detail. Figure 3a shows the results in time series form, as well as the confusion matrix (the test set results will be discussed in section 3.3). Of 86 training documents, 6 were wrongly categorized as 0 and 6 as 1. A ranking of posterior probabilities $P(c|i)$ (Eq. 1; Table 3) shows that terms *i* best representative of class 1 (= 'critical process assumed or demonstrated') include (in stem form): 'powerlaw', 'automat', 'fibre-bundl', 'lattic', 'renormalizationgroup', *etc.*, which are clearly related to the concept of critical phenomenon (Sornette, 2000). On the other side of the probability distribution, we find: 'backslip', 'prestress' or 'coulomb' suggestive of studies more focused on seismotectonic aspects than fundamental physics. However, we also find other terms correlated to one or the other class, which are intrinsically independent of the physical process, such as authors, methods or region names, showing that authors often follow one theoretical view, may favor some methods (e.g., 'lurr', acronym for Load/Unload Response Ratio, or 'ellipt' for use of elliptical areas, see Table 3) and only work on one specific region. Such sampling bias is expected to not generalize well. For example, let's consider the study made by Papadopoulos (1988), labelled 0 and misclassified 1. We can find the reason in the fact that the terms used in many Greek works labelled 1 from the Papazachos group (e.g., Papazachos et al., 2007 and references therein) are also used by Papadopoulos (1988), such as 'hellen' for Hellenic ($P(c = 1|\text{'hellen'}) = 0.77$) or 'peloponnesus' ($P(c = 1|\text{'peloponnesus'}) = 0.69$), references to 'papazacho' ($P(c = 1|\text{'papazacho'}) = 0.81$), use of 'ellipt', *etc.*. Only a larger corpus is likely to reduce the weight of such associations.

For the multiclass categorization, a random classifier would yield a 25% accuracy. The Naïve Bayes model yields, in leave-one-out cross-validation, 78% accuracy and {80%, 72%, 78%, 84%} F1-score for classes 0, 1, 2 and 3, respectively, on the full texts, and 66% accuracy and {74%, 50%, 68%, 73%} F1-score on meta-data only (Table 4). This result is encouraging since the intermediary classes are closely related, with similar lexicons. It means for example that 'criticality assumed' (class 2) and 'criticality demonstrated' (class 3) can be distinguished to some degree from a simple bag-of-words. Results are shown in Figure 3b with the confusion matrix showing how errors are distributed. Note that no document of class 0 is misclassified in class 4 nor *vice versa*. This verifies the proposed ordering of classes 0 (= 'non-criticality assumed or demonstrated') < 1 (= 'theoretically agnostic') < 2 (='criticality assumed') < 3 (= 'criticality demonstrated'). Table 5 shows some of the terms leading to high posterior probabilities $P(c|i)$ for each class $c$, which can easily be related to domain knowledge engineering. For class 0, seismotectonic terms dominate; for class 1, terms related to different regions and statistical tests dominate; for class 3, common methods used in criticality studies dominate; finally, for class 4, fundamental physics terms dominate.

*3.3. Test Data Prediction with the Naïve Bayes Model*

Figure 3 shows the test data results for the 2011-2018 part of the corpus (Table 1) for both binary and multiclass. The Naïve Bayes model built from the training document set yields for the test set 73% accuracy (for both full texts and meta-data). However, the F1-score provides a better metric with 50% and 60%, respectively (Table 6). It means that the model is only slightly better than a random classifier for the binary class. Note also that the search of the term 'critic' is worse than a random classifier for full texts. Results are better for the multiclass prediction (Table 7) with an accuracy between 47 and 53% (25% for a random classifier). F1-scores show that the extreme classes 0 (= 'non-criticality assumed or demonstrated') << 3 (= 'criticality demonstrated') are best predicted with F1-score = 80% and 100%, respectively, for full texts (67 and 100% for the meta-data). For full texts, the Naïve Bayes model predicts the intermediary classes 1 (= 'theoretically agnostic') and 2 (='criticality assumed') at the level of a random classifier (with F1-score = 20 and 25%, respectively). The result is slightly better for the meta-data (with F1-score = 40 and 44%, respectively).

These results prove that the model does not generalize well, which was expected from some of the correlations observed during cross-validation. The results are however encouraging in view of the many difficulties associated with the present data set: (1) the training set is only composed of 86 documents; (2) the class distribution of the training set does not match the target distribution, being unbalanced in different directions with 58% of the training documents in class 1 and only 20% of the test set in class 1 (in binary class); (3) the test set has a rather small statistical overlap with the training set since both sets represent two independent time periods. Although some term weights remain valid, as proven by the results better than random ones, changes in terminology due to different authorships can in part explain the drastic decrease in accuracy.

## 4. Conclusions

We presented the first study that applies machine learning to Seismological text classification, which, we hope, will encourage its use by the community. We showed that the Naïve Bayes model is the best performing classifier, which is likely due to the small size of the corpus considered. While other classifiers ($k$NN, SVM, Random Forest, *etc.*) may work better on larger training document sets, Naïve Bayes combines two advantages, fast computation and transparency. We showed for example that this model allows making the knowledge engineering process more transparent.

Was manual labelling required in the first place? Applying *k*-means to the document-term matrix (DTM) leads to clusters of authors, suggesting a natural clustering of paper by writing style. This is further confirmed by applying Latent Dirichlet Allocation (LDA; Blei et al., 2003) to the DTM with no obvious topic emerging. This suggests that no data-driven guidance is provided in this corpus to define highly abstract classes relating to physical trends. Domain knowledge engineering is therefore still required.

One aim of text classification for earthquake predictability research could be to update theoretical trends online, any time a new research article is published. This would in principle clarify the current state of research in the field. We showed that the Naïve Bayes classifier generalizes poorly on a different time period, meaning that the possibility of new trends, with different term statistics, could be missed. Only training on a larger corpus is likely to improve the text classification. This could be done by applying knowledge engineering on the full earthquake prediction literature.

*Data and Resources:* All the corpus articles are available on journal websites. The corpus metadata and labelling are provided in the supplementary material to this article.

*Acknowledgments:* I thank Pablo Nieto and Marco Broccardo for discussions on the topic of text classification.

**References**

Aggarwal, C. C. (2018). *Machine Learning for Text*, Springer Nature, 493 pp., doi: 10.1007/978-3-319-73531-3.

Bak, P., and C. Tang (1989). Earthquakes as a Self-Organized Critical Phenomenon, *J. Geophys. Res.* **94** 15,635-15,637.

Bennet, K. P., and C. Campbell (2000). Support Vector Machines: Hype or Hallelujah?, *SIGKDD Explorations* **2** 1-13.

Benoit, K. (2018). Quantitative Analysis of Textual Data, Package 'quanteda', available at https://cran.r-project.org/web/packages/quanteda/ (last assessed August 2018)

Blei, D. M., A. Y. Ng, and M. I. Jordan (2003). Latent Dirichlet Allocation, *J. Mach. Learn. Res.* **3** 993-1022.

Breiman, L., J. H. Friedman, R. A. Olshen, and C. J. Stone (1984). *Classification And Regression Trees*, Chapman & Hall/CRC, Taylor & Francis Group, 358 pp.

Breiman, L. (2001). Random Forests, *Machine Learning* **45** 5-32.

Bufe, C. G., and D. J. Varnes (1993). Predictive Modeling of the Seismic Cycle of the Greater San Francisco Bay Region, *J. Geophys. Res.* **98** 9,871-9,883.

Cortes, C., and V. Vapnik (1995). Support-Vector Networks, *Machine Learning* **20** 273-297.

Cover, T. M., and P. E. Hart (1967). Nearest Neighbor Pattern Classification, *IEEE Transac. Information Theory* **IT-13** 21-27.

Domingos, P., and M. Pazzani (1997). On the Optimality of the Simple Bayesian Classifier under Zero-One Loss, *Machine Learning* **29** 103-130.

Forman, G. (2008). BNS Feature Scaling: An Improved Representation over TF-IDF for SVM Text Classification, *ACM 17th Conf. Info. and Knowl. Management* 263-270.

Freund, Y., and R. E. Schapire (1999). A Short Introduction to Boosting, *J. Japanese Soc. AI* **14** 771-780.

Geller, R. J. (1997). Earthquake prediction: a critical review, *Geophys. J. Int.* **131** 425-450.

Glez-Peña, D., A. Laurenco, H. Lopez-Fernandez, M. Reboiro-Jato, and F. Fdez-Riverola (2013). Web scraping technologies in an API world, *Briefing in Bioinformatics* **15** 788-797.

Grimmer, J., and B. M. Stewart (2013). Text as Data: The Promise and Pitfalls of Automatic Content Analysis Methods for Political Texts, *Political Analysis* **21** 267-297, doi: 10.1093/pan/mps028.

Hardebeck, J. L., K. R. Felzer, and A. J. Michael (2008). Improved tests reveal that the accelerating moment release hypothesis is statistically insignificant, *J. Geophys. Res.* **113** B08310, doi: 10.1029/2007JB005410.

Hechenbichler, K., and K. P. Schliep (2004). Weighted k-nearest-neighbor techniques and ordinal classification, Discussion paper 399, SFB 386, Ludwig-Maximilians University, Munich.

Hough, S. (2010). *Predicting the Unpredictable: The Tumultuous Science of Earthquake Prediction*, Princeton University Press, 272 pp.

Joachims, T. (1998). Text Categorization with Support Vector Machines: Learning with Many Relevant Features, *Machine Learning ECML-98* 137-142.

Karatzoglou, A., A. Smola, K. Hornik, and A. Zeileis (2004). kernlab - An S4 Package for Kernelt Methods in R, *J. Stat. Software* **11** 1-20.

Kharde, V. A., and S. S. Sonawane (2016). Sentiment Analysis of Twitter Data: A Survey of Techniques, *Int. J. Comput. Appl.* **139** 5-15.

King, G. C. P. (1983). The accommodation of large strains in the upper lithosphere of the earth and other solids by self-similar fault systems: the geometrical origin of b-value, *Pure Appl. Geophys.* **121** 761-815.

Kohavi, R. (1995). A Study of Cross-Validation and Bootstrap for Accuracy Estimation and Model Selection, *IJCAI'95 Proceed. 14th Int. Joint Conf. AI* **2** 1137-1143.

Kuhn, T. (1970). *The Structure of Scientific Revolutions, Enlarged*, International Encyclopedia of Unified Science, 2nd ed., The University of Chicago Press, 210 pp.

Liaw, A., and M. Wiener (2018). Breiman and Cutler's Random Forests for Classification and Regression, Package 'randomForest', available at https://cran.r-project.org/web/packages/randomForest/ (last assessed August 2018).

Mignan, A., G. C. P. King, and D. Bowman (2007). A mathematical formulation of accelerating moment release based on the stress accumulation model, *J. Geophys. Res.* **112** B07308, doi: 10.1029/2006JB004671.

Mignan, A. (2012). Seismicity precursors to large earthquakes unified in a stress accumulation framework, *Geophys. Res. Lett.* **39** L21308, doi: 10.1029/2012GL053946.

Mignan, A. (2011). Retrospective on the Accelerating Seismic Release (ASR) hypothesis: Controversy and new horizons, *Tectonophysics* **505** 1-16, doi: 10.1016/j.tecto.2011.03.010.

Mignan, A. (2014). The debate on the prognostic value of earthquake foreshocks: A meta-analysis, *Sci. Rep.* **4** 4099, doi: 10.1038/srep04099.

Mouselimis, L. (2018). Kernel k Nearest Neighbors, Package 'KernelKnn', available at https://cran.r-project.org/web/packages/KernelKnn/ (last assessed August 2018).

Ng, S.-K., and M. Wong (1999). Toward Routine Automatic Pathway Discovery from On-line Scientific Text Abstracts, *Genome Informatics* **10** 104-112.

Ng, A. Y., and M. I. Jordan (2001). On Discriminative vs. Generative classifiers: A comparison of logistic regression and naive Bayes, *Adv. Neural Inform. Process. Syst.* **14** 605-610.

Ogata, Y. (1988). Statistical Models for Earthquake Occurrences and Residual Analysis for Point Processes, *J. Am. Stat. Assoc.* **83** 9-27.

Papadopoulos, G. A. (1988). Long-term accelerating foreshock activity may indicate the occurrence time of a strong shock in the Western Hellenic Arc, *Tectonophysics* **152** 179-192.

Papazachos, B. C., G. F. Karakaisis, C. B. Papazachos, and E. M. Scordilis (2007). Evaluation of the Results for an Intermediate-Term Prediction of the 8 January 2006 Mw 6.9 Cythera Earthquake in the Southwestern Aegean, *Bull. Seismol. Soc. Am.* **97** 347-352, doi: 10.1785/0120060075.

Pearce, D., and V. Rantala (1983), New Foundations for Metascience, *Synthese* **56** 1-26.

Rokach, L. (2010). Ensemble-based classifiers, *Artif. Intell. Rev.* **33** 1-39, doi: 10.1007/s10462-009-9124-7.

Rumelhart, D. E., G. E. Hinton, and R. J. Williams (1986). Learning representations by back-propagation errors, *Nature* **323** 533-536.

Salton, G., and M. McGill, eds. (1983). *Introduction to Modern Information Retrieval*, McGraw-Hill.

Sammis, C. G., and D. Sornette (2002). Positive feedback, memory, and the predictability of earthquakes, *PNAS* **99** 2501-2508, doi: 10.1073/pnas.012580999.

Sebastiani, F. (2002). Machine Learning in Automated Text Categorization, *ACM Computing Surveys* **34** 1-47.

Seif, S., A. Mignan, J. D. Zechar, M. J. Werner, and S. Wiemer (2017). Estimating ETAS: The effects of truncation, missing data, and model assumptions, *J. Geophys. Res. Solid Earth* **122** 449-469, doi: 10.1002/2016JB012809.

Sornette, D. (2000). *Critical Phenomena in Natural Sciences, Chaos, Fractal, Selforganization and Disorder: Concepts and Tools*, Springer, 434 pp.

Steinwart, I., and A. Christmann (2008). *Support Vector Machines*, Information Science and Statistics, Springer, 601 pp.

Tsytsarau, M., and T. Palpanas (2012). Survey on mining subjective data on the web, *Data Lin. Knowl. Disc.* **24** 478-514, doi: 10.1007/s10618-011-0238-6.

**Tables**

**Table 1.** Updated corpus† for the period 2011-2018.

| Reference | B / M* | Rationale behind the labelling |
|---|---|---|
| Jiang and Wu (2012) | 0 / 1 | Empirical study. Criticality mentioned in the introduction but not assumed nor demonstrated: "*(AMR) model [...] once related to the critical-point-like behavior of earthquake preparation*" |
| Lagios et al. (2012) | 0 / 1 | Empirical study. Many mentions of "*critical area*" and "*critical time*" (in both abstract and main text) but in a generic manner, without any theoretical assertion |
| Mignan (2012) | 0 / 0 | Explicit "*Non-Critical Precursory Accelerating Seismicity Theory*" |
| Pliakis et al. (2012) | 1 / 3 | Critical-point model with no mention of criticality in the abstract nor title. In the main text, however, "*a mounting body of evidence indicates that the earthquake generation process can be viewed as a critical phenomenon that culminates in a large event that corresponds to some critical point*" |
| Bouchon et al. (2013) | 0 / 0 | Empirical study assuming non-criticality: "*Such models [(where foreshocks trigger one another)] are contradicted by investigations of well-recorded foreshock sequences*" and "*one possible mechanism would be the slow slip of a patch of the subducting plate before the earthquake*". No mention of criticality |
| Guilhem et al. (2013) | 0 / 0 | Statistical test of the precursory accelerating seismicity hypothesis, assuming non-criticality ("*models of stress accumulation*"). Note some mentions of criticality within the text and the use of "*criticized*" in the abstract (part of the feature "*critic\**") |
| Jiang and Wu (2013) | 0 / 0 | Test of the critical point hypothesis, with the following non-critical conclusion: "*potential spatial correlation [...] which might be an evidence for that the observed acceleration may have a geometrical or mechanical* |
| | | *rather than statistical origin*". Note the use of "*critical point*" in both abstract and keywords |
| Karakaisis et al. (2013) | 1 / 2 | Empirical study assuming criticality: "*The generation of accelerating preshocks is considered as a critical phenomenon*" |
| De Santis et al. (2015) | 0 / 1 | Empirical study, theoretically agnostic, often using the term "*criticism*" |
| Felzer et al. (2015) | 1 / 2 | Comment to Bouchon et al. (2013) with simulation based on epidemic-type triggering, implicitly related to criticality (bottom-up cascading): "*Acceleration of seismicity [...] has been explained by the cascade model*". Note that the term "critical" is absent throughout the comment |
| Bouchon et al. (2015) | 0 / 0 | Rebuttal to Felzer et al. (2015) reemphasizing the non-criticality of the process: "*Because these foreshocks cluster in time but do not cluster in space, as the ETAS model implicitly assumes, the ETAS model cannot provide a correct description of them*" |
| Christou et al. (2016) | 0 / 1 | Empirical study, theoretically agnostic, often using the expression "*critical region*" |
| Adamaki and Roberts (2017) | 0 / 1 | Empirical study, theoretically agnostic. Only one mention of the term "*near-critical*" |
| Kazemian and Hatami (2017) | 0 / 1 | Empirical study, theoretically agnostic |
| Huang and Meng (2018) | 0 / 0 | Empirical study with non-critical conclusion: "*We interpret [the accelerating rate of repeaters] as the large-scale slow unlocking process*" |

† the metadata and labelling of the full corpus can be found in the supplementary material.

* See text and Figure 1 for class numbering definition.

**Table 2.** One-leave-out cross-validation results for the binary classification.

| Algorithm † | Corpus | Accuracy | F1 score | Precision | Recall |
|---|---|---|---|---|---|
| | *Full text* | *0.63* | *0.75* | *0.61* | *0.98* |
| *'critic*' keyword search* | *Meta-data** | *0.80* | *0.82* | *0.88* | *0.76* |
| Naïve Bayes | Full text | **0.86** | **0.88** | **0.88** | 0.88 |
| | Meta-data | **0.84** | **0.86** | 0.86 | **0.86** |
| kNN | Full text | **0.85** | **0.88** | **0.82** | 0.94 |
| | Meta-data | 0.77 | 0.81 | 0.76 | **0.88** |
| SVM | Full text | **0.78** | **0.81** | **0.82** | 0.80 |
| | Meta-data | 0.77 | 0.80 | 0.79 | **0.82** |
| Random Forest | Full text | **0.81** | **0.85** | **0.80** | 0.90 |
| | Meta-data | 0.74 | 0.81 | 0.72 | **0.92** |

† Results in italics, based on a dictionary, are shown for machine learning benchmarking.

* Meta-data defined as the combination of title + authors + abstract + keywords.

**Table 3.** Ranked posterior probabilities of binary classes $c$ conditional on selected terms $i$ in the Naïve Bayes model.

| Term $i$ | $P(c = 1|i) > 0.9$ | Term $i$ | $P(c = 1|i) < 0.1$ |
|---|---|---|---|
| 'lurr' | 0.993 | 'backslip' | 0.027 |
| 'powerlaw' | 0.987 | 'sumatra-java' | 0.027 |
| 'automat' | 0.984 | 'prestress' | 0.031 |
| 'fibre-bundl' | 0.984 | 'dip-slip' | 0.032 |
| 'lattic' | 0.976 | 'false-posit' | 0.032 |
| 'renormalizationgroup' | 0.975 | 'non-crit' | 0.032 |
| 'acoustic-emis' | 0.972 | 'wenchuan' | 0.336 |
| 'spinod' | 0.972 | 'dmowska' | 0.036 |
| 'ellipt' | 0.971 | 'cff' | 0.039 |
| 'rank-ord' | 0.970 | 'rtl' | 0.041 |

**Table 4.** One-leave-out cross-validation results for the multiclass classification.

| Algorithm | Corpus | Accuracy | F1 score | | | |
|---|---|---|---|---|---|---|
| | | | 0 | 1 | 2 | 3 |
| Naïve Bayes | Full text | 0.78 | 0.80 | 0.72 | 0.78 | 0.84 |
| | Meta-data | 0.66 | 0.74 | 0.50 | 0.68 | 0.73 |

**Table 5.** Selected terms *i* from the 40 highest posterior probabilities *P*(*c*|*i*) for multi-class *c*.

| Class 0 | Class 1 | Class 2 | Class 3 |
|---|---|---|---|
| 'pas' | 'kunlun' | 'ellipt' | 'fibre-bundle' |
| 'cff' | 'wenchuan' | 'lurr' | 'cell' |
| 'c-valu' | 'palermo' | 'ellips' | 'spin' |
| 'pre-stress' | 'chi-squar' | 'l'aquila' | 'accoustic-emiss' |
| 'non-crit' | 'false-posit' | 'aegean' | 'timestep' |
| 'lobe' | 'subcycl' | 'pre-shock' | 'singular' |
| 'sumatra-java' | 'quiescence-lik' | 'chaotic' | 'temperatur' |
| 'backslip' | 'subregion' | 'powerlaw' | 'finite-tim' |
| 'coulomb' | 'log10' | 'syntheticsequ' | 'dissip' |
| 'scenario' | 'bic' | 'cumulativebenioff' | 'quasi-period' |

**Table 6.** Model prediction of the test data for binary classification.

| Algorithm † | Corpus | Accuracy | F1 score | Precision | Recall |
|---|---|---|---|---|---|
| *'critic*'' keyword search* | *Full text* | *0.27* | *0.27* | *0.17* | *0.67* |
|  | *Meta-data* | *0.53* | *-* | *-* | *-* |
| Naïve Bayes | Full text | **0.73** | 0.50 | 0.40 | **0.67** |
|  | Meta-data | **0.73** | **0.60** | 0.43 | **1.00** |

† Results in italics, based on a dictionary, are shown for machine learning benchmarking.

**Table 7.** Model prediction of the test data for multiclass classification.

| Algorithm | Corpus | Accuracy | F1 score | | | |
|---|---|---|---|---|---|---|
| | | | 0 | 1 | 2 | 3 |
| Naïve Bayes | Full text | **0.47** | 0.80 | 0.20 | 0.25 | 1.00 |
|  | Meta-data | **0.53** | 0.67 | 0.40 | 0.44 | 1.00 |

**Supplementary Material:** 1988-2018 precursory accelerating seismicity corpus metadata with matching binary and multiclass labels.

**Table S1 [page 22].** Corpus metadata in JSON format, with the following attributes: "refID", "title", "authors", "abstract", "keywords", "refs" for reference list (kept null for the present study), "journal", "year" and "doi".

**Table S2 [page 90].** List of binary labels in JSON format with same "refID" as in Table S1.

**Table S3 [page 98].** List of multiclass labels in JSON format with same "refID" as in Table S1.

**Table S1.**

```
[
{
    "refID": "1988-1_Papadopoulos-GA_Tectonophys",
    "title": "Long-term accelerating foreshock activity may indicate the occurrence time of a strong shock in the Western Hellenic Arc",
    "authors": ["Papadopoulos-GA"],
    "abstract": "In the decades prior to the occurrence of the 1899 (M, = 6.6) and 1947 (MS = 7.0) main shocks in the Western Hellenic Arc (WHA) the rate of occurrence of foreshocks (M, 2 5.2) within a radius of 100 km around the epicenters can generally be said to have accelerated. After a long period of very low foreshock activity (stage I), the process culminates in a final rapid acceleration of foreshocks some months before the main shocks (stage II), while the last two months are quiescent (stage III). These three stages are in good correspondence with the three stages of crustal deformation and several precursors to the main shock as predicted by the dilatancy model. The data fit very well power law equations similar to those found for short-term foreshocks of ordinary earthquakes and those associated with the creation of artificial lakes. The position of the rupture zones of past WHA strong shocks implies that the eastern part of the segment is the most probable location for the next strong shock in the arc. A process of accelerating seismic activity, similar to that which preceded the 1899 and 1947 shocks, has been under way since 1966 around this part of the segment. A comparison between changes in the power law curves for the earlier earthquakes and the one now expected, indicates that the latter is now 3-8 months "overdue". Assuming that the long-term accelerating foreshock activity is a seismotectonic peculiarity of the WHA segment, I suggest that the preparation of the next rupture in the WHA has entered a highly mature stage, and that there will very probably be an earthquake in this area within the next few months.",
    "keywords": null,
```

```
        "refs": null,
        "journal": "Tectonophysics",
        "year": 1988,
        "doi": null
},
{
        "refID": "1990-1_Sykes-LR_Nature",
        "title": "Seismic activity on neighbouring faults as a long-term precursor to large earthquakes in the San Francisco Bay area",
        "authors": ["Sykes-LR", "Jaume-SC"],
        "abstract": "Activity in moderate-size earthquakes accelerated in the several decades before the large California earthquakes of 1989, 1906 and 1868. This type of precursor seems to require the presence of several major faults in close enough proximity to one another that moderate-size shocks are selectively triggered on surrounding faults during the latter stages of the cycle of strain buildup to large earthquakes. It may be possible to use quantitative aspects of similar seismic precursors to make predictions of large earthquakes on timescales of a few years to one decade.",
        "keywords": null,
        "refs": null,
        "journal": "Nature",
        "year": 1990,
        "doi": null
},
{
        "refID": "1992-1_Jaume-SC_GRL",
        "title": "Accelerating seismic moment release and outer-rise compression: possible precursors to the next great earthquake in the Alaska Peninsula Region",
        "authors": ["Jaume-SC", "Estabrook-CH"],
        "abstract": "The moment release rate in the Kodiak Island (KI) segment increased prior to the great 1964 Prince William Sound earthquake (MW = 9.2). Starting in 1983 the moment release rate in the Shumagin Island (SI) segment shows a similar increase. In July 1990, an outer-rise reverse earthquake showed arc-normal compression at a depth of 42 km seaward of the Alaska Peninsula (AP) segment. Because accelerating moment release and arc-normal compression have both been proposed as precursors to large and great earthquakes, we suggest this is evidence that the SI-AP region is nearing the end of the seismic cycle between large thrust earthquakes.",
        "keywords": null,
        "refs": null,
        "journal": "Geophysical Research Letters",
        "year": 1992,
        "doi": "10.1029/92GL00260"
},
{
        "refID": "1993-1_Bufe-CG_JGR",
```

```
        "title": "Predictive Modeling of the Seismic Cycle of the Greater San Francisco Bay Region",
        "authors": ["Bufe-CG", "Varnes-DJ"],
        "abstract": "The seismic cycle for the San Francisco Bay region is synthesized by a model combining the pre-and post-1906 seismic histories. The long-term acceleration of seismic release (seismic moment, Benioff strain release, or event count) in the seismic cycle and the shorter-term accelerations preceding the larger earthquakes within that cycle are modeled using an empirical predictive technique, called time-to-failure analysis, in which rate of seismic release is proportional to an inverse power of the remaining time to failure. The exponent of time to failure in the accelerating sequences appears to be scale invariant, and the length of the full cycle is estimated at 269 ± 50 years. The 1989 Loma Prieta earthquake, which is the culmination of the first subcycle in the present long-term seismic cycle, should have been predictable with an uncertainty of 2 years in time and 0.5 in magnitude, although the specific location (at Loma Prieta) was not predictable by this technique. If our model is correct and if the Loma Prieta earthquake is the culmination of a subcycle, the San Francisco Bay region should be entering a relatively long (20—50 years) period of seismic quiescence above magnitude 6. A great earthquake, such as the 1906 San Francisco event, would appear to be more than a century in the future.",
        "keywords": null,
        "refs": null,
        "journal": "Journal of Geophysical Research",
        "year": 1993,
        "doi": "10.1029/93JB00357"
},
{
        "refID": "1994-1_Bufe-CG_PAGEOPH",
        "title": "Seismicity Trends and Potential for Large Earthquakes in the Alaska-Aleutian Region",
        "authors": ["Bufe-CG", "Nishenko-SP", "Varnes-DJ"],
        "abstract": "The high likelihood of a gap-filling thrust earthquake in the Alaska subduction zone within this decade is indicated by two independent methods: analysis of historic earthquake recurrence data and time-to-failure analysis applied to recent decades of instrumental data. Recent (May 1993) earthquake activity in the Shumagin Islands gap is consistent with previous projections of increases in seismic release, indicating that this segment, along with the Alaska Peninsula segment, is approaching failure. Based on this pattern of accelerating seismic release, we project the occurrence of one or moreM≥7.3 earthquakes in the Shumagin-Alaska Peninsula region during 1994—1996. Different segments of the Alaska-Aleutian seismic zone behave differently in the decade or two preceding great earthquakes, some showing acceleration of seismic release (type "A" zones), while others show deceleration (type "D" zones). The largest Alaska-Aleutian earthquakes—in 1957, 1964,
```

```
        and 1965—originated in zones that exhibit type D behavior. Type
        A zones currently showing accelerating release are the Shumagin,
        Alaska Peninsula, Delarof, and Kommandorski segments. Time-to-
        failure analysis suggests that the large earthquakes could occur
        in these latter zones within the next few years.",
            "keywords": ["Alaska-Aleutian seismic zone", "Shumagin
        seismic gap", "accelerating moment release", "time-to-
        failure"],
            "refs": null,
            "journal": "Pure and Applied Geophysics",
            "year": 1994,
            "doi": null
    },
    {
            "refID": "1995-1_Newman-WI_PRE",
            "title": "Log-periodic behavior of a hierarchical failure
        model with applications to precursory seismic activation",
            "authors": ["Newman-WI", "Turcotte-DL", "Gabrielov-AM"],
            "abstract": "Seismic activation has been recognized to
        occur before many major earthquakes including the San Francisco
        Bay area, prior to the 1906 earthquake. There is a serious
        concern that the recent series of earthquakes in Southern
        California is seismic activation prior to a great Southern
        California earthquake. The seismic activation prior to the Loma
        Prieta earthquake has been quantified in terms of a power-law
        increase in the regional Benioff strain release prior to this
        event and there is an excellent fit to a log-periodic increase
        in the Benioff strain release. In order to better understand
        activation a hierarchical seismic failure model has been
        studied. An array of stress-carrying elements is considered
        (formally, a cellular automaton or lattice gas, but analogous
        to the strands of an ideal, frictionless cable). Each element
        has a time to failure that is dependent on the stress the element
        carries and has a statistical distribution of values. When an
        element fails, the stress on the element is transferred to a
        neighboring element; if two adjacent elements fail, stress is
        transferred to two neighboring elements; if four elements fail,
        stress is transferred to four adjacent elemetns, and so forth.
        When stress is transferred to an element its time to failure is
        reduced. The intermediate size failure events prior to total
        failure each have a sequence of precursory failures, and these
        precursory failures each have an embedded precursory sequence
        of smaller failures. The total failure of the array appears to
        be a critical point. There is a sequence of partial failures
        leading up to the total failure that resembles a log-periodic
        sequence.",
            "keywords": null,
            "refs": null,
            "journal": "Physical Review E",
            "year": 1995,
            "doi": "10.1103/PhysRevE.52.4827"
    },
```

```
{
    "refID": "1995-1_Sornette-D_JPhysIFrance",
    "title": "Complex Critical Exponents from Renormalization Group Theory of Earthquakes: Implications for Earthquake Predictions",
    "authors": ["Sornette-D", "Sammis-CG"],
    "abstract": "Several authors have proposed discrete renormalization group models of earthquakes, viewing them as a kind of dynamical critical phenomena. Here, we propose that the assumed discrete scale invariance stems from the irreversible and intermittent nature of rupture which ensures a breakdown of translational invariance. As a consequence, we show that the renormalization group entails complex critical exponents, describing log-periodic corrections to the leading scaling behavior. We use the mathematical form of this solution to fit the time to failure dependence of the Benioff strain on the approach of large earthquakes. This might provide a new technique for earthquake prediction for which we present preliminary tests on the 1989 Loma Prieta earthquake in northern California and on a recent build-up of seismic activity on a segment of the Aleutian-Island seismic zone. The earthquake phenomenology of precursory phenomena such as the causal sequence of quiescence and foreshocks is captured by the general structure of the mathematical solution of the renormalization group.",
    "keywords": null,
    "refs": null,
    "journal": "Journal de Physique I",
    "year": 1995,
    "doi": null
},
{
    "refID": "1996-1_Jaume-SC_JGR",
    "title": "Evolution of moderate seismicity in the San Francisco Bay region, 1850 to 1993: Seismicity changes related to the occurrence of large and great earthquakes",
    "authors": ["Jaume-SC", "Sykes-LR"],
    "abstract": "The rate of seismic activity of moderate-size (M > 5.5) earthquakes in the San Francisco Bay (SFB) region has varied considerably during the past 150 years. As measured by the rate of seismic moment release, seismic activity in the SFB region is observed to accelerate prior to M > 7.0 earthquakes in 1868, 1906, and 1989, and then to decelerate following them. We examine these seismicity changes in the context of the evolution of the stress field in the SFB region as a result of strain accumulation and release using a model of dislocations in an elastic halfspace. We use a Coulomb failure function (CFF) to take into account changes in both shear and normal stresses on potential failure planes of varying strike and dip in the SFB region. We find that the occurrence of a large or great earthquake creates a "stress shadow": a region where the stress driving earthquake deformation is decreased. Interseismic strain
```

accumulation acts to reverse this process, gradually bringing faults in the SFB region out of the stress shadow of a previous large or great earthquake and back into a state where earthquake failure is possible. As the stress shadow generated by a large or great earthquake disappears, it migrates inward toward the fault associated with that large or great event. The observed changes in the rate of occurrence of moderate earthquakes in the SFB region are broadly consistent with this model. In detail, the decrease in seismicity throughout most of the SFB region and a localized increase in the Monterey Bay region following the great 1906 earthquake is consistent with our predicted stress changes. The timing and location of moderate-size earthquakes when the rate of seismicity increases again in the 1950s is consistent with areas in which the 1906 stress shadow had been eliminated by strain accumulation in the SFB region. Those earthquakes that are most inconsistent with our stress evolution model, including the 1911 earthquake southeast of San Jose, are found to occur in regions where dip-slip faulting is common in addition to strike-slip. The 1906 earthquake brought that zone of dip-slip faulting closer to failure, suggesting that the 1911 event may have been a reverse faulting earthquake rather than a strike-slip one similar to the 1984 Morgan Hill earthquake. The occurrence of activity on faults very close to the San Andreas, such as the Lake Elsman earthquakes of 1988 and 1989, appear to be associated with the last disappearence of the stress shadow on the Loma Prieta segment of the San Andreas fault. Thus events of that type may represent an intermediate-term precursor to a large earthquake, such as the 1989 Loma Prieta event. Much of the moderate-size earthquake activity in the SFB region appears to be modulated in time by the buildup and release of stress in large and great earthquakes. A tensorial approach to earthquake prediction, i.e., taking into account changes in the components of the stress tensor, has several advantages over examining scalar changes such as those in seismic activity and moment release rates. This tensorial approach allows for both activation and quiescence (but in different subregions) prior to as well as after large earthquakes.",
        "keywords": null,
        "refs": null,
        "journal": "Journal of Geophysical Research",
        "year": 1996,
        "doi": "10.1029/95JB02393"
    },
    {
        "refID": "1996-1_Saleur-H_JGR",
        "title": "Discrete scale invariance, complex fractal dimensions, and log-periodic fluctuations in seismicity",
        "authors": ["Saleur-H", "Sammis-CG", "Sornette-D"],
        "abstract": "We discuss in detail the concept of discrete scale invariance and show how it leads to complex critical exponents and hence to the log-periodic corrections to scaling exhibited by various measures of seismic activity close to a

large earthquake singularity. Discrete scale invariance is first illustrated on a geometrical fractal, the Sierpinsky gasket, which is shown to be fully described by a complex fractal dimension whose imaginary part is a simple function (inverse of the logarithm) of the discrete scaling factor. Then, a set of simple physical systems (spins and percolation) on hierarchical lattices is analyzed to exemplify the origin of the different terms in the discrete renormalization group formalism introduced to tackle this problem. As a more specific example of rupture relevant for earthquakes, we propose a solution of the hierarchical time-dependent fiber bundle of Newman et al. [1994] which exhibits explicitly a discrete renormalization group from which log-periodic corrections follow. We end by pointing out that discrete scale invariance does not necessarily require an underlying geometrical hierarchical structure. A hierarchy may appear "spontaneously" from the physics and/or the dynamics in a Euclidean (nonhierarchical) heterogeneous system. We briefly discuss a simple dynamical model of such mechanism, in terms of a random walk (or diffusion) of the seismic energy in a random heterogeneous system.",
        "keywords": null,
        "refs": null,
        "journal": "Journal of Geophysical Research",
        "year": 1996,
        "doi": "10.1029/96JB00876"
},
{
        "refID": "1996-1_Varnes-DJ_GJI",
        "title": "The cyclic and fractal seismic series preceding an mb 4.8 earthquake on 1980 February 14 near the Virgin Islands",
        "authors": ["Varnes-DJ", "Bufe-CG"],
        "abstract": "Seismic activity in the 10 months preceding the 1980 February 14, mb 4.8 earthquake in the Virgin Islands, reported on by Frankel in 1982, consisted of four principal cycles. Each cycle began with a relatively large event or series of closely spaced events, and the duration of the cycles progressively shortened by a factor of about 3/4. Had this regular shortening of the cycles been recognized prior to the earthquake, the time of the next episode of seismicity (the main shock) might have been closely estimated 41 days in advance. That this event could be much larger than the previous events is indicated from time-to-failure analysis of the accelerating rise in released seismic energy, using a non-linear time- and slip-predictable foreshock model. Examination of the timing of all events in the sequence shows an even higher degree of order. Rates of seismicity, measured by consecutive interevent times, when plotted on an iteration diagram of a rate versus the succeeding rate, form a triangular circulating trajectory. The trajectory becomes an ascending helix if extended in a third dimension, time. This construction reveals additional and precise relations among the time intervals between times of

relatively high or relatively low rates of seismic activity, including period halving and doubling. The set of 666 time intervals between all possible pairs of the 37 recorded events appears to be a fractal; the set of time points that define the intervals has a finite, non-integer correlation dimension of 0.70. In contrast, the average correlation dimension of 50 random sequences of 37 events is significantly higher, close to 1.0. In a similar analysis, the set of distances between pairs of epicentres has a fractal correlation dimension of 1.52. Well-defined cycles, numerous precise ratios among time intervals, and a non-random temporal fractal dimension suggest that the seismic series is not a random process, but rather the product of a deterministic dynamic system.",
        "keywords": ["earthquake prediction", "fractals", "seismicity", "Virgin Islands"],
        "refs": null,
        "journal": "Geophysical Journal International",
        "year": 1996,
        "doi": "10.1111/j.1365-246X.1996.tb06359.x"
    },
    {
        "refID": "1998-1_Bowman-DD_JGR",
        "title": "An observational test of the critical earthquake concept",
        "authors": ["Bowman-DD", "Ouillon-G", "Sammis-CG", "Sornette-A", "Sornette-D"],
        "abstract": "We test the concept that seismicity prior to a large earthquake can be understood in terms of the statistical physics of a critical phase transition. In this model, the cumulative seismic strain release increases as a power law time to failure before the final event. Furthermore, the region of correlated seismicity predicted by this model is much greater than would be predicted from simple elastodynamic interactions. We present a systematic procedure to test for the accelerating seismicity predicted by the critical point model and to identify the region approaching criticality, based on a comparison between the observed cumulative energy (Benioff strain) release and the power law behavior predicted by theory. This method is used to find the critical region before all earthquakes along the San Andreas system since 1950 with M≥6.5. The statistical significance of our results is assessed by performing the same procedure on a large number of randomly generated synthetic catalogs. The null hypothesis, that the observed acceleration in all these earthquakes could result from spurious patterns generated by our procedure in purely random catalogs, is rejected with 99.5% confidence. An empirical relation between the logarithm of the critical region radius (R) and the magnitude of the final event (M) is found, such that log R∝0.5M, suggesting that the largest probable event in a given region scales with the size of the regional fault network.",
        "keywords": null,
        "refs": null,

```
        "journal": "Journal of Geophysical Research",
        "year": 1998,
        "doi": "10.1029/98JB00792"
},
{
        "refID": "1998-1_Brehm-DJ_BSSA",
        "title": "Intermediate-Term Earthquake Prediction Using Precursory Events in the New Madrid Seismic Zone",
        "authors": ["Brehm-DJ", "Braile-LW"],
        "abstract": "Earthquake prediction may be possible for some mainshock events. The time-to-failure method described by Varnes (1989) and Bufe and Varnes (1990) uses precursory events (foreshocks) to define an accelerated energy release curve. By fitting an equation to the data, a predicted time of failure and magnitude can be calculated. Until recently, this method has been used in only a few studies in tectonically active areas, and for moderate- to large-magnitude mainshock events. Using the microearthquake network data set from the New Madrid Seismic Zone (NMSZ), which is reasonably complete for earthquakes of magnitude ≧1.5 in the area of interest, the method has yielded predicted values of past events as small as mb = 3.5. The network data set used in this evaluation covers the time interval from 29 June 1974 to 20 July 1995 for the NMSZ. There have been 36 earthquakes of magnitude ≧3.5 over the 21-yr period in which the network has been operating. Because precursory events are required for the application of the time-to-failure method, mainshocks that occurred before 1980 do not have enough data to adequately define the accelerated energy release curve. Therefore, we utilized the 26 earthquakes that occurred after 1980 and that had a magnitude ≧3.5. Sixteen of the 26 mainshock events were modeled. In most cases, the precursory sequences yielded predicted times of failure and magnitudes that were reasonably close to the actual mainshock values. The remaining mainshocks, including those occurring before 1980, could not be modeled due to either (1) not enough events to adequately define the precursory sequence or (2) interfering events that disrupt the accelerated energy release curve. In addition, two events were modeled from the Nuttli catalog (Nuttli, 1979) along with one that used a combination of both catalogs. Nineteen earthquakes with magnitudes ≧3.5 were evaluated using the time-to-failure method. The first calculation using the time-to-failure method gave predicted results with large error bounds, essentially no upper bound on the predicted magnitude. An empirical relationship between parameters has helped to constrain the range of the predicted magnitude and, to a lesser extent, the estimated time of failure. This relationship modifies the time-to-failure equation and yields predicted values for magnitudes that have an upper limit. Another empirical relationship suggests that the logarithm of the moment of the mainshock increases linearly with the logarithm of the size of the precursory event search diameter. The relative seismicity of the region also influences the optimum search
```

```
        diameter used to find precursory events. In addition to the
        evaluation of the acceleration sequences associated with the
        mainshocks, an analysis of the occurrence of false-positive
        acceleration sequences (acceleration sequences that do not end
        in a mainshock) was conducted. The preliminary false-positive
        analysis was conducted by randomly selecting potential mainshock
        locations. The results yielded a false-positive acceleration
        sequence occurrence rate of 2%. With the incorporation of the
        empirical relationships into the time-to-failure method, the
        potential for future intermediate-term earthquake predictions
        for the NMSZ is encouraging.",
        "keywords": null,
        "refs": null,
        "journal": "Bulletin of the Seismological Society of
America",
        "year": 1998,
        "doi": null
},
{
        "refID": "1998-1_Gross-S_GJI",
        "title": "A systematic test of time-to-failure analysis",
        "authors": ["Gross-S", "Rundle-JB"],
        "abstract": "Time-to-failure analysis is a technique for
predicting earthquakes in which a failure function is fit to a
time-series of accumulated Benioff strain. Benioff strain is
computed from regional seismicity in areas that may produce a
large earthquake. We have tested the technique by fitting two
functions, a power law proposed by Bufe & Varnes (1993) and a
log-periodic function proposed by Sornette & Sammis (1995). We
compared predictions from the two time-to-failure models to
observed activity and to predicted levels of activity based upon
the Poisson model. Likelihood ratios show that the most
successful model is Poisson, with the simple Poisson model four
times as likely to be correct as the best time-to-failure model.
The best time-failure model is a blend of 90 per cent Poisson
and 10 per cent log-periodic predictions. We tested the accuracy
of the error estimates produced by the standard least-squares
fitter and found greater accuracy for fits of the simple power
law than for fits of the more complicated log-periodic function.
The least-squares fitter underestimates the true error in time-
to-failure functions because the error estimates are based upon
linearized versions of the functions being fitted.",
        "keywords":    ["earthquake   prediction",   "fitting",
"seismicity", "statistics"],
        "refs": null,
        "journal": "Geophysical Journal International",
        "year": 1998,
        "doi": null
},
{
        "refID": "1998-1_Huang-Y_EPL",
```

```
        "title": "Precursors, aftershocks, criticality and self-organized criticality",
        "authors": ["Huang-Y", "Saleur-H", "Sammis-CG", "Sornette-D"],
        "abstract": "We present a simple model of earthquakes on a pre-existing hierarchical fault structure. The system self-organizes at large times in a stationary state with a power law Gutenberg-Richter distribution of earthquake sizes. The largest fault carries irregular great earthquakes preceded by precursors developing over long time scales and followed by aftershocks obeying an Omori's law. The cumulative energy released by precursors follows a time-to-failure power law with log-periodic structures, qualifying a large event as an effective dynamical (depinning) critical point. Down the hierarchy, smaller earthquakes exhibit the same phenomenology, albeit with increasing irregularities.",
        "keywords": null,
        "refs": null,
        "journal": "Europhysics Letters",
        "year": 1998,
        "doi": null
    },
    {
        "refID": "1999-1_Brehm-DJ_BSSA",
        "title": "Intermediate-Term Earthquake Prediction Using the Modified Time-to-Failure Method in Southern California",
        "authors": ["Brehm-DJ", "Braile-LW"],
        "abstract": "Based on retrospective modeling of earthquakes from the southern California earthquake catalog, along with previously published evaluations from the New Madrid Seismic Zone, the modified time-to-failure method may be used as an intermediate-term earthquake prediction technique for locating and predicting the size and time of a future mainshock. Modeling previous mainshocks for hypothesis development indicates that the method predicts the actual magnitude of the mainshock to within approximately ±0.5 magnitude units. The error associated with the time-of-failure is approximately ±1.1 years assuming the last precursory event is known. When the last event in the precursory sequence is not known, the predicted magnitude remains similar, but the predicted time will require refinement as additional events are added, with time, to the sequence. The mainshock location can also be identified within a circular region with a radius on the order of tens of kilometers. Criteria are provided for defining acceleration sequences and mainshock locations. The criteria reduce the number of false predictions but also eliminate some mainshocks from our evaluation. Mainshocks as small as magnitude 5.5, occurring between 1980 and 1995, were evaluated from the Southern California earthquake Catalog (SCC). The results were used in association with previous studies to develop a method that can be used for practical (future prediction) applications. The modified time-to-failure method was used to search the SCC for future
```

```json
        mainshocks occurring after 17 August 1998. One region satisfied
        all the criteria and may be modeled by the modified time-to-
        failure method. The region likely to have a mainshock is a 65-
        km-radius area centered at 31.43° N, 115.47° W (northern Baja
        California, Mexico). The predicted magnitude is 6.36, ±0.55, and
        the predicted time of failure is 1998.565 (7/25/98), ±1.127
        years. The addition of future precursory events will allow
        refinement of the predicted values.",
        "keywords": null,
        "refs": null,
        "journal": "Bulletin of the Seismological Society of
America",
        "year": 1999,
        "doi": null
},
{
        "refID": "1999-1_Brehm-DJ_JOSE",
        "title": "Refinement of the modified time-to-failure method
for intermediate-term earthquake prediction",
        "authors": ["Brehm-DJ", "Braile-LW"],
        "abstract": "The modified time-to-failure method for
intermediate-term earthquake prediction utilizes empirical
relationships to reduce the number of unknown parameters
providing a stable and unique solution set. The only unknown
parameters in the modified time-to-failure method are the time
and size of the impending main shock. The modified time-to-
failure equation is used to model the precursory events and a
prediction contour diagram is constructed with the magnitude and
time-of-failure as the axes of the diagram. The root-mean-square
(rms) is calculated for each set of time and magnitude on the
prediction diagram representing the difference between the model
(calculated) acceleration and the actual accelerated energy
release of the precursory events. A small region, corresponding
to the low rms region on the diagram, defines the prediction.
The prediction has been shown to consistently under-estimate the
magnitude and over-estimate the time-of-failure. These
shortcomings are caused by an underestimation in energy release
of the modified time-to-failure equation at the very end of the
sequence. An empirical correction can be applied to the
predicted results to minimize this problem. A main shock
location search technique has been developed for use with the
modified time-to-failure method. The location technique is used
to systematically search an earthquake catalog and identify
locations corresponding to precursory sequences that display
accelerated energy releases. It has shown good results when
applied in 'retrospective predictions', and is essential for the
practical application of the modified time-to-failure method.
In addition, an observed linear characteristic in long-term
energy release can be used to minimize false predictions. The
refined empirical relationships that eliminate or constrain
unknown constants used in the modified time-to-failure method
and the main shock location search technique are used in a
```

```
        practical application in the New Madrid Seismic Zone (NMSZ). The
        NMSZ, which is 'over due' for a magnitude 6 event according to
        recurrence rates (Johnston and Nava, 1985), makes this region
        ideal for testing the method. One location was identified in the
        NMSZ as a 'high risk' area for an event in the magnitude 4.5
        range. The prediction, if accurate, is of scientific interest
        only because of the relatively small size of the main shock.",
        "keywords": ["Energy Release", "Structural Geology", "Main
Shock", "Empirical Relationship", "Seismic Zone"],
        "refs": null,
        "journal": "Journal of Seismology",
        "year": 1999,
        "doi": null
},
{
        "refID": "1999-1_Jaume-SC_PAGEOPH",
        "title": "Evolving Towards a Critical Point: A Review of
Accelerating Seismic Moment/Energy Release Prior to Large and
Great Earthquakes",
        "authors": ["Jaume-SC", "Sykes-LR"],
        "abstract": "There is growing evidence that some proportion
of large and great earthquakes are preceded by a period of
accelerating seismic activity of moderate-sized earthquakes.
These moderate earthquakes occur during the years to decades
prior to the occurrence of the large or great event and over a
region larger than its rupture zone. The size of the region in
which these moderate earthquakes occur scales with the size of
the ensuing mainshock, at least in continental regions. A number
of numerical simulation studies of faults and fault systems also
exhibit similar behavior. The combined observational and
simulation evidence suggests that the period of increased moment
release in moderate earthquakes signals the establishment of
long wavelength correlations in the regional stress field. The
central hypothesis in the critical point model for regional
seismicity is that it is only during these time periods that a
region of the earth's crust is truly in or near a "self-organized
critical" (SOC) state, such that small earthquakes are capable
of cascading into much larger events. The occurrence of a large
or great earthquake appears to dissipate a sufficient proportion
of the accumulated regional strain to destroy these long
wavelength stress correlations and bring the region out of a SOC
state. Continued tectonic strain accumulation and stress
transfer during smaller earthquakes eventually re-establishes
the long wavelength stress correlations that allow for the
occurrence of larger events. These increases in activity occur
over longer periods and larger regions than quiescence, which
is usually observed within the rupture zone of a coming large
event. The two phenomena appear to have different physical bases
and are not incompatible with one another.",
        "keywords":    ["Accelerating     seismic    moment/energy",
"earthquake forecasting", "critical point hypothesis", "self-
organized criticality", "stress correlations"],
```

```
        "refs": null,
        "journal": "Pure and Applied Geophysics",
        "year": 1999,
        "doi": null
},
{
        "refID": "1999-1_Main-IG_GJI",
        "title": "Applicability of time-to-failure analysis to accelerated strain before earthquakes and volcanic eruptions",
        "authors": ["Main-IG"],
        "abstract": "We examine quantitatively the ranges of applicability of the equation Ω=A + B[1−t/tf]m for predicting 'system-sized' failure times tf in the Earth. In applications Ω is a proxy measure for strain or crack length, and A, B and the index m are model parameters determined by curve fitting. We consider constitutive rules derived from (a) Charles' law for subcritical crack growth; (b) Voight's equation; and (c) a simple percolation model, and show in each case that this equation holds only when m < 0. When m > 0, the general solution takes the form Ω=A+B[1+t/T]m, where T is a positive time constant, and no failure time can be defined. Reported values for volcanic precursors based on rate data are found to be within the range of applicability of time-to-failure analysis (m < 0). The same applies to seismic moment release before earthquakes, at the expense of poor retrospective predictability of the time of the a posteriori-defined main shock. In contrast, reported values based on increasing cumulative Benioff strain occur in the region where a system-sized failure time cannot be defined (m > 0; commonly m ≈ 0.3). We conclude on physical grounds that cumulative seismic moment is preferred as the most direct measure of seismic strain. If cumulative Benioff strain is to be retained on empirical grounds, then it is important that these data either be re-examined with the independent constraint m < 0, or that for the case 0 <m + < 1, a specific correction for the time-integration of cumulative data be applied, of the form ∑Ω=At+B'{1−[1−t/tf]m + 1}.",
        "keywords": null,
        "refs": null,
        "journal": "Geophysical Journal International",
        "year": 1999,
        "doi": "10.1046/j.1365-246x.1999.00004.x"
},
{
        "refID": "1999-1_Sammis-CG_PAGEOPH",
        "title": "Seismic Cycles and the Evolution of Stress Correlation in Cellular Automaton Models of Finite Fault Networks",
        "authors": ["Sammis-CG", "Smith-SW"],
        "abstract": "A cellular automaton is used to study thè relation between the structure of a regional fault network and the temporal and spatial patterns of regional seismicity. Automata in which the cell sizes form discrete fractal
```

hierarchies are compared with those having a uniform cell size. Conservative models in which all the stress is transferred at each step of a cascade are compared with nonconservative ("lossy") models in which a specified fraction of the stress energy is lost from each step. Particular attention is given to the behavior of the system as it is driven toward the critical state by uniform external loading. All automata exhibit a scaling region at times close to the critical state in which the events become larger and energy release increases as a power-law of the time to the critical state. For the hierarchical fractal automata, this power-law behavior is often modulated by fluctuations that are periodic in the logarithm of the time to criticality. These fluctuations are enhanced in the nonconservative models, but are not robust. The degree to which they develop appears to depend on the particular distribution of stresses in the larger cells which varies from cycle to cycle. Once the critical state is reached, seismicity in the uniform conservative automaton remains random in time, space, and magnitude. Large events do not significantly perturb the stress distribution in the system. However, large events in the nonconservative uniform automaton and in the fractal systems produce large stress perturbations that move the system out of the critical state. The result is a seismic cycle in which a large event is followed by a shadow period of quiescence and then a new approach back toward the critical state. This seismic cycle does not depend on the fractal structure, but is a direct consequence of large-scale heterogeneity of these systems in which the size of the largest cell (or the size of the largest nonconservative event) is a significant fraction of the size of the network. In essence, seismic cycles in these models are boundary effects. The largest events tend to cluster in time and the rate of small events remains relatively constant throughout a cycle in agreement with observed seismicity.",
        "keywords": ["Regional seismicity", "seismic cycle", "cellular automaton", "critical point", "fractals"],
        "refs": null,
        "journal": "Pure and Applied Geophysics",
        "year": 1999,
        "doi": null
},
{
        "refID": "1999-1_Yang-WZ_ActaSeismSinica",
        "title": "Seismicity acceleration model and its application to several earthquake regions in China",
        "authors": ["Yang-WZ", "Ma-L"],
        "abstract": "With the theory of subcritical crack growth, we can deduce the fundamental equation of regional seismicity acceleration model. Applying this model to intraplate earthquake regions, we select three earthquake subplates: North China Subplate, Chuan-Dian Block and Xinjiang Subplate, and divide the three subplates into seven researched regions by the difference of seismicity and tectonic conditions. With the modified

```
    equation given by Sornette and Sammis (1995), we analysis the
    seismicity of each region. To those strong earthquakes already
    occurred in these region, the model can give close fitting of
    magnitude and occurrence time, and the result in this article
    indicates that the seismicity acceleration model can also be
    used for describing the seismicity of intraplate. In the
    article, we give the magnitude and occurrence time of possible
    strong earthquakes in Shanxi, Ordos, Bole-Tuokexun, Ayinke-Wuqia
    earthquake regions. In the same subplate or block, the
    earthquake periods for each earthquake region are similar in
    time interval. The constant α in model can be used to describe
    the intensity of regional seismicity, and for the Chinese
    Mainland, α is 0.4 generally. To the seismicity in Taiwan and
    other regions with complex tectonic conditions, the model does
    not fit well at present.",
        "keywords": ["seismicity acceleration model", "subcritical
crack growth", "China", "earthquake region", "fit"],
        "refs": null,
        "journal": "Acta Seismologica Sinica",
        "year": 1999,
        "doi": null
    },
    {
        "refID": "2000-1_Huang-Y_JGR",
        "title": "Reexamination of log periodicity observed in the
seismic precursors of the 1989 Loma Prieta earthquake",
        "authors": ["Huang-Y", "Saleur-H", "Sornette-D"],
        "abstract": "Based on several empirical evidence, a series
of papers has advocated the concept that seismicity prior to a
large earthquake can be understood in terms of the statistical
physics of a critical phase transition. In this model, the
cumulative seismic Benioff strain release ∈ increases as a power
law time-to-failure before the final event. This power law
reflects a kind of scale invariance with respect to the distance
to the critical point: ∈ is the same up to a simple reseating
λz after the time-to-failure has been scaled by a factor λ. A
few years ago, on the basis of a fit of the cumulative Benioff
strain released prior to the 1989 Loma Prieta earthquake,
Sornette and Sammis [1995] proposed that this scale invariance
could be partially broken into a discrete scale invariance,
defined such that the scale invariance occurs only with respect
to specific integer powers of a fundamental scale ratio. The
observable consequence of discrete scale invariance takes the
form of log-periodic oscillations decorating the accelerating
power law. They found that the quality of the fit and the
predicted time of the event are significantly improved by the
introduction of log periodicity. Here we present a battery of
synthetic tests performed to quantify the statistical
significance of this claim. We put special attention to the
definition of synthetic tests that are as much as possible
identical to the real time series except for the property to be
tested, namely, log periodicity. Without this precaution, we
```

would conclude that the existence of log periodicity in the Loma Prieta cumulative Benioff strain is highly statistically significant. In contrast, we find that log-periodic oscillations with frequency and regularity similar to those of the Loma Prieta case are very likely to be generated by the interplay of the low-pass filtering step due to the construction of cumulative functions together with the approximate power law acceleration. Thus the single Loma Prieta case alone cannot support the initial claim, and additional cases and further study are needed to increase the signal-to-noise ratio, if any. The present study will be a useful methodological benchmark for future testing of additional events when the methodology and data to construct reliable Benioff strain function become available.",
        "keywords": null,
        "refs": null,
        "journal": "Journal of Geophysical Research",
        "year": 2000,
        "doi": "10.1029/2000JB900308"
    },
    {
        "refID": "2000-1_Jaume-SC_GeophysMonogrSer",
        "title": "Changes in earthquake size-frequency distributions underlying accelerating seismic moment/energy release",
        "authors": ["Jaume-SC"],
        "abstract": "A considerable number of moderate to great earthquakes have been preceded by an increase in the rate of smaller events in the surrounding region, resulting in an acceleration in the rate of seismic energy/moment release as the time of the mainshock is approached. The so-called 'critical point' models for this behavior postulate that a growing correlation length in the earth's crust underlies this phenomenon. In these models, the correlation length controls the maximum size of events in the earthquake population. As it grows it allows progressively larger earthquakes to occur, and thus changes the earthquake size-frequency distribution at large magnitudes. Here I test this hypothesis by examining changes in the earthquake size-frequency distribution of 17 known cases of accelerating seismic energy/moment release for which the space-time dimensions of this behavior have been defined and adequate earthquake catalogs exist. I find that for 15 of these 17 cases, observed changes in the earthquake size-frequency distribution are consistent with the predictions of the critical point hypothesis. For the other two cases, an increase in the rate of seismicity at all magnitudes appears to have occurred. These results suggest that critical point behavior underlies most but not all cases of accelerating seismic energy/moment release.",
        "keywords": null,
        "refs": null,
        "journal": "Geophysical Monograph Series",
        "year": 2000,
        "doi": "10.1029/GM120p0199"

```
    },
    {
        "refID": "2000-1_Jaume-SC_PAGEOPH",
        "title": "Accelerating Seismic Energy Release and Evolution of Event Time and Size Statistics: Results from Two Heterogeneous Cellular Automaton Models",
        "authors": ["Jaume-SC", "Weatherley-D", "Mora-P"],
        "abstract": "The evolution of event time and size statistics in two heterogeneous cellular automaton models of earthquake behavior are studied and compared to the evolution of these quantities during observed periods of accelerating seismic energy release prior to large earthquakes. The two automata have different nearest neighbor laws, one of which produces self-organized critical (SOC) behavior (PSD model) and the other which produces quasi-periodic large events (crack model). In the PSD model periods of accelerating energy release before large events are rare. In the crack model, many large events are preceded by periods of accelerating energy release. When compared to randomized event catalogs, accelerating energy release before large events occurs more often than random in the crack model but less often than random in the PSD model; it is easier to tell the crack and PSD model results apart from each other than to tell either model apart from a random catalog. The evolution of event sizes during the accelerating energy release sequences in all models is compared to that of observed sequences. The accelerating energy release sequences in the crack model consist of an increase in the rate of events of all sizes, consistent with observations from a small number of natural cases, however inconsistent with a larger number of cases in which there is an increase in the rate of only moderate-sized events. On average, no increase in the rate of events of any size is seen before large events in the PSD model.",
        "keywords": ["Accelerating seismic energy", "heterogeneous cellular automaton", "self-organized criticality", "critical point hypothesis"],
        "refs": null,
        "journal": "Pure and Applied Geophysics",
        "year": 2000,
        "doi": null
    },
    {
        "refID": "2000-1_Papazachos-BC_PAGEOPH",
        "title": "Accelerated Preshock Deformation of Broad Regions in the Aegean Area",
        "authors": ["Papazachos-BC", "Papazachos-CB"],
        "abstract": "Twenty-four regions where accelerating deformation has been observed for a few decades before corresponding strong (M = 6.0—7.5) mainshocks are identified in the broader Aegean area. To a first approximation these preshock regions have elliptical shapes and the radius, R (in km), of a circle with an area equal to the corresponding ellipse is related to the moment magnitude, M, of the mainshock by the equation:
```

```
    log R = 0.42 M-0.68. The dimension of each preshock region is
    about seven to ten times larger than the rupture zone (fault
    length) of the corresponding mainshock. The time variation of
    the cumulative Benioff strain was satisfactorily fitted by a
    power-law relation, which is predicted by statistical physics
    if the mainshock to which accelerating strain rates leads is
    considered as a critical point. The duration, t (in years), of
    the accelerating Benioff strain release period is given by the
    relation: log t = 5.94-0.75 log s r where s r is the mean Benioff
    strain rate release (per year for 104 km2) in the preshock region
    calculated by the complete available data (M≥5.2) for the entire
    instrumental period (1911—1998). The importance of identifying
    and investigating such regions for better understanding the
    dynamics of the active part of the lithosphere as well as for
    earthquake   prediction   and   time-dependent   seismic   hazard
    assessment is discussed.",
        "keywords": ["Accelerated preshock deformation", "Benioff
    strain", "critical phenomena", "Aegean area"],
        "refs": null,
        "journal": "Pure and Applied Geophysics",
        "year": 2000,
        "doi": null
},
{
        "refID": "2000-1_Robinson-R_GJI",
        "title": "A test of the precursory accelerating moment
    release model on some recent New Zealand earthquakes",
        "authors": ["Robinson-R"],
        "abstract": "The proposal that the moment release rate
    increases in a systematic way in a large region around a
    forthcoming large earthquake is tested using three recent, large
    New Zealand events. The three events, 1993—1995, magnitudes 6.7—
    7.0, occurred in varied tectonic settings. For all three events,
    a circular precursory region can be found such that the moment
    release   rate   of   the   included   seismicity   is   modelled
    significantly better by the proposed accelerating model than by
    a linear moment release model, although in one case the result
    is dubious. The 'best' such regions have radii from 122 to 167
    km, roughly in accord with previous observations world-wide, but
    are offset by 50—60 km from the associated main shock epicentre.
    A grid-search procedure is used to test whether these three
    earthquakes could have been forecast using the accelerating
    moment release model. For two of the earthquakes the result is
    positive in terms of location, but the main shock times are only
    loosely constrained.",
        "keywords":  ["earthquake  prediction",  "New  Zealand",
    "seismicity"],
        "refs": null,
        "journal": "Geophysical Journal International",
        "year": 2000,
        "doi": "10.1046/j.1365-246X.2000.00054.x"
},
```

```
{
        "refID": "2000-1_Rundle-JB_PAGEOPH",
        "title": "Precursory Seismic Activation and Critical-point
Phenomena",
        "authors": ["Rundle-JB", "Klein-W", "Turcotte-DL",
"Malamud-BD"],
        "abstract": "In this paper we relate the behavior of
seismicity prior to a characteristic earthquake to the
excitation in proximity to a spinodal instability. We illustrate
the spinodal instability as the upper limit of superheated water
prior to a steam explosion. We draw an analogy between the steam
explosion and a characteristic earthquake, and show that the
power-law activation associated with the spinodal instability
is essentially identical to the power-law increase in Benioff
strain observed prior to characteristic earthquakes. We find
that theory and actual data give very similar results.",
        "keywords": ["Seismicity", "spinodal", "critical-point
phenomena", "characteristic earthquake", "Benioff strain"],
        "refs": null,
        "journal": "Pure and Applied Geophysics",
        "year": 2000,
        "doi": null
},
{
        "refID": "2001-1_Bowman-DD_GRL",
        "title": "Accelerating Seismicity and Stress Accumulation
Before Large Earthquakes",
        "authors": ["Bowman-DD", "King-GCP"],
        "abstract": "The stress field that existed before a large
earthquake can be calculated based on the known source
parameters of the event. This stress field can be used to define
a region that shows greater seismic moment rate changes prior
to the event than arbitrarily shaped regions, allowing us to
link two previously unrelated subjects: Coulomb stress
interactions and accelerating seismicity before large
earthquakes. As an example, we have examined all M ≥ 6.5
earthquakes in California since 1950. While we illustrate the
model using seismicity in California, the technique is general
and can be applied to any tectonically active region. We show
that where sufficient knowledge of the regional tectonics
exists, this method can be used to augment current techniques
for seismic hazard estimation.",
        "keywords": null,
        "refs": null,
        "journal": "Geophysical Research Letters",
        "year": 2001,
        "doi": "10.1029/2001GL013022 "
},
{
        "refID": "2001-1_Bowman-DD_CRAS",
        "title": "Stress transfer and seismicity changes before
large earthquakes",
```

```
        "authors": ["Bowman-DD", "King-GCP"],
        "abstract": "In recent years, observational and theoretical
descriptions of spatio-temporal patterns of seismicity have
focused on two fundamental (and controversial) observations:
static stress (Coulomb) interactions between earthquakes and
accelerating seismic moment release before large earthquakes.
While there have been several documented examples of static
stress changes influencing the space-time pattern of seismicity
following great earthquakes (main shocks and aftershocks), there
have been few attempts to link this method to the evolution of
seismicity before great earthquakes (precursory seismicity and
foreshocks). In this paper, we describe a simple physical model
that links static stress modeling to accelerating moment release
before a large event. For practical reasons, it is not
straightforward to apply this technique as a method of
forecasting future large earthquakes. However, after the large
event has occurred, the region of stress accumulation can be
calculated with precision based on the known source parameters
of the earthquake. This region can then be examined for seismic
moment rate changes prior to the event. As examples, we have
examined all M⩾6.5 earthquakes in California since 1950 in
regions defined by their pre-event stress fields, and find a
period of accelerating moment release before all of these
events. While we illustrate the model using seismicity in
California, the technique is general and can be applied to any
tectonically active region. Where sufficient knowledge of the
regional tectonics exists, this method can be used to augment
current techniques for seismic hazard estimation.",
        "keywords": ["earthquake", "California", "stress
transfer", "accelerating seismicity", "Coulomb stress"],
        "refs": null,
        "journal": "Comptes Rendus de l'Académie des Sciences",
        "year": 2001,
        "doi": "10.1016/S1251-8050(01)01677-9"
},
{
        "refID": "2001-1_DiGiovambattista-R_Tectonophys",
        "title": "An analysis of the process of acceleration of
seismic energy emission in laboratory experiments on destruction
of rocks and before strong earthquakes on Kamchatka and in
Italy",
        "authors": ["DiGiovambattista-R", "Tyupkin-YS"],
        "abstract": "The time-to-failure model is a technique in
which a failure function is fitted to a time series of
accumulated Benioff strain before a large earthquake. We analyze
the relation of the time-to-failure model to the hypothesis of
fractal structure of seismicity. A power law failure function
(Varnes, 1989; Bufe and Varnes, 1993) and its log-periodic
generalization Sornette and Sammis, 1995 are discussed. The
results of application of the log-periodic time-to-failure model
to the analysis of the process of acceleration of seismic energy
emission in the laboratory experiments on rock destruction and
```

```
        before strong earthquakes on Kamchatka and in Italy are
presented.",
        "keywords": ["earthquake prediction", "precursory seismic
activity", "foreshock", "fractal dimension", "fractal
manifold"],
        "refs": null,
        "journal": "Tectonophysics",
        "year": 2001,
        "doi": "10.1016/S0040-1951(01)00088-9"
},
{
        "refID": "2001-1_Papazachos-CB_AnnGeofis",
        "title": "Precursory accelerated Benioff strain in the
Aegean area",
        "authors": ["Papazachos-CB", "Papazachos-BC"],
        "abstract": "Accelerating seismic crustal deformation due
to the occurrence of intermediate magnitude earthquakes leading
to the generation of a mainshock has recently been considered a
critical phenomenon. This hypothesis is tested by the use of a
large data sample concerning the Aegean area. Elliptical
critical regions for fifty-two strong mainshocks, which have
occurred in the Aegean area since 1930, have been identified by
applying a power-law relation between the cumulative Benioff
strain and the time to the main rupture. Empirical relations
between the parameters of this model have been further improved
by the use of a large data sample. The spatial distribution of
preshocks with respect to the mainshock is examined and its
tectonic significance is pointed out. The possibility of using
the results of this work to predict the epicentre, magnitude and
time of ensuing mainshocks are discussed and further work
towards this goal is suggested.",
        "keywords": ["Benioff strain", "critical point", "Aegean
area"],
        "refs": null,
        "journal": "Annali di Geofisica",
        "year": 2001,
        "doi": null
},
{
        "refID": "2001-1_VereJones-D_GJI",
        "title": "Remarks on the accelerated moment release model:
problems of model formulation, simulation and estimation",
        "authors": ["VereJones-D", "Robinson-R", "Yang-WZ"],
        "abstract": "This report summarizes a variety of issues
concerning the development of statistical versions of the so-
called 'accelerated moment release model' (AMR model). Until
such statistical versions are developed, it is not possible to
develop satisfactory procedures for simulating, fitting or
forecasting the model. We propose a hierarchy of simulation
models, in which the increase in moment is apportioned in varying
degrees between an increase in the average size of events and
an increase in their frequency. To control the size
```

```
distribution, we propose a version of the Gutenberg—Richter
power law with exponential fall-off, as suggested in recent
papers by Kagan. The mean size is controlled by the location of
the fall-off, which in turn may be related to the closeness to
criticality of the underlying seismic region. Other points
touched on concern the logical structure of the model, in
particular the identifiability of the parameter assumed to
control the size of the main shock, and appropriate procedures
to use for simulation and estimation. An appendix summarizes
properties of the Kagan distribution. The simulations highlight
the difficulty in identifying an AMR episode with only limited
data.",
        "keywords": null,
        "refs": null,
        "journal": "Geophysical Journal International",
        "year": 2001,
        "doi": "10.1046/j.1365-246x.2001.01348.x"
    },
    {
        "refID": "2001-1_Yang-WZ_JGR",
        "title": "A proposed method for locating the critical
region of a future earthquake using the critical earthquake
concept",
        "authors": ["Yang-WZ", "VereJones-D", "Ma-L"],
        "abstract": "Using the critical point concept and extending
Bowman's idea of critical earthquake, we develop an intersecting
circle method to locate the critical region. A simulation check
shows that this method is effective in finding a given critical
region. We selected several real cases from New Zealand and
China and used this method to find the critical regions before
the occurrence of large earthquakes. The result shows that this
method is valid for detecting a critical region and the epicenter
of mainshock might be in the critical region.",
        "keywords": null,
        "refs": null,
        "journal": "Journal of Geophysical Research",
        "year": 2001,
        "doi": "10.1029/2000JB900311"
    },
    {
        "refID": "2001-1_Zoeller-G_JGR",
        "title": "Observation of growing correlation length as an
indicator for critical point behavior prior to large
earthquakes",
        "authors": ["Zoeller-G", "Hainzl-S", "Kurths-J"],
        "abstract": "We test the critical point concept for
earthquakes in terms of the spatial correlation length. A system
near a critical point is associated with a diverging correlation
length following a power law time-to-failure relation. We
estimate the correlation length directly from an earthquake
catalog using single-link cluster analysis. Therefore we assume
that the distribution of moderate earthquakes reflects the state
```

```
      of the regional stress field. The parameters of the analysis are
      determined by an optimization procedure, and the results are
      tested against a Poisson process with realistic distributions
      of epicenters, magnitudes, and aftershocks. A systematic
      analysis of all earthquakes with M≥6.5 in California since 1952
      is conducted. In fact, we observe growing correlation lengths
      in most cases. The null hypothesis that this behavior can be
      found in random data is rejected with a confidence level of more
      than 99%. Furthermore, we find a scaling relation log R~0.7M
      (log ⟨ξmax ~ 0.5M), between the mainshock magnitude M and the
      critical region R (the correlation length ⟨ξmax before the
      mainshock), which is in good agreement with theoretical
      values.",
      "keywords": null,
      "refs": null,
      "journal": "Journal of Geophysical Research",
      "year": 2001,
      "doi": "10.1029/2000JB900379"
},
{
      "refID": "2002-1_BenZion-Y_PAGEOPH",
      "title": "Accelerated Seismic Release and Related Aspects
of Seismicity Patterns on Earthquake Faults",
      "authors": ["BenZion-Y", "Lyakhovsky-V"],
      "abstract": "Observational studies indicate that large
earthquakes are sometimes preceded by phases of accelerated
seismic release (ASR) characterized by cumulative Benioff strain
following a power law timeto-failure relation with a term (tf –
t)m, where tr is the failure time of the large event and observed
values of m are close to 0.3. We discuss properties of ASR and
related aspects of seismicity patterns associated with several
theoretical frameworks. The subcritical crack growth approach
developed to describe deformation on a crack prior to the
occurrence of dynamic rupture predicts great variability and low
asymptotic values of the exponent m that are not compatible with
observed ASR phases. Statistical physics studies assuming that
system-size failures in a deforming region correspond to
critical phase transitions predict establishment of long-range
correlations of dynamic variables and power-law statistics
before large events. Using stress and earthquake histories
simulated by the model of BEN-ZION (1996) for a discrete fault
with quenched heterogeneities in a 3-D elastic half space, we
show that large model earthquakes are associated with
nonrepeating cyclical establishment and destruction of long-
range stress correlations, accompanied by nonstationary
cumulative Benioff strain release. We then analyze results
associated with a regional lithospheric model consisting of a
seismogenic upper crust governed by the damage rheology of
LYAKHOVSKY et al. (39) over a viscoelastic substrate. We
demonstrate analytically for a simplified 1-D case that the
employed damage rheology leads to a singular power-law equation
for strain proportional to (t f – t)-1/3, and a nonsingular
```

```
      power-law relation for cumulative Benioff strain proportional
      to (t f - t)-1/3,A simple approximate generalization of the
      latter for regional cumulative Benioff strain is obtained by
      adding to the result a linear function of time representing a
      stationary background release. To go beyond the analytical
      expectations, we examine results generated by various
      realizations of the regional lithospheric model producing
      seismicity following the characteristic frequency-size
      statistics, Gutenberg-Richter power-law distribution, and mode
      switching activity. We find that phases of ASR exist only when
      the seismicity preceding a given large event has broad
      frequency-size statistics. In such cases the simulated ASR
      phases can be fitted well by the singular analytical relation
      with m = -1/3, the nonsingular equation with m = 0.2, and the
      generalized version of the latter including a linear term with
      m = 1/3. The obtained good fits with all three relations
      highlight the difficulty of deriving reliable information on
      functional forms and parameter values from such data sets. The
      activation process in the simulated ASR phases is found to be
      accommodated both by increasing rates of moderate events and
      increasing average event size, with the former starting a few
      years earlier than the latter. The lack of ASR in portions of
      the seismicity not having broad frequency-size statistics may
      explain why some large earthquakes are preceded by ASR and other
      are not. The results suggest that observations of moderate and
      large events contain two complementary end-member predictive
      signals on the time of future large earthquakes. In portions of
      seismicity following the characteristic earthquake
      distribution, such information exists directly in the associated
      quasi-periodic temporal distribution of large events. In
      portions of seismicity having broad frequency-size statistics
      withrandom or clustered temporal distribution of large events,
      the ASR phases have predictive information. The extent to which
      natural seismicity may be understood in terms of these end-
      member cases remains to be clarified. Continuing studies of
      evolving stress and other dynamic variables in model
      calculations combined with advanced analyses of simulated and
      observed seismicity patterns may lead to improvements in
      existing forecasting strategies.",
      "keywords": ["Continuum mechanics", "damage rheology",
"heterogeneous faults", "seismicity patterns", "large
earthquake cycles"],
      "refs": null,
      "journal": "Pure and Applied Geophysics",
      "year": 2002,
      "doi": null
},
{
      "refID": "2002-1_Karakaisis-GF_GJI",
      "title": "Accelerating seismic crustal deformation in the
North Aegean Trough, Greece",
```

```
        "authors": ["Karakaisis-GF", "Papazachos-CB", "Savvaidis-AS", "Papazachos-BC"],
        "abstract": "A recently developed algorithm has been applied to define regions of the northern Aegean in which accelerating seismic crustal deformation is currently occurring. An elliptical such region has been found in the western part of the North Aegean. Accelerating deformation, which started three decades ago and has been released by the generation of intermediate-magnitude earthquakes (M ≥ 4.5), is still occurring. Based on these observations we can assume that this region is now in a state (pre-shock deformation) that will lead to a critical point (main shock). The estimated basic parameters of this impending main shock are φ = 39.7°N, λ = 23.7°E for the epicentre, M = 6.0 for the moment magnitude, and tc = 2001.1 for the origin time. The corresponding uncertainties are less than 100 km for the epicentre, ± 0.4 for the magnitude, and ±1.5 yr for the origin time.",
        "keywords": ["accelerating seismic deformation", "Greece", "North Aegean"],
        "refs": null,
        "journal": "Geophysical Journal International",
        "year": 2002,
        "doi": "10.1046/j.0956-540x.2001.01578.x"
},
{
        "refID": "2002-1_Papazachos-BC_Tectonophys",
        "title": "Precursory accelerating seismic crustal deformation in the Northwestern Anatolia Fault Zone",
        "authors": ["Papazachos-BC", "Savvaidis-AS", "Karakaisis-GF", "Papazachos-CB"],
        "abstract": "We present the results of a systematic search for the identification of accelerating seismic crustal deformation in the broader northern Aegean area and in northwestern Turkey. We found that accelerating seismic deformation release, expressed by the generation of intermediate magnitude earthquakes, is currently observed in NW Turkey. On the basis of the critical earthquake model and by applying certain constraints which hold between the basic quantities involved in this phenomenon, it can be expected that this accelerating seismic activity may culminate in the generation of two strong earthquakes in this area during the next few years. The estimated epicenter coordinates of the larger of these probably impending earthquakes are 39.7°N—28.8°E, its magnitude is 7.0 and its occurrence time tc=2003.5. The second strong event is expected to occur at tc=2002.5 with a magnitude equal to 6.4 and epicenter coordinates 40.0°N—27.4°E. The uncertainties in the calculated focal parameters for these expected events are of the order of 100 km for the epicenter, ±0.5 for their magnitude and ±1.5 years for their occurrence time.",
        "keywords": ["Accelerating seismic deformation", "Anatolia", "Aegean"],
        "refs": null,
```

```
        "journal": "Tectonophysics",
        "year": 2002,
        "doi": "10.1016/S0040-1951(02)00030-6"
},
{
        "refID": "2002-1_Papazachos-CB_BSSA",
        "title": "Accelerating Seismic Crustal Deformation in the Southern Aegean Area",
        "authors": ["Papazachos-CB", "Karakaisis-GF", "Savvaidis-AS", "Papazachos-BC"],
        "abstract": "A region of intense accelerating seismic crustal deformation has been identified in the southwestern part of the Hellenic arc (broader area of Cythera island). The identification is performed using a detailed parametric grid search of the broader southern Aegean area for accelerating energy release behavior. The identified region has similar properties with past preshock (critical) regions, which have been identified for strong mainshocks in the Aegean area. Based on such observations, which suggest that this region is at a critical state that can lead to a critical point, that is, to the generation of a mainshock, an estimation is made of the possible epicenter coordinates, magnitude, and origin time of this oncoming large (M ~7.0) earthquake. The estimation procedure is validated on the basis of retrospective analysis of strong events in the Aegean area, as well as by appropriate application on synthetic random catalogs. These results, the existence of similar observations of accelerating seismic deformation in eastern part of southern Aegean and independent information on the time distribution of large earthquakes (M ≥6.8) for the whole southern Aegean indicate that the generation of strong earthquakes in this area in the next few years must be considered as very probable.",
        "keywords": null,
        "refs": null,
        "journal": "Bulletin of the Seismological Society of America",
        "year": 2002,
        "doi": "10.1785/0120000223"
},
{
        "refID": "2002-1_Papazachos-CB_BSSA",
        "title": "Precursory seismic crustal deformation in the area of southern Albanides",
        "authors": ["Papazachos-CB", "Karakaisis-GF", "Savvaidis-AS", "Papazachos-BC"],
        "abstract": "On the basis of growing evidence thatstrong earthquakes are preceded by a periodof accelerating seismicity of moderatemagnitude earthquakes, an attempt is madeto search for such seismicity pattern in NWAegean area. Accelerating seismic crustaldeformation has been identified in the areaof southern Albanides mountain range(border region between Greece, formerYugoslavia and Albania). Based on certainproperties of
```

```
        this   activity   and   on   itssimilarity   with   accelerating
seismicdeformation   observed   before   a   strongearthquake   which
occurred  in  the  sameregion  on  26  May  1960  (M  =  6.5),  we
canconclude  that  a  similar  earthquake  may  begenerated  in  the
same   region   during   thenext   few   years.   This   conclusion   is
inagreement with independent results whichhave been derived on
the basis of the timepredictable model.",
        "keywords":  ["accelerating  seismic  crustal  deformation",
"Aegean area", "Albanides", "precursory seismicity patterns"],
        "refs": null,
        "journal": "Journal of Seismology",
        "year": 2002,
        "doi": null
},
{
        "refID": "2002-1_Sammis-CG_PNAS",
        "title": "Positive feedback, memory, and the predictability
of earthquakes",
        "authors": ["Sammis-CG", "Sornette-D"],
        "abstract":  "We  review  the  'critical  point'  concept  for
large  earthquakes  and  enlarge  it  in  the  framework  of  so-called
'finite-time   singularities.'   The   singular   behavior   associated
with  accelerated  seismic  release  is  shown  to  result  from  a
positive feedback of the seismic activity on its release rate.
The  most  important  mechanisms  for  such  positive  feedback  are
presented.  We  solve  analytically  a  simple  model  of  geometrical
positive  feedback  in  which  the  stress  shadow  cast  by  the  last
large  earthquake  is  progressively  fragmented  by  the  increasing
tectonic stress.",
        "keywords": null,
        "refs": null,
        "journal": "Proceedings of the National Academy of Sciences
of the United States of America",
        "year": 2002,
        "doi": "10.1073/pnas.012580999"
},
{
        "refID": "2002-1_Weatherley-D_PAGEOPH",
        "title":   "Long-range   Automaton   Models   of   Earthquakes:
Power-law  Accelerations,  Correlation  Evolution,  and  Mode-
switching",
        "authors": ["Weatherley-D", "Mora-P", "Xia-MF"],
        "abstract":  "We  introduce  a  conceptual  model  for  the  in-
plane  physics  of  an  earthquake  fault.  The  model  employs  cellular
automaton  techniques  to  simulate  tectonic  loading,  earthquake
rupture,  and  strain  redistribution.  The  impact  of  a  hypothetical
crustal  elastodynamic  Green's  function  is  approximated  by  a
long-range strain redistribution law with a r −p dependance. We
investigate   the   influence   of   the   effective   elastodynamic
interaction  range  upon  the  dynamical  behaviour  of  the  model  by
conducting  experiments  with  different  values  of  the  exponent
(p).  The  results  indicate  that  this  model  has  two  distinct,
```

stable modes of behaviour. The first mode produces a characteristic earthquake distribution with moderate to large events preceeded by an interval of time in which the rate of energy release accelerates. A correlation function analysis reveals that accelerating sequences are associated with a systematic, global evolution of strain energy correlations within the system. The second stable mode produces Gutenberg-Richter statistics, with near-linear energy release and no significant global correlation evolution. A model with effectively short-range interactions preferentially displays Gutenberg-Richter behaviour. However, models with long-range interactions appear to switch between the characteristic and GR modes. As the range of elastodynamic interactions is increased, characteristic behaviour begins to dominate GR behaviour. These models demonstrate that evolution of strain energy correlations may occur within systems with a fixed elastodynamic interaction range. Supposing that similar mode-switching dynamical behaviour occurs within earthquake faults then intermediate-term forecasting of large earthquakes may be feasible for some earthquakes but not for others, in alignment with certain empirical seismological observations. Further numerical investigation of dynamical models of this type may lead to advances in earthquake forecasting research and theoretical seismology.",
        "keywords": ["Critical point hypothesis", "cellular automata", "correlation evolution"],
        "refs": null,
        "journal": "Pure and Applied Geophysics",
        "year": 2002,
        "doi": null
    },
    {
        "refID": "2002-1_Yin-XC_PAGEOPH",
        "title": "Load-Unload Response Ratio and Accelerating Moment/Energy Release Critical Region Scaling and Earthquake Prediction",
        "authors": ["Yin-XC", "Mora-P", "Peng-K", "Wang-Y", "Weatherley-D"],
        "abstract": "The main idea of the Load-Unload Response Ratio (LURR) is that when a system is stable, its response to loading corresponds to its response to unloading, whereas when the system is approaching an unstable state, the response to loading and unloading becomes quite different. High LURR values and observations of Accelerating Moment/Energy Release (AMR/AER) prior to large earthquakes have led different research groups to suggest intermediate-term earthquake prediction is possible and imply that the LURR and AM R/AER observations may have a similar physical origin. To study this possibility, we conducted a retrospective examination of several Australian and Chinese earthquakes with magnitudes ranging from 5.0 to 7.9, including Australia's deadly Newcastle earthquake and the devastating Tangshan earthquake. Both LURR values and best-fit power-law

```
        time-to-failure functions were computed using data within a
range of distances from the epicenter. Like the best-fit power-
law fits in AMR/AER, the LURR value was optimal using data within
a certain epicentral distance implying a critical region for
LURR. Furthermore, LURR critical region size scales with
mainshock magnitude and is similar to the AMR/AER critical
region size. These results suggest a common physical origin for
both the AM R/AER and LURR observations. Further research may
provide clues that yield an understanding of this mechanism and
help lead to a solid foundation for intermediate-term earthquake
prediction.",
        "keywords": ["LURR (Load-Unload Response Ratio)", "AMR
(Accelerating Moment Release)", "AER (Accelerating Energy
Release)", "CPH (Critical Point Hypothesis) ", "earthquake
prediction", "critical region scaling"],
        "refs": null,
        "journal": "Pure and Applied Geophysics",
        "year": 2002,
        "doi": null
    },
    {
        "refID": "2002-1_Zoeller-G_GRL",
        "title": "A systematic spatiotemporal test of the critical
point hypothesis for large earthquakes",
        "authors": ["Zoeller-G", "Hainzl-S"],
        "abstract": "The critical point hypothesis for large
earthquakes predicts two different precursory phenomena in space
and time, an accelerating moment release and the growth of the
spatial correlation length. The objective of this work is to
investigate both methods with respect to their predictive power.
A systematic statistical test based on appropriate random
earthquake catalogs allows to quantify the correlations of a
precursory pattern with the subsequent mainshock activity. The
analysis of target earthquakes in California since 1960 with
magnitudes M ≥ Mcut reveals that these correlations increase
systematically with growing Mcut, and correlations at greater
than 95% confidence are observed for Mcut ≥ 6.5 in the case of
the spatial correlation length. In particular, the seismicity
patterns are found to be significantly correlated with each of
the largest earthquakes (M ≥ 7.0), individually. The
acceleration of the moment release has a similar trend, but is
less significant.",
        "keywords": null,
        "refs": null,
        "journal": "Geophysical Research Letters",
        "year": 2002,
        "doi": "10.1029/2002GL014856"
    },
    {
        "refID": "2003-1_Chen-CC_GJI",
        "title": "Accelerating seismicity of moderate-size
earthquakes before the 1999 Chi-Chi, Taiwan, earthquake: Testing
```

```
        time-prediction of the self-organizing spinodal model of
earthquakes",
        "authors": ["Chen-CC"],
        "abstract": "Seismic activation of moderate-size
earthquakes for the 1999 Chi-Chi, Taiwan, earthquake has been
found. A self-organizing spinodal (SOS) model can explain some
observations concerning seismic activation, but the equal time
durations of the mid and precursory periods during an earthquake
cycle conjectured in the original, published, SOS model have not
been supported in this case. The Chi-Chi test presented here
shows unequal time durations of the mid and precursory periods
of an earthquake cycle. This, in turn, makes the possibility of
time prediction of a characteristic earthquake impossible in the
context of the SOS model. In addition, comparisons with
numerical simulations of the sliding-block model suggest the
change in the system's stiffness is a potential mechanism of
seismic activation.",
        "keywords": ["Chi-Chi earthquake", "seismic activation",
"self-organizing spinodal model"],
        "refs": null,
        "journal": "Geophysical Journal International",
        "year": 2003,
        "doi": "10.1046/j.1365-246X.2003.02071.x"
    },
    {
        "refID": "2003-1_Helmstetter-A_JGR",
        "title": "Foreshocks explained by cascades of triggered
seismicity",
        "authors": ["Helmstetter-A", "Sornette-D"],
        "abstract": "The observation of foreshocks preceding large
earthquakes and the suggestion that foreshocks have specific
properties that may be used to distinguish them from other
earthquakes have raised the hope that large earthquakes may be
predictable. Among proposed anomalous properties are the larger
proportion than normal of large versus small foreshocks, the
power law acceleration of seismicity rate as a function of time
to the mainshock, and the spatial migration of foreshocks toward
the mainshock when averaging over many sequences. Using southern
California seismicity, we show that these properties and others
arise naturally from the simple model that any earthquake may
trigger other earthquakes, without arbitrary distinction between
foreshocks, aftershocks, and mainshocks. We find that foreshock
precursory properties are independent of the mainshock size.
This implies that earthquakes (large or small) are predictable
to the same degree as seismicity rate is predictable from past
seismicity by taking into account cascades of triggering. The
cascades of triggering give rise naturally to long-range and
long-time interactions, which can explain the observations of
correlations in seismicity over surprisingly large length
scales.",
        "keywords": null,
        "refs": null,
```

```
        "journal": "Journal of Geophysical Research",
        "year": 2003,
        "doi": "10.1029/2003JB002409"
},
{
        "refID": "2003-1_Karakaisis-GF_GJI",
        "title": "Accelerating seismic crustal deformation before the Izmit (NW Turkey) large mainshock of 1999 August 17 and the evolution of its aftershock sequence",
        "authors": ["Karakaisis-GF"],
        "abstract": "The large Izmit (NW Turkey) mainshock (1999 August 17, Mw= 7.6), followed by another large earthquake on 1999 November 12 (Mw= 7.2), caused extensive damage and loss of life in a zone approximately 170 km long along the coastal area of the Gulf of Izmit and further east to Adapazarí and Düzce. On the basis of the critical earthquake concept and by applying a recently developed optimization algorithm, an elliptical area surrounding the 1999 August 17 mainshock epicentre was identified, in which accelerating moderate magnitude seismic activity started in 1981 and culminated in the generation of the mainshock. On the other hand, the space and time distribution of the aftershocks of the first mainshock (1999 August 17) and the time variations of the b value of the Gutenberg—Richter recurrence law and the mean aftershock magnitude, suggest that the second mainshock (1999 November 12) might have been anticipated.",
        "keywords": null,
        "refs": null,
        "journal": "Geophysical Journal International",
        "year": 2003,
        "doi": "10.1046/j.1365-246X.2003.01883.x"
},
{
        "refID": "2003-1_Karakaisis-GF_PAGEOPH",
        "title": "Time Variation of Parameters Related to the Accelerating Preshock Crustal Deformation in the Aegean Area",
        "authors": ["Karakaisis-GF", "Savvaidis-AS", "Papazachos-CB"],
        "abstract": "The time variation of two parameters related to accelerating seismic deformation before strong earthquakes in the Aegean area is examined. The first is the b parameter of the Gutenberg-Richter relation and the second is the curvature parameter C, which is a measure of deviation of the accelerating preshock deformation from a linear time variation of this deformation. Following two different procedures, it was found that the b value exhibits a decreasing trend prior to the oncoming earthquake, in agreement with the results of laboratory experiments and other independent observations. C values also show a decreasing trend before main shocks. These results indicate that such time variations of these parameters can be considered as precursory phenomena of ensuing strong earthquakes.",
```

```
        "keywords": ["Accelerating seismicity", "b value",
"curvature parameter C", "Aegean area"],
        "refs": null,
        "journal": "Pure and Applied Geophysics",
        "year": 2003,
        "doi": null
},
{
        "refID": "2003-1_King-GCP_JGR",
        "title": "The evolution of regional seismicity between
large earthquakes",
        "authors": ["King-GCP", "Bowman-DD"],
        "abstract": "We describe a simple model that links static
stress (Coulomb) modeling to the regional seismicity around a
major fault. Unlike conventional Coulomb stress techniques,
which calculate stress changes, we model the evolution of the
stress field relative to the failure stress. Background
seismicity is attributed to inhomogeneities in the stress field
which are created by adding a random field that creates local
regions above the failure stress. The inhomogeneous field is
chosen such that when these patches fail, the resulting
earthquake size distribution follows a Gutenburg-Richter law.
Immediately following a large event, the model produces regions
of increased seismicity (aftershocks) where the overall stress
field has been elevated and regions of reduced seismicity where
the stress field has been reduced (stress shadows). The high
stress levels in the aftershock regions decrease due to loading
following the main event. Combined with the stress shadow from
the main event, this results in a broad seismically quiet region
of lowered stress around the epicenter. Pre-event seismicity
appears as the original stress shadows finally fill as a result
of loading. The increase in seismicity initially occurs several
fault lengths away from the main fault and moves inward as the
event approaches. As a result of this effect, the seismic moment
release in the region around the future epicenter increases as
the event approaches. Synthetic catalogues generated by this
model are virtually indistinguishable from real earthquake
sequences in California and Washington.",
        "keywords": null,
        "refs": null,
        "journal": "Journal of Geophysical Research",
        "year": 2003,
        "doi": "10.1029/2001JB000783"
},
{
        "refID": "2003-1_Turcotte-DL_GJI",
        "title": "Micro and macroscopic models of rock fracture",
        "authors": ["Turcotte-DL", "Newman-WI", "Shcherbakov-R"],
        "abstract": "The anelastic deformation of solids is often
treated using continuum damage mechanics. An alternative
approach to the brittle failure of a solid is provided by the
discrete fibre-bundle model. Here we show that the continuum
```

```
damage model can give exactly the same solution for material
failure as the fibre-bundle model. We compare both models with
laboratory experiments on the time-dependent failure of
chipboard and fibreglass. The power-law scaling obtained in both
models and in the experiments is consistent with the power-law
seismic activation observed prior to some earthquakes.",
    "keywords": ["critical point", "damage mechanics", "fibre-
bundle model", "fracture", "power-law scaling"],
    "refs": null,
    "journal": "Geophysical Journal International",
    "year": 2003,
    "doi": "10.1046/j.1365-246X.2003.01884.x"
},
{
    "refID": "2003-1_Tzanis-A_NHESS",
    "title": "Distributed power-law seismicity changes and
crustal deformation in the SW Hellenic ARC",
    "authors": ["Tzanis-A", "Vallianatos-F"],
    "abstract": "A region of definite accelerating seismic
release rates has been identified at the SW Hellenic Arc and
Trench system, of Peloponnesus, and to the south-west of the
island of Kythera (Greece). The identification was made after
detailed, parametric time-to-failure modelling on a 0.1° square
grid over the area 20° E ? 27° E and 34° N?38° N. The observations
are strongly suggestive of terminal-stage critical point
behaviour (critical exponent of the order of 0.25), leading to
a large earthquake with magnitude 7.1 ± 0.4, to occur at time
2003.6 ± 0.6. In addition to the region of accelerating seismic
release rates, an adjacent region of decelerating seismicity was
also observed. The acceleration/deceleration pattern appears in
such a well structured and organised manner, which is strongly
suggestive of a causal relationship. An explanation may be that
the observed characteristics of distributed power-law seismicity
changes may be produced by stress transfer from a fault, to a
region already subjected to stress inhomogeneities, i.e. a
region defined by the stress field required to rupture a fault
with a specified size, orientation and rake. Around a fault that
is going to rupture, there are bright spots (regions of
increasing stress) and stress shadows (regions relaxing stress);
whereas acceleration may be observed in bright spots,
deceleration may be expected in the shadows. We concluded that
the observed seismic release patterns can possibly be explained
with a family of NE-SW oriented, left-lateral, strike-slip to
oblique-slip faults, located to the SW of Kythera and
Antikythera and capable of producing earthquakes with magnitudes
MS ~ 7. Time-to-failure modelling and empirical analysis of
earthquakes in the stress bright spots yield a critical exponent
of the order 0.25 as expected from theory, and a predicted
magnitude and critical time perfectly consistent with the
figures given above. Although we have determined an approximate
location, time and magnitude, it is as yet difficult to assert
a prediction for reasons discussed in the text. However, our
```

```
        results, as well as similar independent observations by another
        research team, indicate that a strong earthquake may occur at
        the SW Hellenic Arc, in the next few years.",
        "keywords": null,
        "refs": null,
        "journal": "Natural Hazards and Earth System Science",
        "year": 2003,
        "doi": null
},
{
        "refID": "2004-1_Bowman-DD_PAGEOPH",
        "title": "Intermittent Criticality and the Gutenberg-
Richter Distribution",
        "authors": ["Bowman-DD", "Sammis-CG"],
        "abstract": "In recent years there has been renewed
interest in observations of accelerating moment release before
large earthquakes, as well as theoretical descriptions of
seismicity in terms of statistical physics. Most aspects of
these works are encompassed by a concept called intermittent
criticality in which a region alternately approaches and
retreats from a critical -point. From this perspective, the
evolution of seismicity in a region is described in terms of the
growth and destruction of correlation in the stress field over
the course of the seismic cycle. In this paper we test the
concept of intermittent criticality by investigating the
temporal evolution of the Gutenberg-Richter distribution before
and after two successiveM ≥5.0 earthquakes in western Washington
State. The largest event in this distribution, M maxis observed
to systematically increase before each event, producing
accelerating moment release, and then to subsequently decrease.
Associated variations in the b-value are minimal This is the
predicted result if M max is a measure of the correlation length
of the regional stress field.",
        "keywords": null,
        "refs": null,
        "journal": "Pure and Applied Geophysics",
        "year": 2004,
        "doi": "10.1007/s00024-004-2541-z"
},
{
        "refID": "2004-1_Bufe-CG_BSSA",
        "title": "Comparing the November 2002 Denali and November
2001 Kunlun Earthquakes",
        "authors": ["Bufe-CG"],
        "abstract": "Major strike-slip earthquakes recently
occurred in Alaska on the central Denali fault (M 7.9) on 3
November 2002, and in Tibet on the central Kunlun fault (M 7.8)
on 14 November 2001. Both earthquakes generated large surface
waves with MS [U.S. Geological Survey (USGS)] of 8.5 (Denali)
and 8.0 (Kunlun). Each event occurred on an east—west-trending
strike-slip fault situated near the northern boundary of an
intense deformation zone that is characterized by lateral
```

extrusion and rotation of crustal blocks. Each earthquake produced east-directed nearly unilateral ruptures that propagated 300 to 400 km. Maximum lateral surface offsets and maximum moment release occurred well beyond 100 km from the rupture initiation, with the events exhibiting by far the largest separations of USGS hypocenter and Harvard Moment Tensor Centroid (CMT) for strike-slip earthquakes in the 27-year CMT catalog. In each sequence, the largest aftershock was more than two orders of magnitude smaller than the mainshock. Regional moment release had been accelerating prior to the main shocks. The close proximity in space and time of the 1964 Prince William Sound and 2002 Denali earthquakes, relative to their rupture lengths and estimated return times, suggests that these events may be part of a recurrent cluster in the vicinity of a complex plate boundary.",
        "keywords": null,
        "refs": null,
        "journal": "Bulletin of the Seismological Society of America",
        "year": 2004,
        "doi": "10.1785/0120030185"
},
{
        "refID": "2004-1_DiGiovambattista-R_Tectonophys",
        "title": "Seismicity patterns before the M=5.8 2002, Palermo (Italy) earthquake: seismic quiescence and accelerating seismicity",
        "authors": ["DiGiovambattista-R", "Tyupkin-YS"],
        "abstract": "Seismic quiescence and accelerating seismic energy release are considered as possible spatio-temporal patterns of the preparation process of the 6 September 2002 Palermo, Italy, earthquake (M 5.8). The detailed properties of the quiescence are analyzed applying the RTL algorithm. The RTL algorithm is based on the analysis of the RTL prognostic parameter, which is designed in such a way that it has a negative value if, in comparison with long-term background, there is a deficiency of events in the time—space vicinity of the tested point. The RTL parameter increases if activation of seismicity takes place. The RTL algorithm identified that a seismic quiescence started from the beginning of November 2001 and reached its minimum at the end of May 2002. The Palermo 2002 earthquake occurred 2 months after the RTL parameter restored its long-term background level. The application of a log-periodic time-to-failure model gives a "predicted" (in retrospect) magnitude M=6.2 main shock on 5 May 2002.",
        "keywords": ["Seismicity", "Earthquakes", "Seismic quiescence", "Accelerating moment release", "Time-to-failure"],
        "refs": null,
        "journal": "Tectonophysics",
        "year": 2004,
        "doi": "10.1016/j.tecto.2004.04.001"
},

```
{
    "refID": "2004-1_Jaume-SC_JGR",
    "title": "Accelerating seismic release from a self-correcting stochastic model",
    "authors": ["Jaume-SC", "Bebbington-MS"],
    "abstract": "We investigate the conditions under which the "stress-release model," a stochastic version of the elastic rebound model, produces synthetic earthquake sequences characterized by Accelerating Seismic Release (ASR). In this model, the level, or "stress," of the process accumulates linearly with time through tectonic input and decreases as the result of earthquakes. These "stress drops" correspond to some power of the energy released in the earthquakes, either E0.5 (Benioff strain) or E (seismic moment). Earthquakes occur in a point process with rate controlled by the level of the process. We hypothesize that the critical factor in the appearance of ASR is the manner in which the event sizes depend on the level of the process. This is modeled by the square root of energy released following either a tapered Pareto or truncated Gutenberg-Richter distribution, with maximum earthquake size controlled by a "tail-off" or "truncation" point. As the tail-off point becomes large, so does the average size, corresponding to an "acceleration to criticality" of the system. We found that those cases where the underlying level of the process corresponded to accumulated seismic moment produced numerous ASR sequences, whereas those cases using accumulated Benioff strain as the level did not. These results suggest that the occurrence of ASR is strongly dependent on how large earthquakes affect the dynamics of the fault system in which they are embedded, and hopefully provide some insight into the mechanics of acceleration to criticality, i.e., on the possible causes of occurrence/nonoccurrence of ASR.",
    "keywords": null,
    "refs": null,
    "journal": "Journal of Geophysical Research",
    "year": 2004,
    "doi": "10.1029/2003JB002867"
},
{
    "refID": "2004-1_Karakaisis-GF_Tectonophys",
    "title": "Current accelerating seismic excitation along the northern boundary of the Aegean microplate",
    "authors": ["Karakaisis-GF", "Papazachos-CB", "Scordilis-EM", "Papazachos-BC"],
    "abstract": "According to previous observations [Geophys. Res. Lett. 27 (2000) 3957], the generation of large (M≥7.0) earthquakes in the western part of the north Anatolian fault system (Marmara Sea) is followed by strong earthquakes along the Northern Boundary of the Aegean microplate (NAB: northwesternmost Anatolia—northern Aegean—central Greece—Ionian islands). Therefore, it can be hypothesized that a seismic excitation along this boundary should be expected after the occurrence of
```

the Izmit 1999 earthquake (M=7.6). We have applied the method
of accelerating seismic crustal deformation, which is based on
concepts of critical point dynamics in an attempt to locate more
precisely those regions along the NAB where seismic excitation
is more likely to occur. For this reason, a detailed parametric
grid search of the broader NAB area was performed for the
identification of accelerating energy release behavior. Three
such elliptical critical regions have been identified with
centers along this boundary. The first region, (A), is centered
in the eastern part of this boundary (40.2°N, 27.2°E: southwest
of Marmara), the second region, (B), has a center in the middle
part of the boundary (38.8°N, 23.4°E: East Central Greece) and
the third region, (C), in the westernmost part of the boundary
(38.2°N, 20.9°E: Ionian Islands). The study of the time variation
of the cumulative Benioff strain in two of the three identified
regions (A and B) revealed that intense accelerating seismicity
is observed especially after the occurrence of the 1999 Izmit
mainshock. Therefore, it can be suggested that the seismic
excitation, at least in these two regions, has been triggered
by the Izmit mainshock. Estimations of the magnitudes and origin
times of the expected mainshocks in these three critical regions
have also been performed, assuming that the accelerating
seismicity in these regions will lead to a critical point, that
is, to the generation of mainshocks.",
        "keywords": ["Accelerating seismic deformation", "Aegean",
"Earthquake prediction"],
        "refs": null,
        "journal": "Tectonophysics",
        "year": 2004,
        "doi": "10.1016/j.tecto.2004.03.005"
},
{
        "refID": "2004-1_Sammis-CG_PAGEOPH",
        "title": "Anomalous Seismicity and Accelerating Moment
Release Preceding the 2001 and 2002 Earthquakes in Northern Baja
California, Mexico",
        "authors": ["Sammis-CG", "Bowman-DD", "King-GCP"],
        "abstract": "An algorithm recently developed by RUNDLE et
al. (2002) to find regions of anomalous seismic activity
associated with large earthquakes identified the location of an
M w = 5.6 earthquake near Calexico, Mexico. In this paper we
analyze the regional seismicity before this event, and a nearby
M w = 5.7 event, using time-to-failure algorithms developed by
BOWMAN et al. (1998) and BOWMAN and KING (2001a,b). The former
finds the radius of a circular region surrounding the epicenter
that optimizes the time-to-failure acceleration of seismic
release. The latter optimizes acceleration based on the expected
stress accumulation pattern for a dislocation source. Both
methods found a period of accelerating seismicity in an optimal
region, the size of which agrees with previously proposed
scaling relations. This positive result suggests that the Rundle
algorithm may provide a useful technique to identify regions of

```
        accelerating seismicity, which can then be analyzed using signal
        optimization time-to-failure techniques.",
        "keywords": ["Seismicity", "seismic hazard assessment",
"earthquake prediction", "earthquake physics", "earth-quake
stress interactions"],
        "refs": null,
        "journal": "Pure and Applied Geophysics",
        "year": 2004,
        "doi": "10.1007/s00024-004-2569-3"
    },
    {
        "refID": "2004-1_Scordilis-EM_JOSE",
        "title": "Accelerating seismic crustal deformation before
strong mainshocks in Adriatic and its importance for earthquake
prediction",
        "authors": ["Scordilis-EM", "Papazachos-CB", "Karakaisis-
GF", "Karakostas-VG"],
        "abstract":     "Time    accelerating    Benioff    strain
releasebefore the mainshock has been observed inall five cases
of strong (M > 6.0) shallowmainshocks, which have occurred
during thelast four decades in the area surroundingthe Adriatic
Sea. This observation supportsthe idea that strong mainshocks
arepreceded by accelerating seismic crustaldeformation due to
the generation ofintermediate magnitude shocks (preshocks).It
is further shown that the values ofparameters calculated from
these datafollow appropriately modified relations,which have
previously been proposed asadditional constraints to the
criticalearthquake model and to the correspondingmethod of
intermediate term earthquakeprediction. Thus, these results show
thatthe  identification  of  regions  wheretime-accelerating
Benioff  strain  followssuch  constraints  may  lead  to
usefulinformation concerning the epicenter,magnitude and origin
time of oncomingstrong mainshocks in this area. Theprocedure for
identification    of    thetime-acceleration    is    validated
byappropriate application on synthetic butrealistic random
catalogues. Largerdimension  of  critical  regions  in
Adriaticcompared to such regions in the Aegean isattributed to
an order of magnitude smallerseismic deformation of the crust
in theformer in comparison to the latter.",
        "keywords": ["accelerating seismic deformation", "Adriatic
sea", "earthquake prediction"],
        "refs": null,
        "journal": "Journal of Seismology",
        "year": 2004,
        "doi": null
    },
    {
        "refID": "2004-1_Wang-JC_TAO",
        "title": "Investigation of Seismicity in Central Taiwan
Using the Accelerating Seismic Energy Release Model",
        "authors": ["Wang-JC", "Shieh-CF"],
```

```
        "abstract": "A time-to-failure method developed from the
accelerating seismic energy release model is used to scrutinize
the seismicity of central Taiwan for 40 earthquakes with
magnitude greater than 4.5. First, mainshocks and their pre-
events are identified, and then adopted as observed data set.
The nonlinear time-to-failure equation is separated into two
linear equations, and then parameters are estimated by using
linear least-square twice. The model constructed from the
estimated local parameters, is then used to predict time-of-
failure and magnitude of mainshocks. Comparing predicted results
and 40 mainshocks, the maximum misfits are only 0.98 years in
time and 1.2 unit in magnitude, which indicate that accelerating
seismic energy release model could be applied in central Taiwan
as a useful tool for the study of seismicity.",
        "keywords": ["Accelerating seismic energy release model",
"Mainshock", "Pre-shock", "Time-of-failure", "Seismicity"],
        "refs": null,
        "journal":    "Terrestrial,    Atmospheric    and    Oceanic
Sciences",
        "year": 2004,
        "doi": null
},
{
        "refID": "2004-1_Wang-LY_ActaSeismSinica",
        "title":    "Characteristics    of    foreshock    and    its
identification",
        "authors": ["Wang-LY", "Chen-PY", "Wu-ZL", "Bai-TX"],
        "abstract": "In the paper, we analyze 117 moderate-strong
earthquakes occurred in Chinese mainland (M S≥5.5 in the east
and M S≥6.0 in the west) since 1970, among them, 11 earthquakes
(about 9%) have direct foreshocks and 63 earthquakes (about 51%)
have generalized foreshocks. The predominant time interval
between foreshock and main earthquake is no more than 30 days
with a spatial distance less than 50 km and a magnitude
difference over 1. From the digital seismic data in liaoning
Province, we know that direct foreshock had an obvious shear-
stress background before the M S=5.6 and M S=5.1 Xiuyan
earthquakes occurred on Nov. 29, 1999 and Jan.15, 2000.",
        "keywords": ["direct foreshock", "generalized foreshock",
"ambient shear stress"],
        "refs": null,
        "journal": "Acta Seismologica Sinica",
        "year": 2004,
        "doi": null
},
{
        "refID": "2004-1_Wang-Y_PAGEOPH",
        "title": "Spatio-temporal Scanning and Statistical Test of
the Accelerating Moment Release (AMR) Model Using Australian
Earthquake Data",
        "authors": ["Wang-Y", "Yin-C", "Mora-P", "Yin-XC", "Peng-
K"],
```

```
        "abstract": "The Accelerating Moment Release (AMR)
preceding earthquakes with magnitude above 5 in Australia that
occurred during the last 20 years was analyzed to test the
Critical Point Hypothesis. Twelve earthquakes in the catalog
were chosen based on a criterion for the number of nearby events.
Results show that seven sequences with numerous events recorded
leading up to the main earthquake exhibited accelerating moment
release. Two occurred near in time and space to other earthquakes
preceded by AMR. The remaining three sequences had very few
events in the catalog so the lack of AMR detected in the analysis
may be related to catalog incompleteness. Spatio-temporal
scanning of AMR parameters shows that 80% of the areas in which
AMR occurred experienced large events. In areas of similar
background seismicity with no large events, 10 out of 12 cases
exhibit no AMR, and two others are false alarms where AMR was
observed but no large event followed. The relationship between
AMR and Load-Unload Response Ratio (LURR) was studied. Both
methods predict similar critical region sizes, however, the
critical point time using AMR is slightly earlier than the time
of the critical point LURR anomaly.",
        "keywords": ["Critical Point Hypothesis", "Accelerating
Moment Release (AMR) model", "earthquake prediction", "Load-
Unload Response Ratio (LURR)", "Australia earthquakes"],
        "refs": null,
        "journal": "Pure and Applied Geophysics",
        "year": 2004,
        "doi": "10.1007/s00024-004-2563-9"
    },
    {
        "refID": "2004-1_Weatherley-D_PAGEOPH",
        "title": "Accelerating Precursory Activity within a Class
of Earthquake Analogue Automata",
        "authors": ["Weatherley-D", "Mora-P"],
        "abstract": "A statistical fractal automaton model is
described which displays two modes of dynamical behaviour. The
first mode, termed recurrent criticality, is characterised by
quasi-periodic, characteristic events that are preceded by
accelerating precursory activity. The second mode is more
reminiscent of SOC automata in which large events are not
preceded by an acceleration in activity. Extending upon previous
studies of statistical fractal automata, a redistribution law
is introduced which incorporates two model parameters: a
dissipation factor and a stress transfer ratio. Results from a
parameter space investigation indicate that a straight line
through parameter space marks a transition from recurrent
criticality to unpredictable dynamics. Recurrent criticality
only occurs for models within one corner of the parameter space.
The location of the transition displays a simple dependence upon
the fractal correlation dimension of the cell strength
distribution. Analysis of stress field evolution indicates that
recurrent criticality occurs in models with significant long-
```

```
range   stress   correlations.   A   constant   rate   of   activity   is
associated with a decorrelated stress field.",
    "keywords":   ["Critical   point   hypothesis",   "cellular
automata", "accelerating moment release"],
    "refs": null,
    "journal": "Pure and Applied Geophysics",
    "year": 2004,
    "doi": "10.1007/s00024-004-2546-x"
},
{
    "refID": "2005-1_Jiang-CS_BSSA",
    "title": "Test of the Preshock Accelerating Moment Release
(AMR)  in  the  Case  of  the  26  December  2004  Mw  9.0  Indonesia
Earthquake",
    "authors": ["Jiang-CS", "Wu-ZL"],
    "abstract":  "A  case  study  of  the  26  December  2004  Mw  9.0
earthquake  off  the  west  coast  of  northern  Sumatra,  Indonesia,
was   conducted   to   explore   whether   there   was   a   preshock
accelerating  moment  release  (amr)  process  for  the  intermediate
timescale.  The  Harvard  cmt  catalog  was  used  to  calculate  the
cumulative   moment   tensor   directly,   with   clearer   physical
significance  with  regard  to  the  deformation  prior  to  the  great
earthquake.  We  observed  that  average  moment  tensors  at  different
times   over   the   last   decade   before   the   great   earthquake   are
consistent,  and  are  similar  to  the  focal  mechanism  of  the  great
earthquake. However, the widely used cumulative scalar seismic
moment  and  cumulative  Benioff  strain  are  only  an  approximation
of   the   preshock   deformation.   To   test   the   robustness   of   the
accelerating   property   with   respect   to   the   selection   of
spatiotemporal parameters, we calculated the scaling coefficient
m  for  different  spatiotemporal  ranges.  The  curvature  parameter
q  was  used  to  quantify  the  difference  between  the  power-law  fit
and  the  linear  fit  to  ensure  the  statistical  significance  of  the
power-law-like  accelerating  behavior.  Grid  searching  over  the
(tf,  m)  space  was  conducted  to  explore  the  global  stability  of
the  solution.  The  result  showed  that  there  existed  a  reliable
preshock amr process before this great earthquake, with duration
of  a  quarter  of  a  century  and  a  spatial  range  from  800  to  1500
km,  providing  seemingly  positive  evidence  for  the  amr  model.
However, the failure time tf was not well constrained by the amr
analysis,  and  the  amr  model  may  be  problematic  for  a  longer
timescale.",
    "keywords": null,
    "refs": null,
    "journal":   "Bulletin   of   the   Seismological   Society   of
America",
    "year": 2005,
    "doi": "10.1785/0120050018"
},
{
    "refID": "2005-1_Jiang-CS_ActaSeismSinica",
```

        "title": "The December 26, 2004, off the west coast of northern Sumatra, Indonesia, MW=9.0, earthquake and the critical-point-like model of earthquake preparation",
        "authors": ["Jiang-CS", "Wu-ZL"],
        "abstract": "Long-term seismic activity prior to the December 26, 2004, off the west coast of northern Sumatra, Indonesia, M W=9.0 earthquake was investigated using the Harvard CMT catalogue. It is observed that before this great earthquake, there exists an accelerating moment release (AMR) process with the temporal scale of a quarter century and the spatial scale of 1 500 km. Within this spatial range, the M W=9.0 event falls into the piece-wise power-law-like frequency-magnitude distribution. Therefore, in the perspective of the critical-point-like model of earthquake preparation, the failure to forecast/predict the approaching and/or the size of this earthquake is not due to the physically intrinsic unpredictability of earthquakes.",
        "keywords": ["the 2004 off the west coast of northern Sumatra MW=9.0 earthquake", "accelerating moment release (AMR) before earthquakes", "power-law-like frequency-magnitude distribution", "critical-point-like model of earthquake preparation"],
        "refs": null,
        "journal": "Acta Seismologica Sinica",
        "year": 2005,
        "doi": null
    },
    {
        "refID": "2005-1_Papazachos-CB_BSSA",
        "title": "Global Observational Properties of the Critical Earthquake Model",
        "authors": ["Papazachos-CB", "Karakaisis-GF", "Scordilis-EM", "Papazachos-BC"],
        "abstract": "The preshock (critical) regions of 20 mainshocks with magnitudes between 6.4 and 8.3, which occurred recently (since 1980) in a variety of seismotectonic regimes (Greece, Anatolia, Himalayas, Japan, California), were identified and investigated. All these strong earthquakes were preceded by accelerating time-to-mainshock seismic crustal deformation (Benioff strain). The time variation of the cumulative Benioff strain follows a power law with a power value (m = 0.3) in very good agreement with theoretical considerations. We observed that the dimension of the critical region increased with increasing mainshock magnitude and with decreasing long-term seismicity rate of the region. An increase of the duration of this critical (preshock) phenomenon with decreasing long-term seismicity rate was also observed. This spatial and temporal scaling expresses characteristics of the critical earthquake model, which are of importance for earthquake prediction research. We also showed that the critical region of an oncoming mainshock coincides with the preparing

```
        region of this shock, where other precursory phenomena can be
observed.",
        "keywords": null,
        "refs": null,
        "journal": "Bulletin of the Seismological Society of
America",
        "year": 2005,
        "doi": "10.1785/0120040181
"
},
{
        "refID": "2005-1_Robinson-R_GRL",
        "title": "Precursory accelerating seismic moment release
(AMR) in a synthetic seismicity catalog: A preliminary study",
        "authors": ["Robinson-R", "Zhou-S", "Johnston-S",
"VereJones-D"],
        "abstract": "A power-law like acceleration of seismic
moment release (AMR) has been proposed as a precursor to large
earthquakes. Because of problems with real-world data, we have
used a synthetic seismicity model of 256 interacting faults
embedded in a 3-D elastic half-space to search for periods of
AMR preceding the largest events (Mw ~ 7.1). In only 5 of 18
cases does the AMR model fit the data significantly better than
a linear moment release, and then only weakly so. This
proportion, or higher, occurs in 8% of 1000 randomized catalogs.
We conclude that either AMR is unlikely to be a common precursor,
or that factors contributing to the AMR pattern in the real
world are missing from the synthetic model.",
        "keywords": null,
        "refs": null,
        "journal": "Geophysical Research Letters",
        "year": 2005,
        "doi": "10.1029/2005GL022576 "
},
{
        "refID": "2006-1_Jiang-CS_PAGEOPH",
        "title": "Benioff Strain Release Before Earthquakes in
China: Accelerating or Not?",
        "authors": ["Jiang-CS", "Wu-ZL"],
        "abstract": "We systematically analyzed the Benioff strain
release before 65 earthquakes with M S over 6.0 in China from
1978 to 2003 to investigate the generality of the widely
discussed accelerating moment release (AMR) phenomenon before
strong and intermediate-strength earthquakes. In this approach,
a strong or intermediate-strength earthquake is selected as a
"target earthquake," and retrospective analysis of seismic
activity before the "target earthquake" is performed. Simple
searching area (three circular areas with different radius
centered at the epicenter of the "target earthquake") and
unified temporal range (8 years) are taken in the analysis.
Justification of whether AMR exists is by both visual inspection
and by power-law curve fitting. It is found that more than 3/5
```

```
        of the earthquakes under consideration exhibit clear pre-shock
        AMR property, and 1/3 of the events seem to be sensitive to the
        searching area. AMR behavior shows apparent focal mechanism
        dependence: 15 out of 17 dip-slip earthquakes with stable moment
        release characteristics against the changing of searching areas
        exhibit AMR behavior, while 16 out of 25 strike-slip earthquakes
        with stable moment release characteristics exhibit AMR
        behavior.",
        "keywords": ["Accelerating moment release (AMR)",
"earthquakes in China", "critical-point-like model of
earthquakes", "intermediate-term earthquake prediction"],
        "refs": null,
        "journal": "Pure and Applied Geophysics",
        "year": 2006,
        "doi": "10.1007/s00024-006-0107-1"
},
{
        "refID": "2006-1_Mignan-A_JGR",
        "title": "An observational test of the origin of
accelerating moment release before large earthquakes",
        "authors": ["Mignan-A", "Bowman-DD", "King-GCP"],
        "abstract": "A progressive increase of seismic activity
distributed over a wide region around a future earthquake
epicenter is termed accelerating moment release (AMR). This
phenomenon has been observed in several studies over the last
15 years, although there is no consensus about the physical
origin of the effect. In a recent hypothesis known as the stress
accumulation (SA) model, the AMR is thought to result from the
last stage of loading in the earthquake cycle. In this view, the
increasing seismicity is due to minor stress release as the
whole region becomes sufficiently stressed for the major event
to occur. The stress accumulation model makes specific
predictions about the distribution of events in an AMR sequence.
Because the AMR is predicted to be a result of loading on the
main fault, the precursory activity should be concentrated in
the positive lobes of the far-field stresses calculated by a
backslip dislocation model of the main shock. To test this model,
AMR is first found in optimal circular regions around the
epicenters of each of the Mw ≥ 6.5 earthquakes in central and
southern California since 1950. A backslip dislocation model is
then used to determine which of the precursory events occur in
the regions predicted by stress accumulation. AMR is shown to
occur preferentially in the lobes of the backslip stress field
predicted by the stress accumulation model.",
        "keywords": null,
        "refs": null,
        "journal": "Journal of Geophysical Research",
        "year": 2006,
        "doi": "10.1029/2006JB004374"
},
{
        "refID": "2006-1_Mignan-A_EPSL",
```

```
        "title": "Seismic activity in the Sumatra—Java region prior
to the December 26, 2004 (Mw = 9.0—9.3) and March 28, 2005 (Mw
= 8.7) earthquakes",
        "authors":    ["Mignan-A",    "King-GCP",    "Bowman-DD",
"Lacassin-R", "Dmowska-R"],
        "abstract": "A promising approach to assessing seismic
hazards has been to combine the concept of seismic gaps with
Coulomb-stress change modeling to refine short-term earthquake
probability    estimates.    However,    in    practice    the    large
uncertainties in the seismic histories of most tectonically
active regions limit this approach since a stress increase is
only important when a fault is already close to failure. In
contrast, recent work has suggested that Accelerated Moment
Release (AMR) can help to identify when a stretch of fault is
approaching failure without any knowledge of the seismic history
of the region. AMR can be identified in the regions around the
Sumatra Subduction system that must have been stressed before
the 26 December 2004 and 28 March 2005 earthquakes. The effect
is clearest for the epicentral regions with less than a 2%
probability that it could occur in a random catalogue. Less
clear AMR is associated with the regions north of Sumatra around
the Nicobar and Andaman islands where rupture in the December
2004 earthquake was less vigorous. No AMR is found for the region
of the 1833 Sumatran earthquake suggesting that an event in this
region in the near future is unlikely. AMR similar to that before
the December 2004 and March 2005 events is found for a 750 km
stretch of the southeastern Sumatra and western Java subduction
system suggesting that it is close to failure. Should the whole
of this stretch break in a single event the magnitude could be
similar to the December 2004 earthquake.",
        "keywords": ["earthquake stress interactions", "seismicity
and tectonics", "subduction zones", "earthquake forecasting and
prediction", "accelerating moment release"],
        "refs": null,
        "journal": "Earth and Planetary Science Letters",
        "year": 2006,
        "doi": "10.1016/j.epsl.2006.01.058"
},
{
        "refID": "2006-1_Papazachos-BC_JOSE",
        "title": "A forward test of the precursory decelerating and
accelerating seismicity model for California",
        "authors": ["Papazachos-BC", "Scordilis-EM", "Papazachos-
CB", "Karakaisis-GF"],
        "abstract": "Accelerating strain energy released by the
generation of intermediate magnitude preshocks in a broad
(critical)   region,   and   decelerating   energy   released   in   a
narrower (seismogenic) region, is considered as a distinct
premonitory pattern useful in research for intermediate-term
earthquake prediction. Accelerating seismicity in the broad
region is satisfactorily interpreted by the critical earthquake
model and decelerating seismicity in the narrower region is
```

```
        attributed to stress relaxation due to pre-seismic sliding. To
        facilitate the identification of such patterns an algorithm has
        been developed on the basis of data concerning accelerating and
        decelerating preshock sequences of globally distributed already
        occurred strong mainshocks. This algorithm is applied in the
        present work to identify regions, which are currently in a state
        of accelerating seismic deformation and are associated with
        corresponding narrower regions, which are in a state of
        decelerating seismic deformation in California. It has been
        observed that a region which includes known faults in central
        California is in a state of decelerating seismic strain release,
        while the surrounding region (south and north California, etc.)
        is in a state of accelerating seismic strain release. This
        pattern corresponds to a big probably oncoming mainshock in
        central California. The epicenter, magnitude and origin time,
        as well as the corresponding model uncertainties of this
        probably ensuing big mainshock have been estimated, allowing a
        forward testing of the model's efficiency for intermediate-term
        earthquake prediction.",
        "keywords": ["Accelerating strain", "decelerating strain",
"earthquake prediction", "California"],
        "refs": null,
        "journal": "Journal of Seismology",
        "year": 2006,
        "doi": "10.1007/s10950-005-9009-4"
},
{
        "refID": "2006-1_Zhou-S_JGR",
        "title": "Tests of the precursory accelerating moment
release model using a synthetic seismicity model for Wellington,
New Zealand",
        "authors": ["Zhou-S", "Johnston-S", "Robinson-R",
"VereJones-D"],
        "abstract": "We have constructed a synthetic seismicity
model of the Wellington region, New Zealand, including seven
superfaults and 500 subfaults which are randomly positioned.
From this model, a synthetic catalogue of 2000 years duration,
containing events of magnitude 5.0 or more, has been generated.
The properties of the catalogue, such as the long-term slip
rates, b value, average activity rate, and hypocenter
distribution, are in accord with paleoseismic studies and the
real seismicity over the last 40 years. Such a synthetic
catalogue can replace the short, incomplete, and inhomogeneous
historic and instrumental records in research which needs a long
time duration and many strong shocks. We have used our catalogue
to examine tests for the existence of accelerating moment
release (AMR) before large events and compared the results with
those from random (Poisson) catalogues. We find that (1) the
apparent success rate is very dependent on the rules used to
define the test window; (2) when appropriately defined, the AMR
pattern occurs before about 20% of the strong (M ≥ 7.0) shocks
with a typical precursor time of about 22 years; (3) the AMR
```

```
        pattern is found almost equally frequently before large events
        in random catalogues; and (4) there are some false alarms (AMR
        pattern without a large event). This extended study reinforces
        the conclusion in our preliminary report: that in synthetic
        catalogues of the kind we have constructed, the AMR pattern is
        essentially an artefact of the method of sampling.",
        "keywords": null,
        "refs": null,
        "journal": "Journal of Geophysical Research",
        "year": 2006,
        "doi": "10.1029/2005JB003720 "
},
{
        "refID": "2007-1_Mignan-A_JGR",
        "title": "A mathematical formulation of accelerating moment
release based on the stress accumulation model",
        "authors": ["Mignan-A", "King-GCP", "Bowman-DD"],
        "abstract": "Large earthquakes can be preceded by a period
of accelerating seismic activity of moderate-sized earthquakes.
This phenomenon, usually termed accelerating moment release, has
yet to be clearly understood. A new mathematical formulation of
accelerating moment release is obtained from simple stress
transfer considerations, following the recently proposed stress
accumulation model. This model, based on the concept of elastic
rebound, simulates accelerating seismicity from theoretical
stress changes during an idealized seismic cycle. In this view,
accelerating moment release is simply the consequence of the
decrease, due to loading, of the size of a stress shadow due to
a previous earthquake. We show that a power law time-to-failure
equation can be expressed as a function of the loading rate on
the fault that is going to rupture. We also show that the m
value, which is the power law exponent, can be defined as m =
D/3, with D a parameter that takes into account the geometrical
shape of the stress lobes and the distribution of active faults.
In the stress accumulation model, the power law is not due to
critical processes.",
        "keywords": null,
        "refs": null,
        "journal": "Geophysical Research Letters",
        "year": 2007,
        "doi": "10.1029/2006JB004671"
},
{
        "refID": "2007-1_Papazachos-BC_BSSA",
        "title": "Evaluation of the Results for an Intermediate-
Term Prediction of the 8 January 2006 Mw 6.9 Cythera Earthquake
in the
Southwestern Aegean",
        "authors": ["Papazachos-BC", "Karakaisis-GF", "Papazachos-
CB", "Scordilis-EM"],
        "abstract": "During the past few decades the critical
earthquake model, which is based on observations concerning
```

accelerating seismic deformation and concepts of the critical point dynamics, has been proposed by various seismologists as a useful tool for intermediate-term earthquake prediction. A refined approach of this model has been previously applied to search for preshock (critical) regions in the southern Aegean, using all available data until the middle of 2002. A critical region corresponding to a large mainshock had been identified (Papazachos et al., 2002a,b) in the southwestern part of the Aegean, near the Cythera island. The predicted (in 2002) parameters for this ensuing earthquake are φ = 36.5° N, λ = 22.7° E for the epicental geographic coordinates (with a model uncertainty of 120 km), focal depth ≤100 km, moment magnitude M 6.9 ± 0.5, and origin time tc = 2006.4 ± 2.0. The generation of the strong Cythera earthquake on 8 January 2006 with M 6.9, epicenter coordinates φ = 36.2° N and λ = 23.4° E and a focal depth of h = 65 km satisfies this intermediate-term prediction. The region where significant macroseismic effects were anticipated from the predicted mainshock (Cythera, south Peloponnesus, west Crete, and west Cyclades) corresponds to the area where damage by the 8 January 2006 strong earthquake has been observed. The verification of this prediction is strong evidence that the intermediate-term prediction of strong earthquakes is potentially feasible, but additional forward testing of the model is needed to validate this result.",
        "keywords": null,
        "refs": null,
        "journal": "Bulletin of the Seismological Society of America",
        "year": 2007,
        "doi": "10.1785/0120060075"
    },
    {
        "refID": "2007-1_Papazachos-BC_JGR",
        "title": "Currently active regions of decelerating-accelerating seismic strain in central Asia",
        "authors": ["Papazachos-BC", "Scordilis-EM", "Panagiotopoulos-DG", "Papazachos-CB"],
        "abstract": "Accelerating preshock seismic strain in a broad (critical) region and decelerating preshock seismic strain in a narrower (seismogenic) region constitute a model for intermediate-term prediction of strong main shocks. An effort is made in the present work for a forward test of the Decelerating-Accelerating Seismic Strain (D-AS) model by identifying such patterns and estimating the corresponding, probably ensuing, strong main shocks (M ≥ 7.0) in central Asia (20°N—45°N, 42°E—105°E). Five such patterns have been identified, and the origin time, magnitude, and epicentral geographic coordinates of each of the corresponding main shocks have been estimated (predicted). Model uncertainties of the estimated time, magnitude, and space parameters of these probably ensuing main shocks, as well as appropriate statistical tests against a standard Gutenberg-Richter seismicity distribution, are also

```
        presented to allow a future objective evaluation of the model's
        efficiency for intermediate-term earthquake prediction.",
            "keywords": null,
            "refs": null,
            "journal": "Journal of Geophysical Research",
            "year": 2007,
            "doi": "10.1029/2006JB004587"
    },
    {
            "refID": "2008-1_Hardebeck-JL_JGR",
            "title": "Improved tests reveal that the accelerating
        moment release hypothesis is statistically insignificant",
            "authors": ["Hardebeck-JL", "Felzer-KR", "Michael-AJ"],
            "abstract": "We test the hypothesis that accelerating
        moment release (AMR) is a precursor to large earthquakes, using
        data from California, Nevada, and Sumatra. Spurious cases of AMR
        can arise from data fitting because the time period, area, and
        sometimes magnitude range analyzed before each main shock are
        often optimized to produce the strongest AMR signal. Optimizing
        the search criteria can identify apparent AMR even if no robust
        signal exists. For both 1950—2006 California-Nevada M ≥ 6.5
        earthquakes and the 2004 M9.3 Sumatra earthquake, we can find
        two contradictory patterns in the pre—main shock earthquakes by
        data fitting: AMR and decelerating moment release. We compare
        the apparent AMR found in the real data to the apparent AMR
        found in four types of synthetic catalogs with no inherent AMR.
        When spatiotemporal clustering is included in the simulations,
        similar AMR signals are found by data fitting in both the real
        and synthetic data sets even though the synthetic data sets
        contain no real AMR. These tests demonstrate that apparent AMR
        may arise from a combination of data fitting and normal foreshock
        and aftershock activity. In principle, data-fitting artifacts
        could be avoided if the free parameters were determined from
        scaling relationships between the duration and spatial extent
        of the AMR pattern and the magnitude of the earthquake that
        follows it. However, we demonstrate that previously proposed
        scaling relationships are unstable, statistical artifacts caused
        by the use of a minimum magnitude for the earthquake catalog
        that scales with the main shock magnitude. Some recent AMR
        studies have used spatial regions based on hypothetical stress
        loading patterns, rather than circles, to select the data. We
        show that previous tests were biased and that unbiased tests do
        not find this change to the method to be an improvement. The use
        of declustered catalogs has also been proposed to eliminate the
        effect of clustering but we demonstrate that this does not
        increase the statistical significance of AMR. Given the ease
        with which data fitting can find desired patterns in seismicity,
        future studies of AMR-like observations must include complete
        tests against synthetic catalogs that include spatiotemporal
        clustering.",
            "keywords": null,
            "refs": null,
```

```
        "journal": "Journal of Geophysical Research",
        "year": 2008,
        "doi": "10.1029/2007JB005410"
},
{
        "refID": "2008-1_Mignan-A_Tectonophys",
        "title": "Non-Critical Precursory Accelerating Seismicity Theory (NC PAST) and limits of the power-law !t methodology",
        "authors": ["Mignan-A"],
        "abstract": "The hypothesis that Accelerating Moment Release (AMR) is a precursor to large earthquakes is still debated. On one hand, AMR has been claimed to be observed in many cases and on the other hand, it has been proposed that apparent AMR is only due to data-fitting. The debate is in general focused on the validity of the c-value (curvature parameter), which permits to quantify AMR (i.e. cumulative Benioff strain through time), or more generally Precursory Accelerating Seismicity (PAS, i.e. cumulative number of events through time). Contrary to previous studies, which compare c-value optimization in real seismicity catalogues and in random synthetic catalogues, I test c-value optimization in theoretical synthetic catalogues. In that particular case, I assume that PAS exists and that it can be explained by the Non-Critical Precursory Accelerating Seismicity Theory (NC PAST). This theory demonstrates that PAS can emerge from the background seismicity because of the decrease, due to loading, of the size of a stress shadow due to a previous earthquake. I improve the NC PAST by integrating the following characteristics of the background seismicity, (1) the density of random events outside the stress shadow δb0 and (2) the noise ratio δb−/δb0, with δb− being the density of random events inside the stress shadow. Then I perform a spatiotemporal search of PAS using the power-law fit methodology (i.e. c-value) and compare the optimal signal to the expected spatiotemporal extent of the theoretical signal. First I show that the optimal starting time and spatial extent of PAS are poorly controlled, due in part to the intrinsic properties of the c-value, but also to the random behavior of background seismicity. Second I show that theoretical PAS is identified by an optimal c-value (clear acceleration) only if the regional seismic activity (~ δb0) is high and the noise ratio (δb−/δb0) is low. Otherwise the signal tends to disappear and the c-value becomes unstable. As a consequence, even if the power-law fit methodology is a simple approach to test the presence of PAS and can help provide a better understanding of the process engaged, it seems inadequate for robust systematic prospective forecasts.",
        "keywords": ["Earthquake Forecast", "Precursory seismicity", "AMR", "Stress loading", "Stress shadow", "Synthetic catalogue"],
        "refs": null,
        "journal": "Tectonophysics",
        "year": 2008,
```

```
            "doi": "10.1016/j.tecto.2008.02.010"
    },
    {
            "refID": "2008-1_Mignan-A_AdvGeophys",
            "title": "The Stress Accumulation Model: Accelerating Moment Release and Seismic Hazard",
            "authors": ["Mignan-A"],
            "abstract": "This chapter shows that the Stress Accumulation model can be defined with simple stress transfer considerations. At present, a parallel can be made between accelerating moment release and aftershocks, which seem symmetrical about the mainshock time (acceleration/decceleration). Aftershocks are located in positive stress lobes of the stress field caused by the mainshock whereas accelerating moment release is located in positive stress lobes of the prestress field of the future mainshock. Aftershocks are due to the change of stress at the time of the mainshock and accelerating moment release is due to the change of stress during loading before the time of the mainshock. This view permits to better understand the possible behavior of precursory seismicity by defining a simple pattern. This is in contrast with critical concepts where patterns emerge from chaos and are difficult to catch. This has important consequences in earthquake forecasting. Indeed many precursors have been proposed, based on different mathematical or statistical tools around the criticality concepts. However, the application of the Stress Accumulation model to earthquake forecasting also shows that a systematic use of accelerating moment release is complex and depends of the studied region. It shows that proper statistics are necessary to determine correctly the reliability of precursory patterns.",
            "keywords": null,
            "refs": null,
            "journal": "Advances in Geophysics",
            "year": 2008,
            "doi": "10.1016/S0065-2687(07)49002-1"
    },
    {
            "refID": "2008-1_Mignan-A_GRL",
            "title": "Relationship between accelerating seismicity and quiescence, two precursors to large earthquakes",
            "authors": ["Mignan-A", "DiGiovambattista-R"],
            "abstract": "The Non-Critical Precursory Accelerating Seismicity Theory (PAST) has been proposed recently to explain the formation of accelerating seismicity (increase of the a-value) observed before large earthquakes. In particular, it predicts that precursory accelerating seismicity should occur in the same spatiotemporal window as quiescence. In this first combined study we start by determining the spatiotemporal extent of quiescence observed prior to the 1997 Mw = 6 Umbria-Marche earthquake, Italy, using the RTL (Region-Time-Length) algorithm. We then show that background events located in that
```

```
spatiotemporal window form a clear acceleration, as expected by
the Non-Critical PAST. This result is a step forward in the
understanding of precursory seismicity by relating two of the
principal patterns that can precede large earthquakes.",
    "keywords": null,
    "refs": null,
    "journal": "Geophysical Research Letters",
    "year": 2008,
    "doi": "10.1029/2008GL035024"
},
{
    "refID": "2008-1_Papadimitriou-P_JGR",
    "title": "Identification of seismic precursors before large
earthquakes: Decelerating and accelerating seismic patterns",
    "authors": ["Papadimitriou-P"],
    "abstract": "A useful way of understanding both
seismotectonic processes and earthquake prediction research is
to conceive seismic patterns as a function of space and time.
The present work investigates seismic precursors before the
occurrence of an earthquake. It does so by means of a methodology
designed to study spatiotemporal characteristics of seismicity
in a selected area. This methodology is based on two phenomena:
the decelerating moment release (DMR) and the accelerating
moment release (AMR), as they occur within a period ranging from
several months to a few years before the oncoming event. The
combination of these two seismic sequences leads to the proposed
decelerating-accelerating moment release (DAMR) earthquake
sequence, which appears as the last stage of loading in the
earthquake cycle. This seismic activity appears as a foreshock
sequence and can be supported by the stress accumulation model
(SAM). The DAMR earthquake sequence constitutes a double seismic
precursor identified in space and time before the occurrence of
an earthquake and can be used to improve seismic hazard
assessment research. In this study, the developed methodology
is applied to the data of the 1989 Loma Prieta (California), the
1995 Kobe (Japan), and the 2003 Lefkada (Greece) earthquakes.
The last part of this study focuses on the application of the
methodology to the Ionian Sea (western Greece) and forecasts two
earthquakes in that area.",
    "keywords": null,
    "refs": null,
    "journal": "Journal of Geophysical Research",
    "year": 2008,
    "doi": "10.1029/2007JB005112"
},
{
    "refID": "2009-1_Chouliaras-G_NHESS",
    "title": "Seismicity anomalies prior to 8 June 2008, Mw=6.4
earthquake in Western Greece",
    "authors": ["Chouliaras-G"],
    "abstract": "The epicentral area of the Mw=6.4, 8 June 2008
main shock in northwestern Peloponesus, Western Greece, had been
```

```
forecasted as a candidate for the occurrence of a strong
earthquake by independent scientific investigations. This study
concerns the seismicity of a large area surrounding the
epicenter of the main shock using the seismological data from
the monthly bulletins of the Institute of Geodynamics of the
National Observatory of Athens. This data set is the most
detailed earthquake catalog available for anomalous seismicity
pattern investigations in Greece. The results indicate a
decrease in seismicity rate seven years prior to the 8 June main
shock which constituted a two and a half year long seismic
quiescence surrounding the epicentral area. This quiescence
anomaly was succeeded by a period of acceleration in seismic
activity for five years approximately, until the occurrence of
the main shock.",
        "keywords": null,
        "refs": null,
        "journal": "Natural Hazards and Earth System Science",
        "year": 2009,
        "doi": "10.5194/nhess-9-327-2009"
},
{
        "refID": "2009-2_Chouliaras-G_NHESS",
        "title": "Seismicity anomalies prior to the 13 December
2008, Ms=5.7 earthquake in Central Greece",
        "authors": ["Chouliaras-G"],
        "abstract": "This investigation has applied a recent
methodology to identify seismic quiescence and seismic
acceleration, prior to the occurrence of the 13 December 2008,
Ms=5.7 earthquake in Central Greece. Anomalous seismic
quiescence is observed around the epicentral area almost twelve
years prior to the main shock and it lasted for a period of
about four and a half years. After this period an acceleration
in seismic activity began and lasted until the main shock.
Modeling this seismic sequence with the time-to-failure equation
and with a fixed value of the exponent 'm' equal to 0.32, shows
a successful estimation of the occurrence time of the main event
within a few days. The physical meaning of this particular choice
of the 'm' value is discussed.",
        "keywords": null,
        "refs": null,
        "journal": "Natural Hazards and Earth System Science",
        "year": 2009,
        "doi": "10.5194/nhess-9-501-2009"
},
{
        "refID": "2009-1_Greenhough-J_GRL",
        "title": "Comment on 'Relationship between accelerating
seismicity and quiescence, two precursors to large earthquakes'
by Arnaud Mignan and Rita Di Giovambattista",
        "authors": ["Greenhough-J", "Bell-AF", "Main-IG"],
        "abstract": "The significance levels of many reported
episodes of Accelerating Moment Release (AMR, a cumulative func-
```

```
tion of earthquake magnitude with time) have been shown to be
too low to reject a range of alternative hypotheses [Hardebeck
et al., 2008]. While Mignan [2008] acknowl- edges the deficiency
of power-law fitting alone for fore- casting large events via
AMR, this and the proceeding study [Mignan and Di
Giovambattista, 2008] do not address an underlying problem of
applying standard regression methods to cumulative data. We
consider this a timely opportunity to emphasize why regression
on any cumulative quantity requires the utmost care and is at
best avoided. This cautionary comment is relevant to a wide
range of applications in geophysics and elsewhere.",
        "keywords": null,
        "refs": null,
        "journal": "Geophysical Research Letters",
        "year": 2009,
        "doi": "10.1029/2009GL039846"
},
{
        "refID": "2009-1_Mignan-A_GRL",
        "title": "Reply to comment by J. Greenhough et al. on
'Relationship between accelerating seismicity and quiescence,
two precursors to large earthquakes'",
        "authors": ["Mignan-A", "DiGiovambattista-R"],
        "abstract": "The existence of Accelerating Seismic Release
(ASR) prior to large earthquakes has been largely debated in the
last decade and opponents are even stronger since the landmark
paper by Hardebeck et al. [2008] (personal communication from
StatSeis 2007, 2008 and 2009 meet- ings attendees). In this
reply, we discuss the limits of approaches employed to verify
the 'non-existence' of ASR and we emphasize the advantages of
the method proposed by Mignan and Di Giovambattista [2008] to
identify ASR, over the classic ASR regression method [e.g.,
Bowman et al., 1998].",
        "keywords": null,
        "refs": null,
        "journal": "Geophysical Research Letters",
        "year": 2009,
        "doi": "10.1029/2009GL039871"
},
{
        "refID": "2010-1_Bebbington-MS_PAGEOPH",
        "title": "Repeated Intermittent Earthquake Cycles in the
San Francisco Bay Region",
        "authors": ["Bebbington-MS", "Harte-DS", "Jaume-SC"],
        "abstract": "Forecasts of future earthquake hazard in the
San Francisco Bay region (SFBR) are dependent on the
distribution used for the possible magnitude of future events.
Based on the limited observed data, it is not possible to
statistically distinguish between many distributions with very
different tail behavior. These include the modified and
truncated Gutenberg–Richter distributions, and a composite
distribution assembled by the Working Group on California
```

Earthquake Probabilities. There is consequent ambiguity in the estimated probability of very large, and hence damaging, events. A related question is whether the energy released in earthquakes is a small or large proportion of the stored energy in the crust, corresponding loosely to the ideas of self-organized criticality, and intermittent criticality, respectively. However, the SFBR has experienced three observed accelerating moment release (AMR) cycles, terminating in the 1868 Hayward, 1906 San Andreas and 1989 Loma Prieta events. A simple stochastic model based on elastic rebound has been shown to be capable of producing repeated AMR cycles in large synthetic catalogs. We propose that such catalogs can provide the basis of a test of a given magnitude distribution, via comparisons between the AMR properties of the real and synthetic data. Our results show that the truncated Gutenberg—Richter distribution produces AMR behavior closest to the observed AMR behavior. The proviso is that the magnitude parameters b and m max are such that a sequence of large events that suppresses activity for several centuries is unlikely to occur. Repeated simulation from the stochastic model using such distributions produces 30-year hazard estimates at various magnitudes, which are compared with the estimates from the 2003 Working Group on California Earthquake Probabilities.",
        "keywords": ["Earthquake magnitude", "accelerating moment release", "truncated Gutenberg—Richter", "tail behavior", "hazard estimates"],
        "refs": null,
        "journal": "Pure and Applied Geophysics",
        "year": 2010,
        "doi": "10.1007/s00024-010-0064-6"
},
{
        "refID": "2010-1_Jiang-CS_ConcurrComput",
        "title": "Seismic moment release before the May 12, 2008, Wenchuan earthquake in Sichuan of southwest China",
        "authors": ["Jiang-CS", "Wu-ZL"],
        "abstract": "Whether seismic moment release before great earthquakes exhibits accelerating or quiescence behavior is one of the controversial topics in the study of intermediate-term earthquake forecast or time-dependent seismic hazard. The May 12, 2008, Wenchuan earthquake provides a unique opportunity to check whether accelerating moment release (AMR) or quiescence existed before this great earthquake. To systematically analyze the precursory moment release, considering the special characteristics of this great inland thrust event, we took four upgraded approaches using the local earthquake catalogue with cutoff magnitude ML3.0. We propose a BIC criterion as a development of the curvature parameter q to identify the statistically significant acceleration or quiescence behavior as compared with linear increase. We use an 'eclipse method' as a development of the 'interfering event consideration' to eliminate the interference from the nearby seismically active

```
        fault zones. We consider the distribution of m-coefficient in
        the (T, R, Mc) space, to explore the variation of moment release
        behavior with temporal window length T and spatial window radius
        R centered at the nucleation point, and cutoff magnitude Mc of
        the catalogue in use. We use not only circular windows but also
        'crack-like' windows to investigate the overall behavior of the
        moment release associated with the Wenchuan earthquake. Through
        this retrospective case study, it is observed that moment
        release before the Wenchuan earthquake did show accelerating
        moment release (AMR) and quiescence behavior for different
        spatio-temporal ranges, with robustness to some extent against
        the changing of parameters, indicating the preparation process
        of this great earthquake. However, for this earthquake, to
        constrain the failure time in a forward forecasting mode is
        shown to be difficult.",
        "keywords": null,
        "refs": null,
        "journal": "Concurrency and Computation",
        "year": 2010,
        "doi": "10.1002/cpe.1522"
},
{
        "refID": "2010-1_Mignan-A_Tectonophys",
        "title": "Testing the Pattern Informatics index on
synthetic seismicity catalogs based on the Non-Critical PAST",
        "authors": ["Mignan-A", "Tiampo-K"],
        "abstract": "The Non-Critical Precursory Accelerating
Seismicity Theory (or Non-Critical PAST) has recently been
proposed to explain the formation of accelerating seismicity
that may be observed before large earthquakes. It has led to the
possibility of constructing synthetic seismicity catalogs where
patterns of accelerating seismicity (~ activation) and
quiescence, which occur in the same space–time window, can be
evaluated by existing forecasting techniques. In this study, the
Pattern Informatics (or PI) index is tested on synthetic
catalogs where a realistic spatiotemporal clustering has been
added on top of the theoretical precursory seismicity. We show
that the PI index is successful at identifying the precursory
quiescent signal but fails in identifying precursory
accelerating seismicity directly, being more sensitive to
aftershock sequences of background events than to the
activation-like behavior of the acceleration, formed by
background events alone. We also show that the PI index has a
high success rate in finding precursory quiescence, even for a
low signal-to-noise ratio. As for the few false negatives, they
are usually due to interfering aftershocks as well, which skew
seismicity rates to higher means. The Non-Critical PAST, by
helping to better understand the behavior of the PI algorithm
in synthetic catalogs, gives new perspectives on how to improve
it and on how to use it in real catalogs.",
        "keywords": ["AMR", "Pattern informatics", "Earthquake
simulator", "Power-law"],
```

```
        "refs": null,
        "journal": "Tectonophysics",
        "year": 2010,
        "doi": "10.1016/j.tecto.2009.10.023"
},
{
        "refID": "2010-1_Papazachos-BC_JOSE",
        "title": "Present patterns of decelerating—accelerating seismic strain in South Japan",
        "authors": ["Papazachos-BC", "Karakaisis-GF", "Scordilis-EM", "Papazachos-CB", "Panagiotopoulos-DG"],
        "abstract": "Decelerating generation of preshocks in a narrow (seismogenic) region and accelerating generation of other preshocks in a broader (critical) region, called decelerating—accelerating seismic strain (D-AS) model has been proposed as appropriate for intermediate-term earthquake prediction. An attempt is made in the present work to identify such seismic strain patterns and estimate the corresponding probably ensuing large mainshocks (M ≥ 7.0) in south Japan (30—38° N, 130—138° E). Two such patterns have been identified and the origin time, magnitude, and epicenter coordinates for each of the two corresponding probably ensuing mainshocks have been estimated. Model uncertainties of predicted quantities are also given to allow an objective forward testing of the efficiency of the model for intermediate-term earthquake prediction.",
        "keywords": ["Decelerating seismic strain", "Accelerating seismic strain", "Earthquake prediction", "Japan"],
        "refs": null,
        "journal": "Journal of Seismology",
        "year": 2010,
        "doi": "10.1007/s10950-009-9165-z"
},
{
        "refID": "2010-1_DeSantis-A_Tectonophys",
        "title": "The 2009 L'Aquila (Central Italy) seismic sequence as a chaotic process",
        "authors": ["DeSantis-A", "Cianchini-G", "Qamili-E", "Frepoli-A"],
        "abstract": "In this paper we demonstrate that the seismic sequence of foreshocks culminating with the recent Mw = 6.3 main shock on April 6, 2009 in L'Aquila (Central Italy) evolved as a chaotic process. To do this, we apply a nonlinear retrospective prediction to this seismic sequence and look at the temporal behaviour of the error between predicted and actual occurrence of the main shock when gradually increasing parts of the sequence are considered. This is a generalisation of the typical nonlinear approach which is quite powerful to detect chaos in relatively short time series. The method of prediction is based on the Accelerated Strain Release (ASR) analysis in time and on the nonlinear forecasting approach in a reconstructed phase space. We find that i) the temporal decay of the prediction error is consistent with an exponential function with a time
```

```
constant τ of about 10 days and ii) at around 6 days before the
main shock, ASR analysis is quite powerful for anticipating the
time of occurrence with an uncertainty of about a day. Due to
its retrospective characteristics, the latter result could be
affected by changes on some a-priori parameters used in the
application of the ASR technique. However, we consider these
findings, together with those obtained from the phase-space
analysis, to be strong evidence that the studied sequence of
foreshocks was produced by a physical process dominated by a
significant chaotic component characterised by a K-entropy = 1/τ
of about 0.1 day−1. This result could have important
implications for the predictability of the possible main shock
for those seismic sequences showing analogous nonlinear chaotic
properties.",
        "keywords": ["Seismic sequence", "Chaotic process",
"Accelerated Strain Release", "Nonlinear prediction", "Phase-
space reconstruction"],
        "refs": null,
        "journal": "Tectonophysics",
        "year": 2010,
        "doi": "10.1016/j.tecto.2010.10.005"
},
{
        "refID": "2012-1_Jiang-CS_EPS",
        "title": "Insights into the long-to-intermediate-term pre-
shock accelerating moment release (AMR) from the March 11, 2011,
off the Pacific coast of Tohoku, Japan, M 9 earthquake",
        "authors": ["Jiang-CS", "Wu-ZL"],
        "abstract": "Great earthquakes with extending rupture
areas, such as the March 11, 2011, off the Pacific coast of
Tohoku, Japan, M 9 earthquake, provide opportunities to inspect
some details of the pre-shock moment release with the reference
of the earthquake preparation and rupture processes. To this
end, we investigated the cumulative seismic moment tensor for
different segments of the earthquake fault. For the 3 decades
time scale, pre-shock accelerating moment release (AMR) can be
observed, with potential correlation with the segmentation of
earthquake fault.",
        "keywords": ["Accelerating moment release", "Tohoku M 9
earthquake", "fault segmentation"],
        "refs": null,
        "journal": "Earth, Planets and Space",
        "year": 2012,
        "doi": null
},
{
        "refID": "2012-1_Lagios-E_Tectonophys",
        "title": "Combined Seismicity Pattern Analysis, DGPS and
PSInSAR studies in the broader area of Cephalonia (Greece)",
        "authors": ["Lagios-E", "Papadimitriou-P", "Novali-F",
"Sakka-V", "Fumagalli-A", "Vlachou-K", "DelConte-S"],
```

```
        "abstract":    "Ground    deformation    studies    based    on
Differential GPS (DGPS) measurements and Permanent Scatterer
Interferometric (PSI)  analysis  have  been  conducted  on  the
islands of Cephalonia and Ithaca covering the period 1992 to
2010. DGPS measurements for the period 2001 to 2010 revealed
horizontal  clockwise  rotation  of  Cephalonia  and  velocities
ranging from 3 to 8 mm/yr with the largest values occurring at
the western and southern parts of the island. Considering its
vertical deformation, two periods are distinguished on the basis
of DGPS and PSI: The first one (1992 to 2003) shows generally
an almost linear slight subsidence (around 1 mm/yr) which is
consistent with expected neotectonic movements of the island.
The second one (2003 to 2010) has been tentatively attributed
to dilatancy in which reversal to uplift (2—4 mm/yr) occurred
mainly along the southern and southeastern parts of the island,
while larger magnitudes (>4 mm/yr) took place at the western
part. These non-linear high rates of uplift started at about
mid-2005, and were of increasing rate at the southern part, but
of decreasing rate at the western part; they may indicate a
major regional crustal deformation  process in an environment
that  has  previously  supported  offshore  large  magnitude
earthquakes. Parallel analysis of the observed seismicity in the
broader  area  identified  two  seismically  critical  areas  on  the
basis of the decelerating—accelerating seismicity: a major one
south of Cephalonia and west of Zakynthos, and another minor one
at the NW part of Peloponnese. Critical time estimates of the
occurrence of a future strong seismic event in the above critical
areas were also made based on: (i) accelerating seismicity, and
(ii) the temporal analysis of the seismicity.",
        "keywords":    ["Differential    GPS",    "PS    Interferometry",
"Seismicity Pattern Analysis", "Cephalonia", "Ionian Islands"],
        "refs": null,
        "journal": "Tectonophysics",
        "year": 2012,
        "doi": "10.1016/j.tecto.2011.12.015"
    },
    {
        "refID": "2012-1_Mignan-A_GRL",
        "title":  "Seismicity  precursors  to  large  earthquakes
unified in a stress accumulation framework",
        "authors": ["Mignan-A"],
        "abstract": "Various  seismicity  patterns  before  large
earthquakes have been reported in the literature. They include
foreshocks    (medium-term    acceleration    and    short-term
activation), quiescence, doughnut patterns and event migration.
The existence of these precursory patterns is however debated.
Here, we develop an approach based on the concept of stress
accumulation  to  unify  and  categorize  all  claimed  seismic
precursors in a same physical framework. We first extend the
Non-Critical Precursory Accelerating Seismicity Theory (N-C
PAST), which already explains most precursors, to additionally
include short-term activation. Theoretical results are then
```

```
        compared to the time series observed prior to the 2009Mw = 6.3
        L'Aquila, Italy, earthquake. We finally show that different
        precursory paths are possible before large earthquakes, with
        possible coupling of different patterns or non-occurrence of
        any. This is described by a logic tree defined from the combined
        probabilities of occurrence of the mainshock at a given stress
        state and of precursory silent slip on the fault. In the case
        of the L'Aquila earthquake, the observed precursory path is
        coupling of quiescence and accelerating seismic release,
        followed by activation. These results provide guidelines for
        future research on earthquake predictability.",
        "keywords": null,
        "refs": null,
        "journal": "Geophysical Research Letters",
        "year": 2012,
        "doi": "10.1029/2012GL053946 "
},
{
        "refID": "2012-1_Pliakis-D_AnnGeophys",
        "title": "A first principles approach to understand the
physics of precursory accelerating seismicity",
        "authors": ["Pliakis-D", "Papakostas-T", "Vallianatos-F"],
        "abstract": "Observational studies from rock fractures to
earthquakes indicate that fractures and many large earthquakes
are preceded by accelerating seismic release rates (accelerated
seismic deformation). This is characterized by cumulative
Benioff strain that follows a power law time-to-failure relation
of the form C(t) = K + A(Tf − t)m, where Tf is the failure time
of the large event, and m is of the order of 0.2-0.4. More recent
theoretical studies have been related to the behavior of
seismicity prior to large earthquakes, to the excitation in
proximity of a spinodal instability. These have show that the
power-law activation associated with the spinodal instability
is essentially identical to the power-law acceleration of
Benioff strain observed prior to earthquakes with m = 0.25-0.3.
In the present study, we provide an estimate of the generic
local distribution of cracks, following the Wackentrapp-
Hergarten-Neugebauer model for mode I propagation and
concentration of microcracks in brittle solids due to remote
stress. This is a coupled system that combines the equilibrium
equation for the stress tensor with an evolution equation for
the crack density integral. This inverse type result is obtained
through the equilibrium equations for a solid body. We test
models for the local distribution of cracks, with estimation of
the stress tensor in terms of the crack density integral, through
the Nash-Moser iterative method. Here, via the evolution
equation, these estimates imply that the crack density integral
grows according to a (Tf − t)0.3-law, in agreement with
observations.",
        "keywords": ["seismicity", "fracture", "Benioff"],
        "refs": null,
        "journal": "Annals of Geophysics",
```

```
        "year": 2012,
        "doi": "10.4401/ag-5363   "
},
{
        "refID": "2013-1_Bouchon-M_NatureGeo",
        "title": "The long precursory phase of most large interplate earthquakes",
        "authors": ["Bouchon-M", "Durand-V", "Marsan-D", "Karabulut-H", "Schmittbuhl-J"],
        "abstract": "Many earthquakes are preceded by foreshocks. However, the mechanisms that generate foreshocks and the reason why they occur before some earthquakes and not others are unknown. Here we use seismic catalogues from the best instrumented areas of the North Pacific to analyse the foreshock sequences preceding all earthquakes there between 1999 and 2011, of magnitude larger than 6.5 and at depths shallower than 50 km. The data set comprises 31 earthquakes at plate boundaries, and 31 in plate interiors. We find that there is a remarkable contrast between the foreshock sequences of interplate compared with intraplate earthquakes. Most large earthquakes at plate interfaces in the North Pacific were preceded by accelerating seismic activity in the months to days leading up to the mainshock. In contrast, foreshocks are much less frequent in intraplate settings. We suggest that at plate boundaries, the interface between the two plates begins to slowly slip before the interface ruptures in a large earthquake. This relatively long precursory phase could help mitigate earthquake risk at plate boundaries.",
        "keywords": null,
        "refs": null,
        "journal": "Nature Geosciences",
        "year": 2013,
        "doi": "10.1038/NGEO1770"
},
{
        "refID": "2013-1_Guilhem-A_GJI",
        "title": "Testing the accelerating moment release (AMR) hypothesis in areas of high stress",
        "authors": ["Guilhem-A", "Buergmann-R", "Freed-AM", "Ali-ST"],
        "abstract": "Several retrospective analyses have proposed that significant increases in moment release occurred prior to many large earthquakes of recent times. However, the finding of accelerating moment release (AMR) strongly depends on the choice of three parameters: (1) magnitude range, (2) area being considered surrounding the events and (3) the time period prior to the large earthquakes. Consequently, the AMR analysis has been criticized as being a posteriori data-fitting exercise with no new predictive power. As AMR has been hypothesized to relate to changes in the state of stress around the eventual epicentre, we compare here AMR results to models of stress accumulation in California. Instead of assuming a complete stress drop on all
```

surrounding fault segments implied by a back-slip stress lobe method, we consider that stress evolves dynamically, punctuated by the occurrence of earthquakes, and governed by the elastic and viscous properties of the lithosphere. We study the seismicity of southern California and extract events for AMR calculations following the systematic approach employed in previous studies. We present several sensitivity tests of the method, as well as grid-search analyses over the region between 1955 and 2005 using fixed magnitude range, radius of the search area and period of time. The results are compared to the occurrence of large events and to maps of Coulomb stress changes. The Coulomb stress maps are compiled using the coseismic stress from all M > 7.0 earthquakes since 1812, their subsequent post-seismic relaxation, and the interseismic strain accumulation. We find no convincing correlation of seismicity rate changes in recent decades with areas of high stress that would support the AMR hypothesis. Furthermore, this indicates limited utility for practical earthquake hazard analysis in southern California, and possibly other regions.",
        "keywords": ["Earthquake interaction", "forecasting, and prediction", "Seismicity and tectonics", "Fractures and faults", "Crustal structure"],
        "refs": null,
        "journal": "Geophysical Journal International",
        "year": 2013,
        "doi": "10.1093/gji/ggt298"
},
{
        "refID": "2013-1_Jiang-CS_PAGEOPH",
        "title": "Intermediate-Term Medium-Range Precursory Accelerating Seismicity Prior to the 12 May 2008, Wenchuan Earthquake",
        "authors": ["Jiang-CS", "Wu-ZL"],
        "abstract": "In the study of the predictability of great earthquakes in the perspective of seismicity analysis, two issues are presently controversial, and need more detailed studies based on real earthquake cases. The first issue is whether there exists pre-shock accelerating behavior of seismicity which is robust against the changing of spatio-temporal ranges for the sampling of seismic events, and the second is whether such an accelerating behavior is physically associated with an approach to the critical point. To answer these two questions, a retrospective case study was conducted on the 12 May 2008, Wenchuan earthquake, using the local earthquake catalogue in Sichuan and Yunnan Provinces, China, with cutoff magnitude M L3.0, from 1977 to 2008. The results show that the answer to the first question appears to be 'yes'; that is, in a finite spatial domain within the last couple of years before the event, clear accelerating seismicity could be observed. The answer to the second question cannot be obtained merely by examining seismicity data. However, detailed analysis of the accelerating behavior reveals a potential spatial

```
correlation between the accelerating region and a known
asperity, which might be an evidence for that the observed
acceleration may have a geometrical or mechanical rather than
statistical origin.",
        "keywords": ["Precursory accelerating seismicity", "2008
Wenchuan earthquake", "asperity", "critical point model of
earthquakes", "intermediate-term medium-range earthquake
forecast"],
        "refs": null,
        "journal": "Pure and Applied Geophysics",
        "year": 2013,
        "doi": "10.1007/s00024-011-0413-0"
    },
    {
        "refID": "2013-1_Karakaisis-GF_JOSE",
        "title": "Recent reliable observations and improved tests
on synthetic catalogs with spatiotemporal clustering verify
precursory decelerating—accelerating seismicity",
        "authors": ["Karakaisis-GF", "Papazachos-CB", "Scordilis-
EM"],
        "abstract": "We examined the seismic activity which
preceded six strong mainshocks that occurred in the Aegean
(M = 6.4—6.9, 33—43° N, 19—28° E) and two strong mainshocks that
occurred in California (M = 6.5—7.1, 32—41° N, 115—125° W) during
1995—2010. We find that each of these eight mainshocks has been
preceded by a pronounced decelerating and an equally easily
identifiable accelerating seismic sequence with the time to the
mainshock. The two preshock sequences of each mainshock occurred
in separate space, time, and magnitude windows. In all eight
cases, very low decelerating seismicity, as well as very low
accelerating seismicity, is observed around the actual epicenter
of the ensuing mainshock. Statistical tests on the observed
measures of decelerating, q d, and accelerating, q a, seismicity
against similar measures calculated using synthetic catalogs
with spatiotemporal clustering based on the ETAS model show that
there is an almost zero probability for each one of the two
preshock sequences which preceded each of the eight mainshocks
to be random. These results support the notion that every strong
shallow mainshock is preceded by a decelerating and an
accelerating seismic sequence with predictive properties for the
ensuing mainshock.",
        "keywords": ["Precursory accelerating seismicity",
"Precursory decelerating seismicity"],
        "refs": null,
        "journal": "Journal of Seismology",
        "year": 2013,
        "doi": "10.1007/s10950-013-9372-5"
    },
    {
        "refID": "2015-1_Bouchon-M_NatureGeo",
        "title": "Reply to 'Artificial seismic acceleration'",
```

```
        "authors":     ["Bouchon-M",    "Durand-V",    "Marsan-D",
"Karabulut-H", "Schmittbuhl-J"],
        "abstract": "In our study1, we show that most large
magnitude M ≥ 6.5 interplate earthquakes are preceded by an
acceleration of seismic activity. The Correspondence from Felzer
et al. questions our interpretation of this acceleration. It has
long been recognized that one characteristic of seismic events
is their natural tendency to cluster both in space and time, as
evidenced by the presence of aftershocks following an
earthquake. The debate raised by Felzer et al. is whether
foreshocks result only from this tendency to cluster, that is,
a first shock triggers others and eventually one of them triggers
a large earthquake by something akin to a random throw. Felzer
and colleagues advocate this interpretation. Alternatively,
foreshocks may indicate an underlying mechanical process, such
as slow fault slip, in which the foreshocks are simply the
seismically visible signature — an interpretation we claim our
observations favour.",
        "keywords": null,
        "refs": null,
        "journal": "Nature Geosciences",
        "year": 2015,
        "doi": null
    },
    {
        "refID": "2015-1_DeSantis-A_Tectonophys",
        "title": "Accelerating moment release revisited: Examples
of application to Italian seismic sequences",
        "authors":          ["DeSantis-A",         "Cianchini-G",
"DiGiovambattista-R"],
        "abstract": "From simple considerations we propose a
revision of the Accelerating Moment Release (AMR) methodology
for improving our knowledge of seismic sequences and then,
hopefully in a close future, to reach the capability of
predicting the main-shock location and occurrence with
sufficient accuracy. The proposed revision is based on the
introduction of a "reduced" Benioff strain for the earthquakes
of the seismic sequence where, for the same magnitude and after
a certain distance from the main-shock epicentre, the closer the
events the more they are weighted. In addition, we retain the
usual expressions proposed by the ordinary AMR method for the
estimation of the corresponding main-shock magnitude, although
this parameter is the weakest of the analysis. Then, we apply
the revised method to four case studies in Italy, three of which
are the most recent seismic sequences of the last 9 years
culminating with a shallow main-shock, and one is instead a
1995–1996 swarm with no significant main-shock. The application
of the R-AMR methodology provides the best results in detecting
the precursory seismic acceleration, when compared with those
found by ordinary AMR technique. We verify also the stability
of the results in space, applying the analysis to real data with
moving circles in a large area around each main-shock epicentre,
```

```
        and the efficiency of the revised technique in time, comparing
the results with those obtained when applying the same analysis
to simulated seismic sequences.",
        "keywords": ["Earthquake interaction", "Forecasting and
prediction", "Seismicity and tectonics", "Seismic attenuation",
"Seismic sequence", "Foreshocks"],
        "refs": null,
        "journal": "Tectonophysics",
        "year": 2015,
        "doi": "10.1016/j.tecto.2014.11.015"
},
{
        "refID": "2015-1_Felzer-KR_NatureGeo",
        "title": "Artificial seismic acceleration",
        "authors": ["Felzer-KR", "Page-MT", "Michael-AJ"],
        "abstract": "In their 2013 Letter, Bouchon et al. claim to
see a significant acceleration of seismicity before magnitude
≥6.5 mainshock earthquakes that occur in interplate regions, but
not before intraplate mainshocks. They suggest that this
accelerating seismicity reflects a preparatory process before
large plate- boundary earthquakes. We concur that their
interplate data set has significantly more foreshocks than their
intraplate data set; however, we disagree that the foreshocks
indicate a precursory phase that is predictive of large events
in particular. Acceleration of seismicity in stacked foreshock
sequences has been seen before and has been explained by the
cascade model, in which earthquakes occasionally trigger
aftershocks larger than themselves. In this model, the time lags
between the smaller mainshocks and larger aftershocks follow the
inverse power law common to all aftershock sequences, creating
an apparent acceleration when stacked (see Supplementary
Information).",
        "keywords": null,
        "refs": null,
        "journal": "Nature Geosciences",
        "year": 2015,
        "doi": null
},
{
        "refID": "2016-1_Christou-EV_ResGeophys",
        "title": "Time dependent seismicity along the western coast
of Canada",
        "authors": ["Christou-EV", "Karakaisis-GF", "Scordilis-
EM"],
        "abstract": "Decelerating generation of intermediate
magnitude earthquakes (preshocks) in a narrow region
(seismogenic region) and accelerating generation of relatively
larger such earthquakes in a broader region (critical region)
has been proposed as an appropriate model for intermediate-term
earthquake prediction. We examined the seismic activity which
preceded the Mw=7.7 (October 28, 2012) thrust event that
occurred off the west coast of Haida Gwaii, Canada (formerly the
```

```
        Queen   Charlotte   islands),   by   applying   the  decelerating-
        accelerating seismic strain model. We found that this mainshock
        was preceded by a pronounced accelerating seismic sequence with
        the time to the mainshock, as well as by an equally easily
        identifiable decelerating seismic sequence. Both precursory
        seismic  sequences  occurred  in  different  space,  time  and
        magnitude windows. The behavior of previous mainshocks that
        occurred close to the 2012 earthquake was also examined by the
        time and magnitude predictable regional model. An attempt was
        also made to identify such seismic strain patterns, which may
        also be related to the generation of strong mainshocks along the
        western coast of Canada.",
        "keywords": ["Time dependent seismicity", "Canada", "Haida
        Gwaii 2012 earthquake"],
        "refs": null,
        "journal": "Research in Geophysics",
        "year": 2016,
        "doi": "10.4081/rg.2016.5730"
},
{
        "refID": "2017-1_Adamaki-AK_PAGEOPH",
        "title":  "Precursory  Activity  Before  Larger  Events  in
        Greece Revealed by Aggregated Seismicity Data",
        "authors": ["Adamaki-AK", "Roberts-RG"],
        "abstract": "We investigate the seismicity rate behaviour
        in and around Greece during 2009, seeking significant changes
        in rate preceding larger events. For individual larger events
        it is difficult to clearly distinguish precursory rate changes
        from  other,  possibly  unrelated,  variations  in  seismicity.
        However, when we aggregate seismicity data occurring within a
        radius of 10 km and in a 50-day window prior to earthquakes
        with, e.g. magnitude ≥3.5, the resulting aggregated time series
        show a clearly increasing trend starting 2—3 weeks prior to the
        "mainshock" time. We apply statistical tests to investigate if
        the observed behaviour may be simply consistent with random
        (poissonian) variations, or, as some earlier studies suggest,
        with clustering in the sense that high activity rates at some
        time may imply increased rates later, and thus (randomly)
        greater probability of larger coming events than for periods of
        lower seismicity. In this case, rate increases have little
        useful predictive power. Using data from the entire catalogue,
        the aggregated rate changes before larger events are clearly and
        strongly statistically significant and cannot be explained by
        such clustering. To test this we choose events at random from
        the catalogue as potential "mainshocks". The events preceding
        the  randomly  chosen  earthquakes  show  less  pronounced  rate
        increases compared to the observed rate changes prior to larger
        events. Similar behaviour is observed in data sub-sets. However,
        statistical  confidence  decreases  for  geographical  subsets
        containing few "mainshocks" as it does when data are weighted
        such that "mainshocks" with many preceding events are strongly
        downweighted relative to those with fewer. The analyses suggest
```

```
        that genuine changes in aggregated rate do occur prior to larger
events and that this behaviour is not due to a small number of
mainshocks with many preceding events dominating the analysis.
It does not automatically follow that it will be possible to
routinely observe precursory changes prior to individual larger
events, but there is a possibility that this may be feasible,
e.g. with better data from more sensitive networks.",
        "keywords": ["Temporal seismicity patterns", "aggregated
data", "precursory activity", "Greece"],
        "refs": null,
        "journal": "Pure and Applied Geophysics",
        "year": 2017,
        "doi": "10.1007/s00024-017-1465-6"
},
{
        "refID": "2017-1_Kazemian-J_PAGEOPH",
        "title": "Temporal Variations of Seismic Parameters in
Tehran Region",
        "authors": ["Kazemian-J", "Hatami-MR"],
        "abstract": "The study of earthquake precursors can lead
to deliver an entirely measurable, time varying estimation of
the coming events which is strongly based on physical and
geological principals and fully responsive to any future
examination. The temporal variation of the seismic parameters
such as "b" values, in the Gutenberg–Richter formula logN = a –
bM, some details of the pre-shock accelerating moment release
(AMR) and the variation of the Thirumalai and Mountain (TM)
metric before the past big event is investigated in the vicinity
of the Tehran city. The temporal variation of the b-value in
some cases strongly supports the assumption that it has
potential to be used as a precursory signal. In addition,
analysis of the pre-shock AMR along with the inverse TM metric
signal in the region shows deviations from the background long-
term behavior prior to some of the bigger events in the studied
area. The aggregation of these signs suggests that a combination
of studied physical precursors has a potential which could be
employed in a medium term earthquake warnings.",
        "keywords": ["Earthquake precursors", "b-value variation",
"accelerated moment release", "inverse TM metric"],
        "refs": null,
        "journal": "Pure and Applied Geophysics",
        "year": 2017,
        "doi": "10.1007/s00024-017-1549-3"
},
{
        "refID": "2018-1_Huang-H_GRL",
        "title": "Slow Unlocking Processes Preceding the 2015 Mw
8.4 Illapel, Chile, Earthquake",
        "authors": ["Huang-H", "Meng-L"],
        "abstract": "On 16 September 2015, the Mw 8.4 Illapel
earthquake occurred in central Chile with no intense foreshock
sequences documented in the regional earthquake catalog. Here
```

```
we employ the matched-filter technique based on an enhanced
template data set of previously catalogued events. We perform a
continuous search over an ~4-year period before the Illapel
mainshock to recover the uncatalogued small events and repeating
earthquakes. Repeating earthquakes are found both to the north
and south of the mainshock rupture zone. To the south of the
rupture zone, the seismicity and repeater-inferred aseismic slip
progressively accelerate around the Illapel epicenter starting
from ~140 days before the mainshock. This may indicate an
unlocking process involving the interplay of seismic and
aseismic slip. The acceleration culminates in a M 5.3 event of
low-angle thrust mechanism, which occurred ~36 days before the
Mw 8.4 mainshock. It is then followed by a relative quiescence
in seismicity until the mainshock occurred. This quiescence
might correspond to an intermediate period of stable slip before
rupture initiation. In addition, to the north of the mainshock
rupture area, the last aseismic-slip episode occurs within ~175—
95 days before the mainshock and accumulates the largest amount
of slip in the observation period. The simultaneous occurrence
of aseismic-slip transients over a large area is consistent with
large-scale slow unlocking processes preceding the Illapel
mainshock.",
    "keywords": null,
    "refs": null,
    "journal": "Geophysical Research Letters",
    "year": 2018,
    "doi": "10.1029/2018GL077060"
}
]

Table S2.

[
{
    "refID": "1988-1_Papadopoulos-GA_Tectonophys",
    "label": "0"
},
{
    "refID": "1990-1_Sykes-LR_Nature",
    "label": "0"
},
{
    "refID": "1992-1_Jaume-SC_GRL",
    "label": "0"
},
{
    "refID": "1993-1_Bufe-CG_JGR",
    "label": "0"
},
{
    "refID": "1994-1_Bufe-CG_PAGEOPH",
```

```json
        "label": "0"
},
{
        "refID": "1995-1_Newman-WI_PRE",
        "label": "1"
},
{
        "refID": "1995-1_Sornette-D_JPhysIFrance",
        "label": "1"
},
{
        "refID": "1996-1_Jaume-SC_JGR",
        "label": "0"
},
{
        "refID": "1996-1_Saleur-H_JGR",
        "label": "1"
},
{
        "refID": "1996-1_Varnes-DJ_GJI",
        "label": "1"
},
{
        "refID": "1998-1_Bowman-DD_JGR",
        "label": "1"
},
{
        "refID": "1998-1_Brehm-DJ_BSSA",
        "label": "0"
},
{
        "refID": "1998-1_Gross-S_GJI",
        "label": "0"
},
{
        "refID": "1998-1_Huang-Y_EPL",
        "label": "1"
},
{
        "refID": "1999-1_Brehm-DJ_BSSA",
        "label": "0"
},
{
        "refID": "1999-1_Brehm-DJ_JOSE",
        "label": "0"
},
{
        "refID": "1999-1_Jaume-SC_PAGEOPH",
        "label": "1"
},
{
        "refID": "1999-1_Main-IG_GJI",
```

```
                "label": "1"
        },
        {
                "refID": "1999-1_Sammis-CG_PAGEOPH",
                "label": "1"
        },
        {
                "refID": "1999-1_Yang-WZ_ActaSeismSinica",
                "label": "0"
        },
        {
                "refID": "2000-1_Huang-Y_JGR",
                "label": "1"
        },
        {
                "refID": "2000-1_Jaume-SC_GeophysMonogrSer",
                "label": "1"
        },
        {
                "refID": "2000-1_Jaume-SC_PAGEOPH",
                "label": "1"
        },
        {
                "refID": "2000-1_Papazachos-BC_PAGEOPH",
                "label": "1"
        },
        {
                "refID": "2000-1_Robinson-R_GJI",
                "label": "1"
        },
        {
                "refID": "2000-1_Rundle-JB_PAGEOPH",
                "label": "1"
        },
        {
                "refID": "2001-1_Bowman-DD_CRAS",
                "label": "0"
        },
        {
                "refID": "2001-1_Bowman-DD_GRL",
                "label": "0"
        },
        {
                "refID": "2001-1_DiGiovambattista-R_Tectonophys",
                "label": "1"
        },
        {
                "refID": "2001-1_Papazachos-CB_AnnGeofis",
                "label": "1"
        },
        {
                "refID": "2001-1_VereJones-D_GJI",
```

```
                "label": "1"
        },
        {
                "refID": "2001-1_Yang-WZ_JGR",
                "label": "1"
        },
        {
                "refID": "2001-1_Zoeller-G_JGR",
                "label": "1"
        },
        {
                "refID": "2002-1_BenZion-Y_PAGEOPH",
                "label": "1"
        },
        {
                "refID": "2002-1_Karakaisis-GF_GJI",
                "label": "1"
        },
        {
                "refID": "2002-1_Papazachos-BC_JOSE",
                "label": "1"
        },
        {
                "refID": "2002-1_Papazachos-BC_Tectonophys",
                "label": "1"
        },
        {
                "refID": "2002-1_Papazachos-CB_BSSA",
                "label": "1"
        },
        {
                "refID": "2002-1_Sammis-CG_PNAS",
                "label": "1"
        },
        {
                "refID": "2002-1_Weatherley-D_PAGEOPH",
                "label": "1"
        },
        {
                "refID": "2002-1_Yin-XC_PAGEOPH",
                "label": "1"
        },
        {
                "refID": "2002-1_Zoeller-G_GRL",
                "label": "1"
        },
        {
                "refID": "2003-1_Chen-CC_GJI",
                "label": "1"
        },
        {
                "refID": "2003-1_Helmstetter-A_JGR",
```

```json
            "label": "1"
    },
    {
            "refID": "2003-1_Karakaisis-GF_GJI",
            "label": "1"
    },
    {
            "refID": "2003-1_Karakaisis-GF_PAGEOPH",
            "label": "1"
    },
    {
            "refID": "2003-1_King-GCP_JGR",
            "label": "0"
    },
    {
            "refID": "2003-1_Turcotte-DL_GJI",
            "label": "1"
    },
    {
            "refID": "2003-1_Tzanis-A_NHESS",
            "label": "1"
    },
    {
            "refID": "2004-1_Bowman-DD_PAGEOPH",
            "label": "1"
    },
    {
            "refID": "2004-1_Bufe-CG_BSSA",
            "label": "0"
    },
    {
            "refID": "2004-1_DiGiovambattista-R_Tectonophys",
            "label": "0"
    },
    {
            "refID": "2004-1_Jaume-SC_JGR",
            "label": "1"
    },
    {
            "refID": "2004-1_Karakaisis-GF_Tectonophys",
            "label": "1"
    },
    {
            "refID": "2004-1_Sammis-CG_PAGEOPH",
            "label": "0"
    },
    {
            "refID": "2004-1_Scordilis-EM_JOSE",
            "label": "1"
    },
    {
            "refID": "2004-1_Wang-JC_TAO",
```

```
                "label": "1"
        },
        {
                "refID": "2004-1_Wang-LY_ActaSeismSinica",
                "label": "1"
        },
        {
                "refID": "2004-1_Wang-Y_PAGEOPH",
                "label": "1"
        },
        {
                "refID": "2004-1_Weatherley-D_PAGEOPH",
                "label": "1"
        },
        {
                "refID": "2005-1_Jiang-CS_ActaSeismSinica",
                "label": "1"
        },
        {
                "refID": "2005-1_Jiang-CS_BSSA",
                "label": "0"
        },
        {
                "refID": "2005-1_Papazachos-CB_BSSA",
                "label": "1"
        },
        {
                "refID": "2005-1_Robinson-R_GRL",
                "label": "0"
        },
        {
                "refID": "2006-1_Jiang-CS_PAGEOPH",
                "label": "0"
        },
        {
                "refID": "2006-1_Mignan-A_EPSL",
                "label": "0"
        },
        {
                "refID": "2006-1_Mignan-A_JGR",
                "label": "0"
        },
        {
                "refID": "2006-1_Papazachos-BC_JOSE",
                "label": "1"
        },
        {
                "refID": "2006-1_Zhou-S_JGR",
                "label": "0"
        },
        {
                "refID": "2007-1_Mignan-A_JGR",
```

```
                "label": "0"
        },
        {
                "refID": "2007-1_Papazachos-BC_BSSA",
                "label": "1"
        },
        {
                "refID": "2007-1_Papazachos-BC_JGR",
                "label": "1"
        },
        {
                "refID": "2008-1_Hardebeck-JL_JGR",
                "label": "0"
        },
        {
                "refID": "2008-1_Mignan-A_AdvGeophys",
                "label": "0"
        },
        {
                "refID": "2008-1_Mignan-A_GRL",
                "label": "0"
        },
        {
                "refID": "2008-1_Mignan-A_Tectonophys",
                "label": "0"
        },
        {
                "refID": "2008-1_Papadimitriou-P_JGR",
                "label": "1"
        },
        {
                "refID": "2009-1_Chouliaras-G_NHESS",
                "label": "0"
        },
        {
                "refID": "2009-1_Greenhough-J_GRL",
                "label": "0"
        },
        {
                "refID": "2009-1_Mignan-A_GRL",
                "label": "0"
        },
        {
                "refID": "2009-2_Chouliaras-G_NHESS",
                "label": "0"
        },
        {
                "refID": "2010-1_Bebbington-MS_PAGEOPH",
                "label": "0"
        },
        {
                "refID": "2010-1_DeSantis-A_Tectonophys",
```

```
                "label": "1"
        },
        {
                "refID": "2010-1_Jiang-CS_ConcurrComput",
                "label": "0"
        },
        {
                "refID": "2010-1_Mignan-A_Tectonophys",
                "label": "0"
        },
        {
                "refID": "2010-1_Papazachos-BC_JOSE",
                "label": "0"
        },
        {
                "refID": "2012-1_Jiang-CS_EPS",
                "label": "0"
        },
        {
                "refID": "2012-1_Lagios-E_Tectonophys",
                "label": "0"
        },
        {
                "refID": "2012-1_Mignan-A_GRL",
                "label": "0"
        },
        {
                "refID": "2012-1_Pliakis-D_AnnGeophys",
                "label": "1"
        },
        {
                "refID": "2013-1_Bouchon-M_NatureGeo",
                "label": "0"
        },
        {
                "refID": "2013-1_Guilhem-A_GJI",
                "label": "0"
        },
        {
                "refID": "2013-1_Jiang-CS_PAGEOPH",
                "label": "0"
        },
        {
                "refID": "2013-1_Karakaisis-GF_JOSE",
                "label": "1"
        },
        {
                "refID": "2015-1_Bouchon-M_NatureGeo",
                "label": "0"
        },
        {
                "refID": "2015-1_DeSantis-A_Tectonophys",
```

```
        "label": "0"
},
{
        "refID": "2015-1_Felzer-KR_NatureGeo",
        "label": "1"
},
{
        "refID": "2016-1_Christou-EV_ResGeophys",
        "label": "0"
},
{
        "refID": "2017-1_Adamaki-AK_PAGEOPH",
        "label": "0"
},
{
        "refID": "2017-1_Kazemian-J_PAGEOPH",
        "label": "0"
},
{
        "refID": "2018-1_Huang-H_GRL",
        "label": "0"
}
]
```

**Table S3.**

```
[
{
        "refID": "1988-1_Papadopoulos-GA_Tectonophys",
        "label": "0"
},
{
        "refID": "1990-1_Sykes-LR_Nature",
        "label": "0"
},
{
        "refID": "1992-1_Jaume-SC_GRL",
        "label": "1"
},
{
        "refID": "1993-1_Bufe-CG_JGR",
        "label": "1"
},
{
        "refID": "1994-1_Bufe-CG_PAGEOPH",
        "label": "1"
},
{
        "refID": "1995-1_Newman-WI_PRE",
        "label": "3"
},
```

```json
    {
        "refID": "1995-1_Sornette-D_JPhysIFrance",
        "label": "3"
    },
    {
        "refID": "1996-1_Jaume-SC_JGR",
        "label": "0"
    },
    {
        "refID": "1996-1_Saleur-H_JGR",
        "label": "3"
    },
    {
        "refID": "1996-1_Varnes-DJ_GJI",
        "label": "2"
    },
    {
        "refID": "1998-1_Bowman-DD_JGR",
        "label": "2"
    },
    {
        "refID": "1998-1_Brehm-DJ_BSSA",
        "label": "1"
    },
    {
        "refID": "1998-1_Gross-S_GJI",
        "label": "1"
    },
    {
        "refID": "1998-1_Huang-Y_EPL",
        "label": "3"
    },
    {
        "refID": "1999-1_Brehm-DJ_BSSA",
        "label": "1"
    },
    {
        "refID": "1999-1_Brehm-DJ_JOSE",
        "label": "1"
    },
    {
        "refID": "1999-1_Jaume-SC_PAGEOPH",
        "label": "2"
    },
    {
        "refID": "1999-1_Main-IG_GJI",
        "label": "3"
    },
    {
        "refID": "1999-1_Sammis-CG_PAGEOPH",
        "label": "3"
    },
```

```json
    {
        "refID": "1999-1_Yang-WZ_ActaSeismSinica",
        "label": "1"
    },
    {
        "refID": "2000-1_Huang-Y_JGR",
        "label": "2"
    },
    {
        "refID": "2000-1_Jaume-SC_GeophysMonogrSer",
        "label": "3"
    },
    {
        "refID": "2000-1_Jaume-SC_PAGEOPH",
        "label": "3"
    },
    {
        "refID": "2000-1_Papazachos-BC_PAGEOPH",
        "label": "2"
    },
    {
        "refID": "2000-1_Robinson-R_GJI",
        "label": "2"
    },
    {
        "refID": "2000-1_Rundle-JB_PAGEOPH",
        "label": "3"
    },
    {
        "refID": "2001-1_Bowman-DD_CRAS",
        "label": "0"
    },
    {
        "refID": "2001-1_Bowman-DD_GRL",
        "label": "0"
    },
    {
        "refID": "2001-1_DiGiovambattista-R_Tectonophys",
        "label": "3"
    },
    {
        "refID": "2001-1_Papazachos-CB_AnnGeofis",
        "label": "2"
    },
    {
        "refID": "2001-1_VereJones-D_GJI",
        "label": "2"
    },
    {
        "refID": "2001-1_Yang-WZ_JGR",
        "label": "2"
    },
```

```json
    {
            "refID": "2001-1_Zoeller-G_JGR",
            "label": "2"
    },
    {
            "refID": "2002-1_BenZion-Y_PAGEOPH",
            "label": "3"
    },
    {
            "refID": "2002-1_Karakaisis-GF_GJI",
            "label": "2"
    },
    {
            "refID": "2002-1_Papazachos-BC_JOSE",
            "label": "2"
    },
    {
            "refID": "2002-1_Papazachos-BC_Tectonophys",
            "label": "2"
    },
    {
            "refID": "2002-1_Papazachos-CB_BSSA",
            "label": "2"
    },
    {
            "refID": "2002-1_Sammis-CG_PNAS",
            "label": "3"
    },
    {
            "refID": "2002-1_Weatherley-D_PAGEOPH",
            "label": "3"
    },
    {
            "refID": "2002-1_Yin-XC_PAGEOPH",
            "label": "2"
    },
    {
            "refID": "2002-1_Zoeller-G_GRL",
            "label": "2"
    },
    {
            "refID": "2003-1_Chen-CC_GJI",
            "label": "2"
    },
    {
            "refID": "2003-1_Helmstetter-A_JGR",
            "label": "2"
    },
    {
            "refID": "2003-1_Karakaisis-GF_GJI",
            "label": "2"
    },
```

```json
    {
        "refID": "2003-1_Karakaisis-GF_PAGEOPH",
        "label": "2"
    },
    {
        "refID": "2003-1_King-GCP_JGR",
        "label": "0"
    },
    {
        "refID": "2003-1_Turcotte-DL_GJI",
        "label": "3"
    },
    {
        "refID": "2003-1_Tzanis-A_NHESS",
        "label": "2"
    },
    {
        "refID": "2004-1_Bowman-DD_PAGEOPH",
        "label": "2"
    },
    {
        "refID": "2004-1_Bufe-CG_BSSA",
        "label": "1"
    },
    {
        "refID": "2004-1_DiGiovambattista-R_Tectonophys",
        "label": "1"
    },
    {
        "refID": "2004-1_Jaume-SC_JGR",
        "label": "2"
    },
    {
        "refID": "2004-1_Karakaisis-GF_Tectonophys",
        "label": "2"
    },
    {
        "refID": "2004-1_Sammis-CG_PAGEOPH",
        "label": "0"
    },
    {
        "refID": "2004-1_Scordilis-EM_JOSE",
        "label": "2"
    },
    {
        "refID": "2004-1_Wang-JC_TAO",
        "label": "2"
    },
    {
        "refID": "2004-1_Wang-LY_ActaSeismSinica",
        "label": "2"
    },
```

```
        {
                "refID": "2004-1_Wang-Y_PAGEOPH",
                "label": "2"
        },
        {
                "refID": "2004-1_Weatherley-D_PAGEOPH",
                "label": "3"
        },
        {
                "refID": "2005-1_Jiang-CS_ActaSeismSinica",
                "label": "2"
        },
        {
                "refID": "2005-1_Jiang-CS_BSSA",
                "label": "1"
        },
        {
                "refID": "2005-1_Papazachos-CB_BSSA",
                "label": "2"
        },
        {
                "refID": "2005-1_Robinson-R_GRL",
                "label": "1"
        },
        {
                "refID": "2006-1_Jiang-CS_PAGEOPH",
                "label": "1"
        },
        {
                "refID": "2006-1_Mignan-A_EPSL",
                "label": "0"
        },
        {
                "refID": "2006-1_Mignan-A_JGR",
                "label": "0"
        },
        {
                "refID": "2006-1_Papazachos-BC_JOSE",
                "label": "2"
        },
        {
                "refID": "2006-1_Zhou-S_JGR",
                "label": "1"
        },
        {
                "refID": "2007-1_Mignan-A_JGR",
                "label": "0"
        },
        {
                "refID": "2007-1_Papazachos-BC_BSSA",
                "label": "2"
        },
```

```json
    {
        "refID": "2007-1_Papazachos-BC_JGR",
        "label": "2"
    },
    {
        "refID": "2008-1_Hardebeck-JL_JGR",
        "label": "1"
    },
    {
        "refID": "2008-1_Mignan-A_AdvGeophys",
        "label": "0"
    },
    {
        "refID": "2008-1_Mignan-A_GRL",
        "label": "0"
    },
    {
        "refID": "2008-1_Mignan-A_Tectonophys",
        "label": "0"
    },
    {
        "refID": "2008-1_Papadimitriou-P_JGR",
        "label": "2"
    },
    {
        "refID": "2009-1_Chouliaras-G_NHESS",
        "label": "1"
    },
    {
        "refID": "2009-1_Greenhough-J_GRL",
        "label": "1"
    },
    {
        "refID": "2009-1_Mignan-A_GRL",
        "label": "1"
    },
    {
        "refID": "2009-2_Chouliaras-G_NHESS",
        "label": "1"
    },
    {
        "refID": "2010-1_Bebbington-MS_PAGEOPH",
        "label": "0"
    },
    {
        "refID": "2010-1_DeSantis-A_Tectonophys",
        "label": "2"
    },
    {
        "refID": "2010-1_Jiang-CS_ConcurrComput",
        "label": "1"
    },
```

```json
    {
        "refID": "2010-1_Mignan-A_Tectonophys",
        "label": "0"
    },
    {
        "refID": "2010-1_Papazachos-BC_JOSE",
        "label": "1"
    },
    {
        "refID": "2012-1_Jiang-CS_EPS",
        "label": "1"
    },
    {
        "refID": "2012-1_Lagios-E_Tectonophys",
        "label": "1"
    },
    {
        "refID": "2012-1_Mignan-A_GRL",
        "label": "0"
    },
    {
        "refID": "2012-1_Pliakis-D_AnnGeophys",
        "label": "3"
    },
    {
        "refID": "2013-1_Bouchon-M_NatureGeo",
        "label": "0"
    },
    {
        "refID": "2013-1_Guilhem-A_GJI",
        "label": "0"
    },
    {
        "refID": "2013-1_Jiang-CS_PAGEOPH",
        "label": "0"
    },
    {
        "refID": "2013-1_Karakaisis-GF_JOSE",
        "label": "2"
    },
    {
        "refID": "2015-1_Bouchon-M_NatureGeo",
        "label": "0"
    },
    {
        "refID": "2015-1_DeSantis-A_Tectonophys",
        "label": "1"
    },
    {
        "refID": "2015-1_Felzer-KR_NatureGeo",
        "label": "2"
    },
```

```
    {
        "refID": "2016-1_Christou-EV_ResGeophys",
        "label": "1"
    },
    {
        "refID": "2017-1_Adamaki-AK_PAGEOPH",
        "label": "1"
    },
    {
        "refID": "2017-1_Kazemian-J_PAGEOPH",
        "label": "1"
    },
    {
        "refID": "2018-1_Huang-H_GRL",
        "label": "0"
    }
]
```